\documentclass[journal,twoside]{IEEEtran}

\usepackage[pdftex]{graphicx}
\graphicspath{{./images/}}

\usepackage{amssymb}
\usepackage{latexsym}

\usepackage{url}
\usepackage{xcolor}

\usepackage{hyperref}
\usepackage{multirow}

\usepackage{lineno}
\usepackage{hhline}
\usepackage{amsmath}
\usepackage[ruled,vlined]{algorithm2e}
\usepackage{eucal}
\usepackage{bm}
\usepackage{subcaption}
\usepackage{booktabs}

\usepackage{array}
\newcolumntype{H}{>{\setbox0=\hbox\bgroup}c<{\egroup}@{}}

\SetKwInput{KwInput}{Input}                
\SetKwInput{KwOutput}{Output}              

\makeatletter
\newcommand{\nosemic}{\renewcommand{\@endalgocfline}{\relax}}
\newcommand{\dosemic}{\renewcommand{\@endalgocfline}{\algocf@endline}}
\newcommand{\pushline}{\Indp}
\makeatother

\makeatletter
\newcommand\notsotiny{\@setfontsize\notsotiny\@vipt\@viipt}
\makeatother

\definecolor{DarkRed}{rgb}{0.5,0.1,0.1}
\definecolor{newcolor}{rgb}{.8,.349,.1}

\newcommand{\mycolor}{black}
\newcommand{\revisecolor}{black}
\def\myfigurescale{0.35}
\def\myboxplotscalesmall{0.22}
\def\myboxplotscalebig{0.275}

\begin{document}

\title{\huge RA-GCN: Graph Convolutional Network for  Disease Prediction Problems with Imbalanced Data}

\date{}

\author{ \small
\IEEEauthorblockN{{\bf Mahsa Ghorbani \thanks{Corresponding authors: Mahsa Ghorbani, Hamid R. Rabiee \newline  Email-addresses: mahsa.ghorbani@tum.de, mahsa.ghorbani@sharif.edu, \newline rabiee@sharif.edu}}
\IEEEauthorrefmark{1}\IEEEauthorrefmark{2}}, 
\IEEEauthorblockN{{\bf Anees Kazi}\IEEEauthorrefmark{2}},
\IEEEauthorblockN{{\bf Mahdieh Soleymani Baghshah}\IEEEauthorrefmark{1}},
\IEEEauthorblockN{{\bf Hamid R. Rabiee} \IEEEauthorrefmark{1}},
\IEEEauthorblockN{{\bf Nassir Navab}\IEEEauthorrefmark{2},\IEEEauthorrefmark{3}},
\and
\\ 
\IEEEauthorblockA{\IEEEauthorrefmark{1} \small Department of Computer Engineering, Sharif University of Technology, Tehran, Iran} \\
\IEEEauthorblockA{\IEEEauthorrefmark{2} \small Computer Aided Medical Procedures, Department of Informatics, Technical University of Munich, Germany} \\
\IEEEauthorblockA{\IEEEauthorrefmark{3} \small Whiting School of Engineering, Johns Hopkins University, Baltimore, USA}
}

\markboth{2021}%
{G\MakeLowercase{horbani} \MakeLowercase{\textit{et al.}}:A\MakeLowercase{ccepted for publication in} M\MakeLowercase{edical} I\MakeLowercase{mage} A\MakeLowercase{nalysis} (2021)}

\maketitle

\renewcommand\IEEEkeywordsname{Keywords}


\begin{abstract}
Disease prediction is a well-known classification problem in medical applications. 
Graph Convolutional Networks (GCNs) provide a powerful tool for analyzing the patients' features relative to each other. This can be achieved by modeling the problem as a graph node classification task, where each node is a patient.
Due to the nature of such medical datasets, class imbalance is a prevalent issue in the field of disease prediction, where the distribution of classes is skewed.
When the class imbalance is present in the data, the existing graph-based classifiers tend to be biased towards the major class(es) and neglect the samples in the minor class(es). On the other hand, the correct diagnosis of the rare positive cases (true-positives) among all the patients is vital in a healthcare system.
In conventional methods, such imbalance is tackled by assigning appropriate weights to classes in the loss function which is still dependent on the relative values of weights, sensitive to outliers, and in some cases biased towards the minor class(es).
In this paper, we propose a Re-weighted Adversarial Graph Convolutional Network (RA-GCN) to prevent the graph-based classifier from emphasizing the samples of any particular class. This is accomplished by associating a graph-based neural network to each class, which is responsible for weighting the class samples and changing the importance of each sample for the classifier. Therefore, the classifier adjusts itself and determines the boundary between classes with more attention to the important samples. \textcolor{\mycolor}{The parameters of the classifier and weighting networks are trained by an adversarial approach. 
We show experiments on synthetic and three publicly available medical datasets.} Our results demonstrate the superiority of RA-GCN compared to recent methods in identifying the patient's status
on all three datasets. The detailed analysis of our method is provided as quantitative and qualitative experiments on synthetic datasets.
\end{abstract}


\section{Introduction}
Disease prediction using medical data has recently shown a potential application area for machine learning-based methods \cite{uddin2019comparing,mohan2019effective}. 
Disease prediction analyzes effective factors for each patient to distinguish healthy and diseased cases or to recognize the type of disease. Predicting the disease state has considerable value in the clinical domain because the identification of patients with a higher probability of developing a chronic disease enhances the chance for better treatments \cite{bayati2015low}. 
Deep neural networks are popular machine learning models that started to outperform other methods in a wide range of domains. Most of the state-of-the-art machine learning methods across a variety of areas adopted deep neural networks and have achieved unprecedented performance on a broad range of tasks over the past years \cite{lundervold2019overview,yamashita2018convolutional}.

\begin{figure}
    \centering
    \includegraphics[scale=0.35]{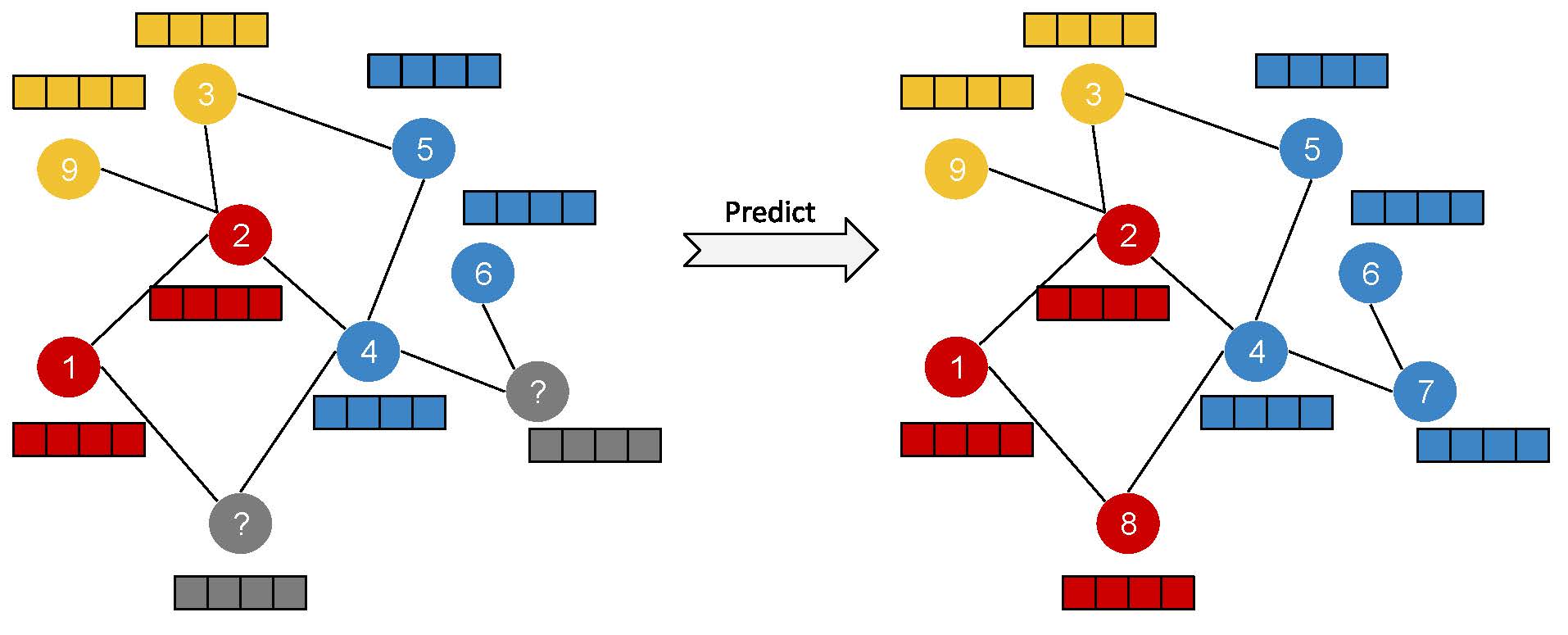}
    \caption{Node classification task. Label for a subset of nodes and the graph between all nodes are available. The goal is to predict the label of unlabeled (grey) nodes.}
    \label{fig:node_classification}
\end{figure}

Prevalent deep neural networks do not incorporate the interaction and association between the patients in their architecture. Considering the relationships between the patients is beneficial as it helps to analyze and study the similar cohort of patients together. By viewing the patients as nodes and their associations as edges, graphs provide a natural way of representing the interactions among a population.
Moreover, when a graph between patients is constructed from a subset of their features, those features are summarized in the graph edges and omitted from the feature set. This results in feature dimensionality reduction to avoids the overfitting due to the large number of features  \cite{bishop2006pattern,caruana2001overfitting,zhang2018feature}.
Consequently, the focus of developing deep learning methods for such data with an underlying graph structure has witnessed a tremendous amount of attention over the last few years \cite{zhang2020deep}. 
Geometric deep learning \cite{bronstein2017geometric} is the field that studies extending neural networks to graph-structured data by including neighborhoods between nodes. Studies in this field explore the methods of generalizing the key concepts of deep learning, such as convolutional neural networks, to graph-structured data and propose graph-based neural network architectures.
By modeling the medical data with a graph, the disease prediction problem turns into the node classification task (depicted in Fig.~\ref{fig:node_classification}) which has been explored recently by utilizing Graph Neural Network (GNN) models \cite{kazi2019graph, parisot2017spectral,ghorbani2019mgcn}. As shown in  Fig.~\ref{fig:node_classification}, the task is to predict the class label of the unknown test samples within the cohort (patients with gray nodes).

However, there are still open challenges, especially in the biomedical datasets including missing values \cite{cheng2020novel}, dataset size \cite{ashraf2020deep}, and low contrast imaging \cite{subramani2020fast}. Such challenges hinder the optimal performance of any model for the classification task. 
One of these challenges is also the class imbalance which happens when the number of samples across the classes are disproportional. The class imbalance can vary from a slight to a severe one where the major class contains tens, hundreds, or thousands of times more samples than the minor one.
It is common to describe the imbalance of the dataset in terms of imbalance ratio (ratio of the major class size to the minor class \cite{sun2009classification}) or summarize the distribution as a percentage of samples in each class. 
Most of the algorithms and loss functions for training a classifier aim to maximize accuracy metric for training data, so they have reasonable outcomes when the classification task is defined on the datasets with balanced classes \cite{sun2009classification}. 
However, in an imbalanced setting, maximizing the accuracy encourages the classifier to favor the major class. Accordingly, a high overall accuracy classifier does not imply a well-discriminating decision boundary between the major and minor classes in imbalanced datasets.

In this paper, we propose a model that handles the class imbalance by modifying the objective function and prevents bias towards either of the classes. We adopt a well-known variant of Graph Convolutional Networks (GCNs) proposed by Kipf et al. \cite{kipf2016semi} which has achieved high performance for the node classification task in different applications (See review paper \cite{wu2020comprehensive}). 
In our proposed model, in addition to the GCN-based classifier, a class-specific GCN is used which is responsible for providing a weight distribution over the training samples of the corresponding class. These weights are employed in the weighted cross-entropy as the objective function. Modifying the objective function by giving more or less penalty on samples, purposefully biases the classifier to provide a more accurate prediction of high weighted samples and build a discriminating hidden space across them. The classifier and the weighting networks are trained in an iterative adversarial process. The classifier minimizes the penalty of misclassification and the weighting networks maximize the weights of the misclassified nodes against the classifier.
The proposed method is named Re-weighted Adversarial Graph Convolutional Network (RA-GCN) which is designed for node classification task in the imbalanced datasets.
Our contributions are:
\begin{itemize}
    
    \item \textcolor{\mycolor}{A novel model to address class imbalance for disease prediction using GCNs.}

    \item Proposing separate weighting networks that learn a parametric function for weighting each sample in each class to help the classifier fit better between classes and avoid bias towards either of the classes by adopting an adversarial training approach.
    \item  Proposing a simple and easy-to-learn model, \textcolor{\mycolor}{compared to the previous methods}, by weighting the existing training samples instead of generating new samples.
    \item Demonstrating RA-GCN's ability to enhance the performance of the GNNs with experiments on both real medical and designed synthetic datasets which are similar to real ones in the number of samples and features. In this paper, we demonstrate the superiority of RA-GCN in terms of performance and stability over the recent methods. 
\end{itemize}
\label{sec:introduction}
\section{Related Work}
Our method is mainly related to GCNs, their importance in disease prediction, and the effect of class imbalance in medical datasets on the decision of the classifiers. Accordingly, this section is dedicated to reviewing the literature on the related areas.  \\
\textbf{Graph Convolutional Networks}: Due to the success of convolutional neural networks on common and regular domains such as computer vision and speech recognition, most of the GNNs are devoted to \textcolor{\revisecolor}{re-defining} the concept of convolution on graphs (see review paper of \cite{wu2020comprehensive}). Depending on the definition of filtering, GNNs are categorized into spectral-based (as reviewed in \cite{shuman2013emerging,bruna2013spectral,defferrard2016convolutional,bronstein2017geometric}) and spatial-based models (as reviewed in \cite{zhang2019graph}).
ChebyNet \cite{defferrard2016convolutional} is among the known spectral methods which developed a deep learning approach to deal with graph-structured data. Further, \cite{kipf2016semi} proposed \textcolor{\mycolor}{a variant of GCNs} which simplified the ChebyNet without the need to perform the convolution in the spectral domain and bridged the gap between spectral and spatial-based approaches \cite{balcilar2020bridging}. 
 Many variants of GCNs are proposed which include applying the attention mechanism to the graphs \cite{velivckovic2017graph}, increasing the receptive field of convolutional filters \cite{li2019label}, and improving the scalability by sampling techniques \cite{hamilton2017inductive,rossi2020sign,ramezani2020gcn}.
\\
\textbf{GCNs in disease prediction}: In recent years, GCNs have been adopted in different applications, especially in medical domains e.g., brain analysis \cite{li2020braingnn}, mammogram analysis \cite{du2019zoom}, and segmentation \cite{soberanis2020uncertainty}. The disease prediction problem has been also widely explored by GCN-based methods \cite{parisot2017spectral,anirudh2019bootstrapping,li2020graph,ravindra2020disease}.
Following the promising results of GCNs in the medical domain, \cite{parisot2017spectral} exploit GCNs in the disease prediction problem with multi-modal datasets. They chose ChebyNet with constant filter size as the classifier and investigated the affinity graph construction by computing the distance between the subjects for brain analysis. InceptionGCN \cite{kazi2019inceptiongcn} extends ChebyNet and designs filters with different kernel sizes, and chooses the optimal one for the final disease prediction. A branch of methods focuses on improving the graph structure. \cite{huang2020edge} and \cite{song2021graph} start with a pre-constructed graph and update it during the training, but \cite{cosmo2020latent} learn the whole graph from features directly in an end-to-end manner. Besides the methods that concentrate on the graph structure, a set of methods investigate the missing-value problem as an important challenge of medical datasets in disease prediction \cite{wu2020inferring,wang2020graph}. However, to the best of our knowledge, none of the previous work consider the class imbalance problem in the graph-based methods for disease prediction.
\\
\textbf{Class imbalance in non-medical and medical applications}: Methods for handling class imbalance for non-graph datasets can be categorized as data-level and cost-sensitive approaches. In the data-level approach, re-sampling is done to balance the classes \cite{burnaev2015influence,dong2018imbalanced,rayhan2017cusboost}. There are different forms of re-sampling such as randomly or targeted under-sampling the major class(es) and oversampling the minor one(s). 
Even in solving class imbalance for non-graph datasets, oversampling the minority class(es) causes the overfitting to the duplicate samples drawn from the minor class(es). On the other hand, under-sampling the majority class(es) causes the exclusion of samples that are required for discrimination \cite{lin2020boosting}. These weaknesses lead many methods to cost-sensitive approaches. Modifying the cost function is a common solution for dealing with the problem \cite{zhang2018cost,cenggoro2018deep,sze2017weight}. This approach alleviates the problem by assigning different weights to each class, so the misclassification of samples in the minor class(es) penalizes the classifier more intensely \cite{wang2020packetcgan}. The definition of class weights is a key point to address in training the classifier \cite{sun2009classification}.
It should be noted that in this approach, all the samples in each class have the same weight. 
The idea of re-weighting the samples has been also employed in the boosting algorithms (e.g., AdaBoost  \cite{freund1996experiments}). However, boosting algorithms have a different approach to learn the classifier. Boosting algorithms improve the prediction by training a sequence of weak models where each of them compensates for the weaknesses of its predecessors, and their weighted linear combination forms a strong classifier. AdaBoost is a boosting algorithm that iteratively identifies misclassified samples and increases their weights which makes the next classifier put extra attention on them. On the contrary, RA-GCN learns one strong classifier instead of an ensemble of weak classifiers. Furthermore, RA-GCN learns the weighting function based on the classifier’s output in an adversarial manner, instead of using a pre-defined re-weighting function. Due to the sensitivity of decision-making in the medical domain, some methods exclusively concentrate on the problem of class imbalance in this field \cite{zhang2018imbalanced,zhao2018framework}. Cancer \cite{fotouhi2019comprehensive}, diabetes \cite{alghamdi2017predicting}, Parkinson's \cite{vuttipittayamongkol2020improved}, and Alzheimer's diagnosis \cite{cruz2018class} are some examples that have been studied widely in this domain.
\\
\textbf{Class imbalance in graph-structured data}: 
The studies related to the imbalanced classification in graph-structured data are limited. One of the reasons is that re-sampling is challenging in the context of graphs since each data sample not only contains features but also has relations to others.  Although the generation of features for a new sample has several approaches, adding a new sample to the graph needs to determine its relation \textcolor{\mycolor}{ with respect to the rest of the samples}. Additionally, over-sampling changes the original graph structure, which is an influential factor in the trained model. On the other hand, removing data samples from a class might affect a significant number of connections and may create disconnected components. 
Dual Regularized GCN (DR-GCN) is a variant of GCNs dedicated to addressing the class imbalance problem in the GCN architecture \cite{ijcai2020-398}. 
DR-GCN has three components. First one is the GCN-based classifier.
As second component, a conditional Generative Adversarial Net (cGAN) including a discriminator and generator is added to the proposed architecture. 
After the training, the generator can generate samples with the desired label and the discriminator can discriminate the fake and real samples. Since the low-dimensional representation of nodes from the classifier's hidden space is the input to the cGAN, the first component gets updated during the training of cGAN and the separation between classes will be enhanced in its hidden space. In order to prevent the network from overfitting with the labeled samples, a distribution alignment between the labeled and unlabeled samples is imposed as the third component. For this purpose, it is assumed that the labeled and unlabeled samples followed two multivariate Gaussian distributions. The goal of the third component is to minimize the Kullback-Leibler divergence between these two distributions. Cross-entropy without any modifications is used as the main objective function and the second and third components regularize the classifier to avoid overfitting to the imbalanced data.
Although both the proposed method (RA-GCN) and DR-GCN use adversarial training to face the class imbalance issue, they have totally different approaches. DR-GCN uses adversarial training between the discriminator and the conditional generator of cGAN along with the training process of the main classifier. This causes the DR-GCN to struggle with the convergence and stability problems in the training of cGAN. The problem becomes serious in the domains with a limited number of data for training, such as medical datasets. However, the proposed RA-GCN utilizes an adversary between the main classifier and the weighting networks with the purpose of learning weights for training samples which is a much simpler and easier task than generating new samples as in DR–GCN. In addition to that, DR-GCN forces the separation between the classes in the hidden space of the classifier by adding another cGAN with its own loss function. Nevertheless, RA-GCN trains the classifier's parameters directly by modifying the classification loss function and penalizing the misclassification based on the sample weights which are learned by weighting networks.

\section{Method}
Graph-based deep learning methods have recently become popular because of the novelty and success in different applications.
One of the important flaws of recently proposed GNNs, is their inability to deal with imbalanced datasets. 
In this section, we provide the details of our method to dynamically address the class imbalance in GCNs. To this end, we propose a model to simultaneously learn graph-based weighting networks to automatically weight the loss value calculated by the classifier's output for each sample. In the following, we first provide the notation and definition of the problem, next an overview of the specific type of graph-convolutional layer (GC-layer) utilized in our model is described. Finally, the proposed method is discussed in detail.
\subsection{ Problem Definition and Preliminary}
Assume that the graph $G$ is given with $N$ nodes, represented by $G(V, E, X)$, where $V$ is the set of nodes ($|V| = N$), $E$ is the set of edges, and $X \in \mathbb{R}^{N\times F}$ indicates the node feature matrix.
$A \in \mathbb{R}^{N\times N}$ is the unweighted and undirected adjacency matrix of the graph. \textcolor{\mycolor}{Every node $v_i$ has a corresponding feature vector $x_i$, a one-hot label vector $y_i$, and the true class label $c_i \in C$ where $C$ is the set of classes. The label information is available for a subset of nodes and the task is to learn the parametric function $f_{\theta}(X,A)$ which takes the adjacency matrix and node features as input, and its goal is to predict the true label of the unlabeled nodes. It should be noted that a probabilistic classifier predicts a probability distribution over the $|C|$ classes and the class with maximum probability is selected as the label. In our proposed method $q_i$ is the output probability distribution of classifier defined over $|C|$ classes for the sample $x_i$ where $c$-th element represents the confidence of the classifier about assigning label $c$ to $x_i$. Thus, the problem will be formulated as follow:}
\begin{linenomath*}
\begin{equation}
    Q =  f_{\theta}(X,A),
\end{equation}
\end{linenomath*}
where $Q\in \mathbb{R}^{N\times |C|}$ is the prediction matrix of the classifier for all nodes including unlabeled ones.

\subsection{Background: Graph Convolutional Network (GCN)}
\label{sec:background}
In traditional neural networks (such as multi-layer perceptron with fully connected layers), there is no explicit relation between the data samples, and they are assumed to be independent.
GCNs aim to take the neighborhood graph between the samples into consideration and create the feature map of each node not only by its own features but also using its neighbors \cite{park2020sumgraph}.
For our method, we employed the definition of GC-layer proposed by Kipf et al. \cite{kipf2016semi} in their method which is a spectral-rooted spatial convolution network that has shown prospering results in the node classification task \cite{zhang2019graph} and has influenced many other contributions \cite{balcilar2020bridging}.
The utilized GC-layer first aggregates each node's features and its neighbors (based on the structure of the graph) and then finds a latent representation for each node using a fully connected layer (FC-layer). 
Assume $A\in \mathbb{R}^{N\times N}$ is the adjacency matrix of the given graph and  $X \in \mathbb{R}^{N\times F}$ indicates the features of nodes. Let $Deg$ be a degree matrix, and $I_N$ be the identity matrix of size $N$. 
The graph convolution which we also use in our method comprises of the following two steps:

    \textbf{Step1}: $X' = \hat{Deg}^{-\frac{1}{2}}\hat{A}\hat{Deg}^{-\frac{1}{2}}X$, where $ \hat{A} = A + I_N$ and $\hat{Deg}_{ii}=\sum_j \hat{A}_{ij}$.
    
    \textbf{Step2}: $Z = f_\theta(X')$, where $f_\theta$ is an FC-layer with parameters $\theta$ and an arbitrary non-linear activation function.

In the first step, the input graph structure is changed by adding a self-loop for every node, and then the features of each node are replaced by an average between the node features and its neighbors in the graph. Then, in the second step, the updated features are given to an FC-layer for mapping to a latent space.
In the rest of the paper, "GCN" refers to a network composed of the described GC-layers except otherwise expressed.

\subsection{The Proposed Model}
\begin{figure*}[!htb]
    \centering
    \includegraphics[width=\textwidth,height=\textheight,keepaspectratio]{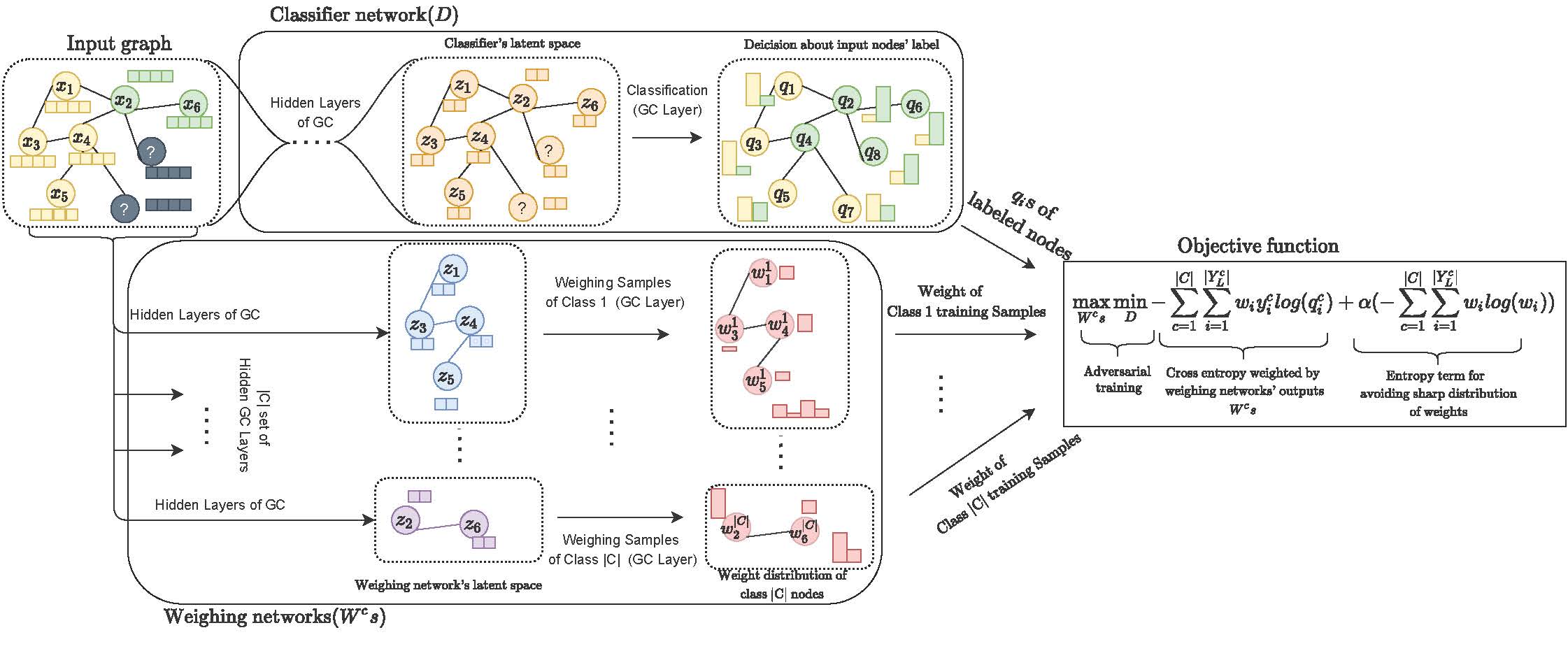}
    \caption{The proposed RA-GCN model. The input graph with the corresponding node features is processed by two components. In the first one (the upper part), the classifier, after processing the input data by several layers of GC, maps the input to a new features space. In the final step of classification, the classifier predicts the label of each input node which indicates its confidence about assigning the input node to every class ($q_i$s).
    In the second component (the lower part), a weighting network composed of multiple GC-layers is utilized for each class to assign appropriate weight to the samples of the corresponding class so as to improve the classifier's performance and prevent it from biasing towards either of the classes. The outputs of the components go to the objective function for updating their parameters. The classifier intends to minimize the objective and the weighting networks aim to maximize it in an adversary manner.}
    \label{fig:aric_model2}
\end{figure*}

Weighted cross-entropy loss ($\mathcal{L}_{WCE}$) is a prevalent objective function for the classification of imbalanced data, which is defined as follows:
\begin{linenomath*}
\begin{equation}
\label{eq:wce}
    \mathcal{L}_{WCE} = -\sum_{c=1}^{|C|} \sum_{i=1}^{|Y_L^c|}  \beta^c y_{i}^{c} \; log(q_{i}^{c}),
\end{equation}
\end{linenomath*}
where $y_{i}^{c}$ and $q_{i}^{c}$ are the $c$-th element of $y_i$ and $q_i$, respectively, and $Y_L^c$ is the set of labeled nodes with label $c$. $\mathcal{L}_{WCE}$ needs predefined weights for samples in every class ($\beta^c$). This is usually done by weighting samples of each class proportional to the inverse of the class frequency \cite{cui2019class}.  
However, the class-weighting method assigns the same weight for the incorrect samples as the samples that are already classified correctly. We modify this approach by learning appropriate weights for all samples by dynamically increasing the weights of misclassified samples against the true-classified ones. In this way, the classifier is forced to refine its decision boundary and focus on the misclassified samples to correct them for both minor and major classes.
Fig.~\ref{fig:aric_model2} depicts an overview of the proposed model. 

\begin{figure}[!htb]
    \centering
    \includegraphics[scale=0.28]{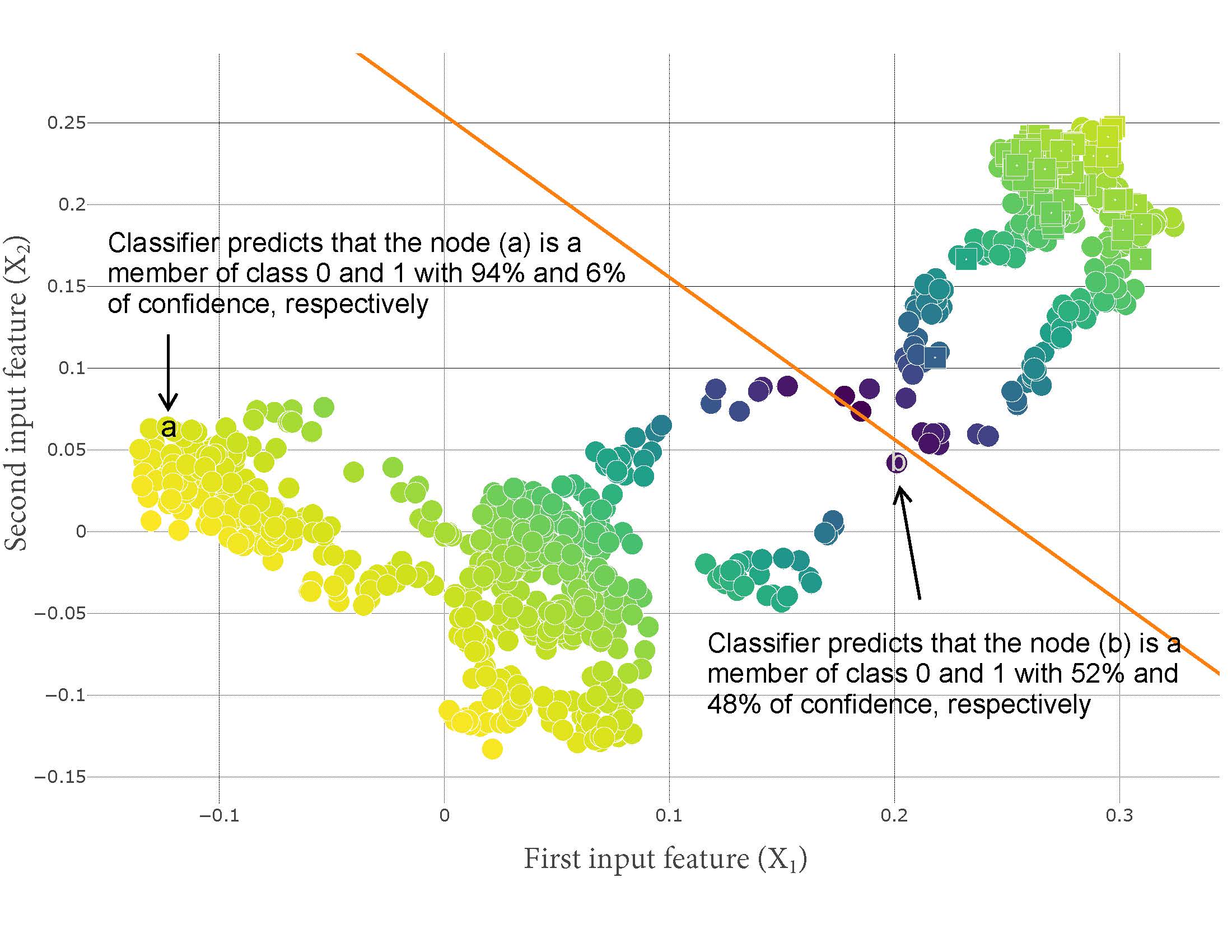}
    \caption{Confidence of the classifier in the classification task. Node (a) is an easy sample for classification since it is far from the decision boundary and the classifier is confident about its decision for node (a)}. Node (b) is more challenging because its features are more similar to the nodes with the opposite label (squared) and the classifier assigns it to both classes with close probabilities.
    \label{fig:class-prob}
\end{figure}

\begin{figure}[!htb]
    \centering
    \includegraphics[scale=0.28]{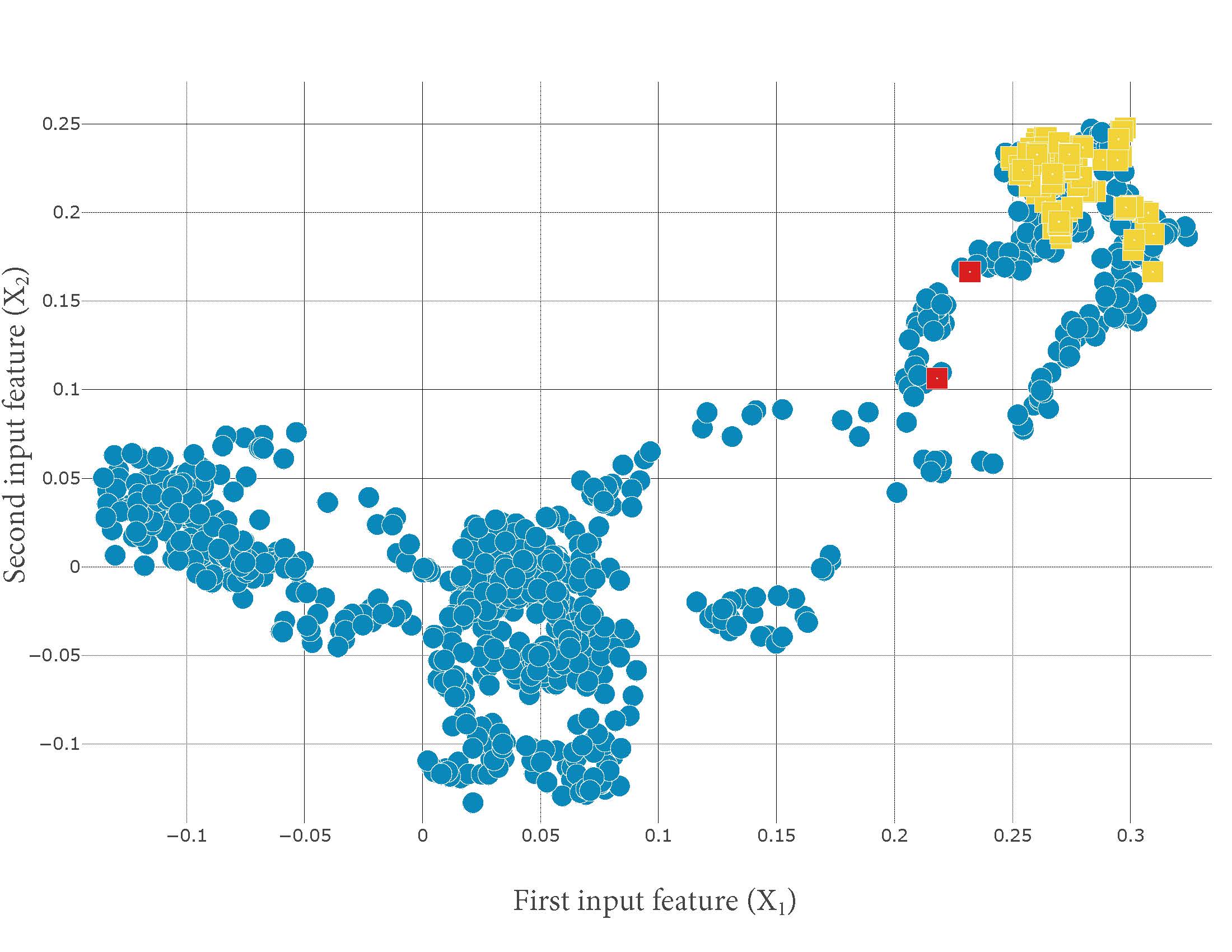}
    \caption{Outliers in the dataset. Circles (blue samples) are the samples from the first class and squares (yellow and red samples) are the samples from the second class. Samples in red have an abnormal distance from other points with the same label (yellow ones). These points are called outliers. Outliers can be the result of many circumstances such as an error in experiments, data collection, or labeling.}
    \label{fig:outlier}
\end{figure}

As it is demonstrated in Fig.~\ref{fig:aric_model2}, RA-GCN consists of two components: 1) A Classifier network, 2) A set of weighting networks which are described below. \\
\textbf{Classifier network  ($D$)}: 
In the proposed model, $D$ is a generic arbitrary classifier. In our method, we use GCN which takes the adjacency matrix and node features and its task is to predict the class label of the input nodes. As we said before, a probabilistic classifier predicts a probability distribution over classes. Fig.~\ref{fig:class-prob} represents that the classifier predicts the label for node (a) with high probability, but node (b) is more challenging to decide. Since our classification task has a single label per node, we use a softmax activation function in the final layer of classifier $D$.
\\ 
\textbf{Weighting networks ($W^cs$)}: 
In order to learn the weighting function for samples of each class, a separate GCN is dedicated to each class to learn the weights for its samples. The weighting network (W-GCN) for class $c$ is denoted by $W^c$. Every $W^c$ takes the samples from class $c$ and the adjacency matrix between them as input and assigns a weight to each sample.
It is worth noting that GCN utilizes the neighbors of a node if they exist; otherwise, a node is considered as an isolated one with a self-loop edge. Therefore, if taking the class-specific sub-graphs for weighting networks leads to isolated nodes, GCN still works properly.
To limit the weights of samples and make the sum of weights in every class in the same range, we normalize the sample weights of each class using a softmax function.
\\ \textbf{End-to-end adversarial training process}: As discussed above, to emphasize the misclassified samples and the correctly classified ones but with low confidence, we design an adversary process between the classifier network $D$ and the weighting networks $W^{c_i}$s. To this end, we adopt the weighted cross-entropy as the classifier loss function and make the weighting networks to provide the sample weights dynamically for the classifier.
We use a two-player min-max game \cite{goodfellow2020generative} to formulate the adversarial process, where the classifier and the set of weighting networks both try to optimize the following objective:
\begin{linenomath*}
\begin{equation}
\small
 \max_{W^cs} \min_{D} \underbrace{-\sum_{c=1}^{|C|} \sum_{i=1}^{|Y_L^c|}  \; w_{i}y_{i}^{c} \; log(q_{i}^{c})}_{\substack{\text{\scriptsize Dynamic weighted cross entropy}}},
 \label{eq:ra-gcn-init}
\end{equation}
\end{linenomath*}
such that $q_i$ and $w_i$ are the outputs of networks $D$ and $W^{c_i}$ respectively, for the sample $x_i$ ($W^{c_i}$ is the corresponding weighting network for $x_i$).
This is adversarial training because while the classifier minimizes the objective function, W-GCNs have to maximize it. 
During the training, the classifier tries to correctly classify the training samples, especially the samples with high weights, to minimize the Eq. \ref{eq:ra-gcn-init}. Whereas, the adversary makes the weighting networks update their parameters by putting larger weights on challenging nodes (the samples that are not correctly classified or those on which the classifier is not confident about its decision) to maximize the objective function. However, as explained before, In order to prevent the weighting networks from increasing the sample weights without limitation and do not prioritize any class to another one, the total summation of sample weights in each class is limited by applying a softmax normalization.
Alternatively, when the provided weights by weighting networks increase the objective function value, the classifier faces a high penalty. This makes the classifier update its parameters and try to boost its performance with respect to the latest generated weights without bias towards either of classes.
Analogous to \cite{goodfellow2020generative}, the parameters of $W^c$s have to update after the training of $D$ is completed, however optimizing $D$ to full convergence is highly expensive; therefore, we update $W^cs$ for one step after updating the classifier $D$ for $k$ steps.
At the end of the adversarial game, there will be an equilibrium between the classifier and the weighting networks. We hypothesize that the dynamically changing node weights put more attention on the samples that might be ignored when the classifier is trained by the conventional WCE or CE loss functions.
However, in the real datasets, there are samples in every class that are significantly different from other observations. In Fig.~\ref{fig:outlier}, nodes colored in red are such samples that have the same labels as yellow nodes but lie at an abnormal distance from other samples. These samples are conventionally called outliers. In the presence of the outliers, the weighting networks put more weight on outliers to maximize the penalty for the classifier. This makes the classifier overfit on outliers or alternate between the correct classification of outliers and finding an appropriate boundary which results in instability.
Using GCN as weighting network is beneficial to overcome this problem by avoiding sharp weighting distributions. This happens by smoothing the features of each node by its same-labeled neighbors and generating a softened weight distribution as a result.
In addition to that, we add another term and call it "entropy term". The entropy term is the sum of weight distribution entropies. This term acts as a regularizer that penalizes such sharp weighting distributions where a few samples are assigned a large weight compared to other samples. $\alpha$ is a free coefficient that determines the importance of the regularizer term against the weighted cross-entropy. By adding the entropy term, the overall objective function becomes as follow: 
\begin{linenomath*}
\begin{equation}
\small
 \max_{W^cs} \min_{D} \underbrace{-\sum_{c=1}^{|C|} \sum_{i=1}^{|Y_L^c|}  \; w_{i}y_{i}^{c} \; log(q_{i}^{c})}_{\substack{\text{\scriptsize Dynamic weighted cross entropy}}} +
 \alpha \underbrace{(-\sum_{c=1}^{|C|} \sum_{i=1}^{|Y_L^c|} w_{i} \; log(w_i))}_{\substack{\text{\scriptsize 
 Entropy term
 }}}
 \label{eq:ra-gcn}
\end{equation}
\end{linenomath*}
Eventually, for the testing phase, we remove the weighting networks $W^c$s and obtain an efficient classifier for the test set.
Algorithm \ref{alg:ra-gcn} describes the training steps of RA-GCN. 

\begin{algorithm}[!htb]
\SetAlgoLined
  \KwInput{Attributed Graph $G$, node features $X$, adjacency matrix $A$, and labeled nodes $Y_L$ }
  \KwOutput{Classifier for node classification (D)}
  \For{number of training iterations}
  {
  \For{k steps}
  {
  Get the output of the classifier network ($q_i$s for training samples)\;
  Get the outputs of weighting networks ($w_i$s for training samples) \;
  Update the parameters of the classifier ($\theta_D$) by descending their gradient : \\
  \pushline \pushline
  
  $\nabla_{\theta_D} \; -\sum_{c=1}^{|C|} \sum_{i=1}^{|Y_L^c|}  \; w_{i}y_{i}^{c} \; log(q_{i}^{c})$ \\
  }
  Get the output of the classifier network\;
  Get the output of weighting networks \;
  Update the parameters of each weighting network ($\theta_{W^c}$) by ascending the gradient: \\
  \pushline 
  \footnotesize
  $\nabla_{\theta_{W^c}} -\sum_{i=1}^{|Y_L^c|}  \; w_{i}y_{i}^{c} \; log(q_{i}^{c}) + \alpha (-\sum_{i=1}^{|Y_L^c|} w_{i} \; log(w_i))$
  
  }
  \caption{Training Procedure of RA-GCN. $k$ is the number of steps to update the classifier towards the convergence. We used $k=1$ in our experiments to avoid expensive updates.}
  \label{alg:ra-gcn}
\end{algorithm}

\section{Experiments and Results}
To evaluate our model's performance as well as other existing methods, we construct the population graph based on a subset of features in the original imbalanced datasets.
Unlike \cite{cosmo2020latent,parisot2017spectral}, finding the best way for graph construction is not the focus of this paper. Therefore, we follow the similar steps of \cite{parisot2017spectral} for graph construction, explained in the next subsection. Moreover, to keep the setting simple and avoid the effect of edge weights on the classification task, a binary graph is constructed between samples.

Our experiments are divided into two sets of studies on real and synthetic datasets. For each one, we introduce the datasets, the corresponding task, the experimental setup, and finally, the results and discussion of the experiments are provided. 

\subsection{Graph Construction}
\label{sec:graph_cons}
For graph construction, one or multiple features from the original datasets are chosen (called $X^{adj}$). We compute the distance between every pair of nodes according to $X^{adj}$ and connect the nodes within a distance less than a threshold $\gamma$. $\gamma$ is the hyper-parameter chosen empirically (e.g., cross-validation) for each dataset. Mathematically, the adjacency matrix between samples (nodes) is defined as follows:
\begin{linenomath*}
\begin{equation}
\label{eq:graph-dist}
 a_{ij} = \begin{cases} \mbox{1,} & \mbox{if } dist(x_i^{adj},x_j^{adj})<\gamma \\ \mbox{0,} & \mbox{otherwise} \end{cases}
\end{equation}
\end{linenomath*}
 where $a_{ij}$ is the element in the $i$-th row and $j$-th column of adjacency matrix ($A$) and $x^{adj}_i$ represents the adjacency feature(s) of node $i$. It should be noted that the affinity graph is constructed between all the pairs of nodes (train, validation, and test) as we follow the transductive setting. After graph construction, $X^{adj}$ will be excluded from the input features to avoid redundancy.
 
 \subsection{Method Comparisons}
One of the most effective methods for dealing with the class imbalance problem is the weighted cross-entropy (Eq.\ref{eq:wce}), and the weights are given by:
\begin{linenomath*}
\begin{equation}
 \beta^c = 1-\frac{|Y_L^c|}{\sum_{i=1}^{|C|}|Y_L^i|},
\end{equation}
\end{linenomath*}
where $|Y_L^c|$ is the number of nodes in the training set with label $c$. The intuition behind this approach of weighting is to make the sum of the sample weights of all classes equal.

We compare the results of our proposed method with multi-layer perceptron (MLP) as a non-graph feature-based model and the GCN proposed by Kipf et al. \cite{kipf2016semi} as the baselines since they are successful neural networks for the classification task \cite{wu2020comprehensive}.
As it is described in Section \ref{sec:background}, GCN utilizes FC-layers in its architecture, but it incorporates the graph between samples at the start of every layer. Therefore, depending on the graph structure, the extra information in the graph for GCN can either improve the class imbalance by enforcing the separation between classes, or it can intensify the imbalance problem by the domination of major class samples throughout the feature propagation \cite{ijcai2020-398}.
Despite the strength of MLP and GCN, the class distribution is not considered in their architecture. To deal with this issue in training both methods, we use weighted cross-entropy as loss function and add "weighted" to their name in experiments.
DR-GCN \cite{ijcai2020-398} is the recent method dedicated to the class imbalance problem in the GCN. DR-GCN does not follow conventional methods. It adds regularization terms to enforce the separation between classes in the latent space by including cGAN.  

\subsection{Metrics}
In the following experiments, accuracy, macro F1, and ROC AUC are reported as measurements to assess the compared methods. The value of all the metrics is in the range $[0,1]$, and the higher values show better classification performance. 
Accuracy is one of the most frequently used metrics in classification. High accuracy is achievable in an imbalanced dataset by a biased model towards the major class. However, along with other metrics, accuracy can reflect the effect of the class imbalance problem on the classifiers’ performance \cite{branco2016survey}. For measuring accuracy, all samples, without considering their class, contribute equally to compute the metric.  To reflect the classifiers’ performance on the minor classes as well as the major classes, we add ROC AUC and macro F1. Since in their calculation, each metric is computed independently for each class and then averaged (by treating all classes equally), they are good measures to be studied besides the other metrics when the class imbalance exists in the dataset.  ROC AUC is a single-valued score widely used in the presence of class imbalance \cite{he2013imbalanced}, but it should be noted that it can be sub-optimal in case of the low sample size of minority instances \cite{fernandez2018learning}. Therefore, to thoroughly analyze the models’ performance, macro F1 is also utilized as a preferable and reliable metric. F1 score is the weighted average of precision and recall, and both false positives and false negatives are considered in its calculation. When false positive and false negative have similar costs, accuracy is a better measurement, but in uneven class distribution, where the cost of classes are different, we should look at both precision and recall \cite{das2020progress}.
The details about the definition of the metrics are provided in \ref{supsec:metric}.

\subsection{Experiments on Real Datasets}
\label{sec:exp-real}
In this section, we evaluate the proposed RA-GCN on three real datasets. The datasets are Pima Indian Diabetes (Diabetes) \cite{smith1988using}, Parkinson's Progression Markers Initiative (PPMI) \cite{marek2011parkinson}, and Haberman's survival (Haberman) \cite{haberman1976generalized}.  All the datasets are medical datasets, and they are different in the number of samples, the number of features, and the imbalance ratio to cover different challenges in the targeted problem.
An overview of the real datasets is provided in Table~\ref{tab:real_data}. In the following section, we will describe each dataset in detail.

\begin{table}[!htb]
\centering
\caption{Details about the real datasets. Table has the characteristics of the real datasets including the number of samples, number of features, density of graph, and imbalance ratio. All the datasets have two classes, and the imbalance ratio is reported as the number of samples in the major class divided by the number of samples in the minor one.}
\notsotiny
\begin{tabular}{|c|c|c|c|c|c|}
\hline
Datasets & \begin{tabular}[c]{@{}c@{}}No. of\\ Samples\end{tabular} & \begin{tabular}[c]{@{}c@{}}No. of\\ Features\end{tabular} & \begin{tabular}[c]{@{}c@{}}No. of Features\\ for Graph\end{tabular} & Adj Density & \begin{tabular}[c]{@{}c@{}}Imbalance\\ Ratio\end{tabular} \\ \hline \hline
Diabetes    & $768$            & $7$      & 1       & $0.086$               & $500/268   = 1.866$ \\ \hline
PPMI        & $324$            & $300$    & 4       & $0.0005$              & $249/75   = 3.320$  \\ \hline
Haberman    & $306$            & $2$      & 1        & $0.076$               & $225/81  =    2.778$ \\ \hline
\end{tabular}
\label{tab:real_data}
\end{table}
\subsubsection{Datasets}
\textbf{Pima Indian Diabetes (Diabetes) \cite{smith1988using}}: The dataset is produced by the "National Institute of Diabetes and Digestive and Kidney Diseases". The goal of this dataset is to recognize the diabetic status of patients (binary classification).
Every patient has $7$ numeric features which show the diagnostic measurements of diabetes including the number of pregnancies, plasma glucose, blood pressure, skin thickness, Body Mass Index  (BMI), insulin level, diabetes pedigree function, and age. We use plasma glucose of patients for graph construction  (with $\gamma = 4$ in Eq. \ref{eq:graph-dist}) and the rest measurements as node features.

\textbf{Parkinson's Progression Markers Initiative (PPMI)  \cite{marek2011parkinson}}: This dataset is about detecting Parkinson’s disease vs. normal samples (binary classification). PPMI dataset contains brain MRI and non-imaging information. Non-imaging information includes Unified Parkinson’s Disease Rating Scale (UPDRS is a numeric rating used to gauge the severity and progression of Parkinson's disease), Montreal Cognitive Assessment scores (MoCA is a cognitive screening test for detecting brain diseases with numeric scores), and demographic information (age and gender). The MRI images are processed and then are used as the features of samples. A 3D-autoencoder is used to encode raw image intensities, and the encoded representation is used as the input features. More details about this process are described in \cite{baur2018deep} and \cite{vivar2020simultaneous}. 
Non-imaging features are utilized for graph construction (patients with equal test scores in UPDRS and MOCA who have the same gender and have age difference less than 2 are connected). The encoded representation by trained 3D-autoencoder with imaging features are used as node features.

\textbf{Haberman's survival (Haberman) \cite{haberman1976generalized}}: This dataset is the output of the study about patients' survival status who had undergone surgery for breast cancer, which was done between 1958 and 1970 at the University of Chicago's Billings Hospital. The dataset has two classes for predicting the survival status of patients.
The dataset contains numeric features including the number of auxiliary nodes (Lymph Nodes), age, and operation year. The number of auxiliary nodes is used for graph construction (with $\gamma = 2$) and the rest features are node features.

\subsubsection{Implementation Details}
In all experiments, for RA-GCN and also the rest of the methods, the hyper-parameters are chosen based on their best performance on the validation set. In RA-GCN, for the classifier, we chose a two-layer GCN (one hidden layer and one output layer) with 4 hidden units for Diabetes and PPMI and 2 hidden units for Haberman datasets. To simplify our model, we use the same structure and setting for the class-specific weighting networks. We adopted a two-layer GCN with two hidden units for the weighting networks in the following experiments. Rectified Linear Unit (ReLU) \cite{nair2010rectified} is used as the activation function of the hidden layers. We have also applied a dropout layer with a dropping rate of 0.5 
to avoid overfitting.
The entropy term's coefficient $\alpha$ is set to 0.5, 0.1, 1 for Diabetes, PPMI, and Haberman datasets respectively.
In order to train the networks regarding the objective function in Eq.~\ref{eq:ra-gcn}, stochastic gradient descent using Adam optimizer \cite{kingma2014adam} is used with the learning rate of 0.001 for the classifier $D$ and 0.01 for the weighting networks.
We use $60\%$, $20\%$, and $20\%$ for train, validation, and test split, respectively in all datasets and the imbalance ratio is kept in the splits.
Inspired by Parisot et al. \cite{parisot2017spectral}, the absolute difference between the features is used as a distance for graph construction, and the graphs are simple.
All the implementations are in PyTorch \cite{paszke2017automatic} and the results of competitors are obtained with the authors' source code.
For all methods, the results are reported on the test set using the best configuration (selected based on the macro F1 value on validation set) for each method per dataset.  The reported results are the mean and standard deviation of all the metrics for each method on $5$ different random splits of data.

 \subsubsection{Results and Discussion}
\begin{figure*}[!htb]
    \centering
    \begin{subfigure}[b]{\textwidth}
        \centering
        \includegraphics[scale=\myboxplotscalebig]{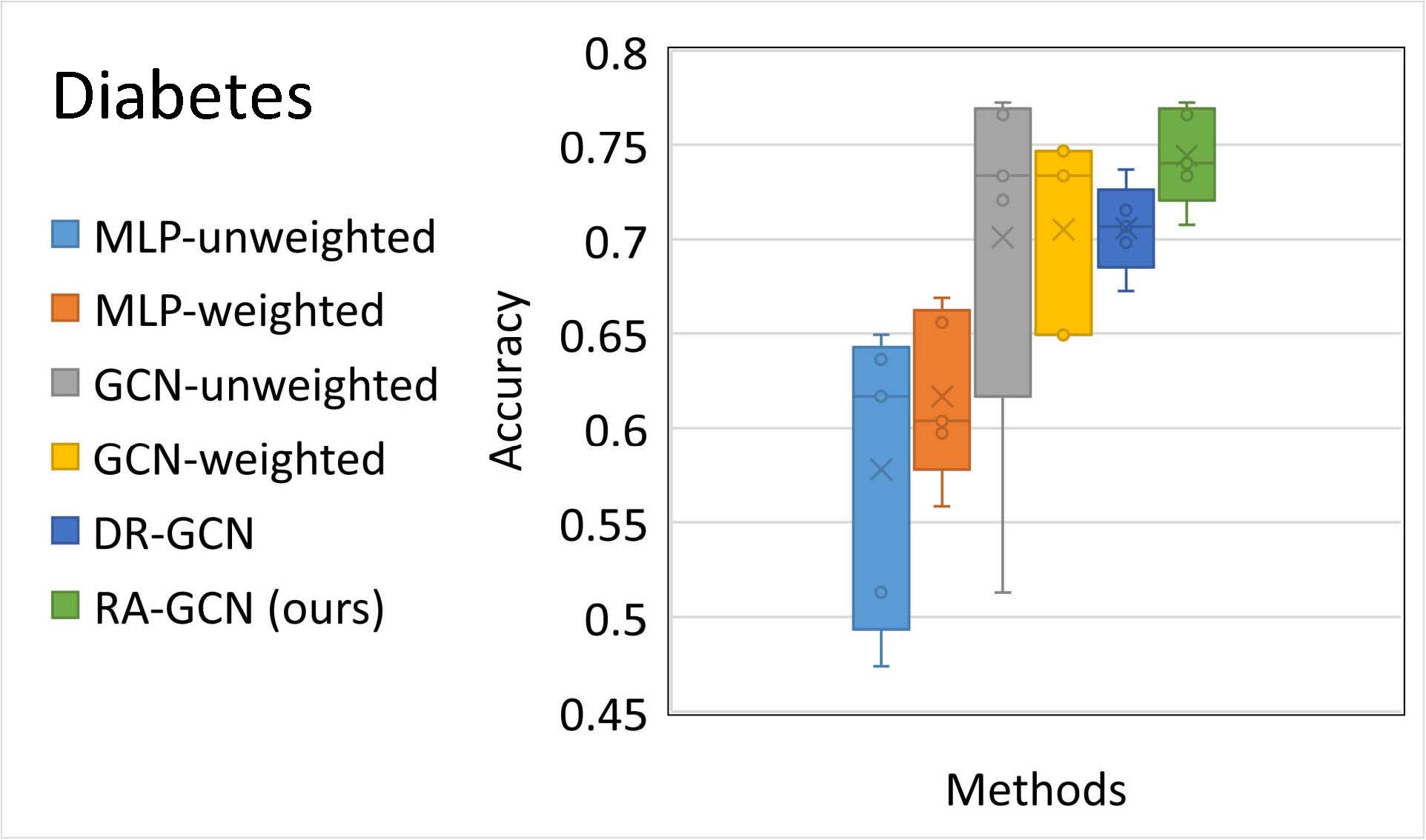}
        \hspace{2em}
       \includegraphics[scale=\myboxplotscalesmall]{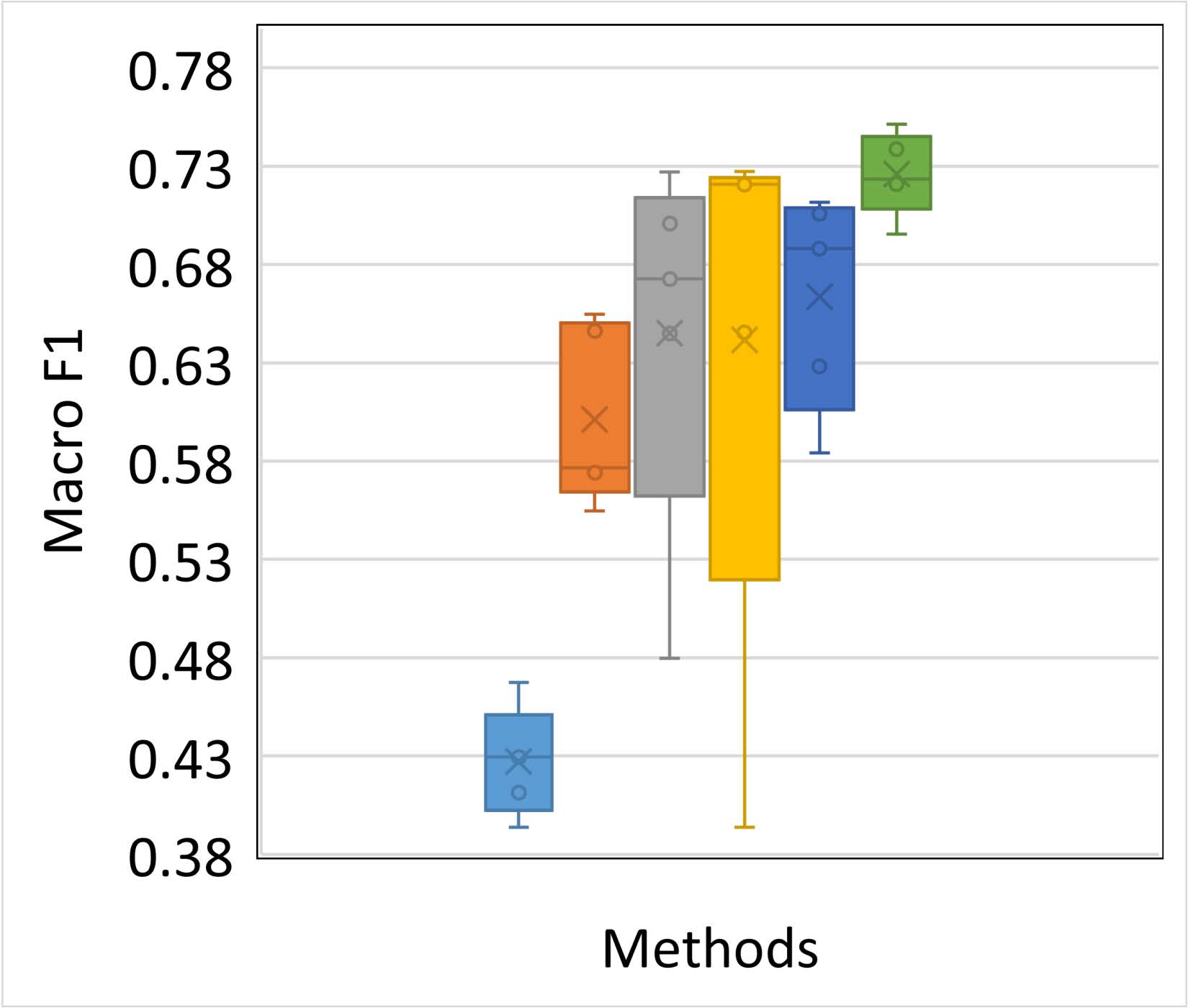}
       \hspace{2em}
       \includegraphics[scale=\myboxplotscalesmall]{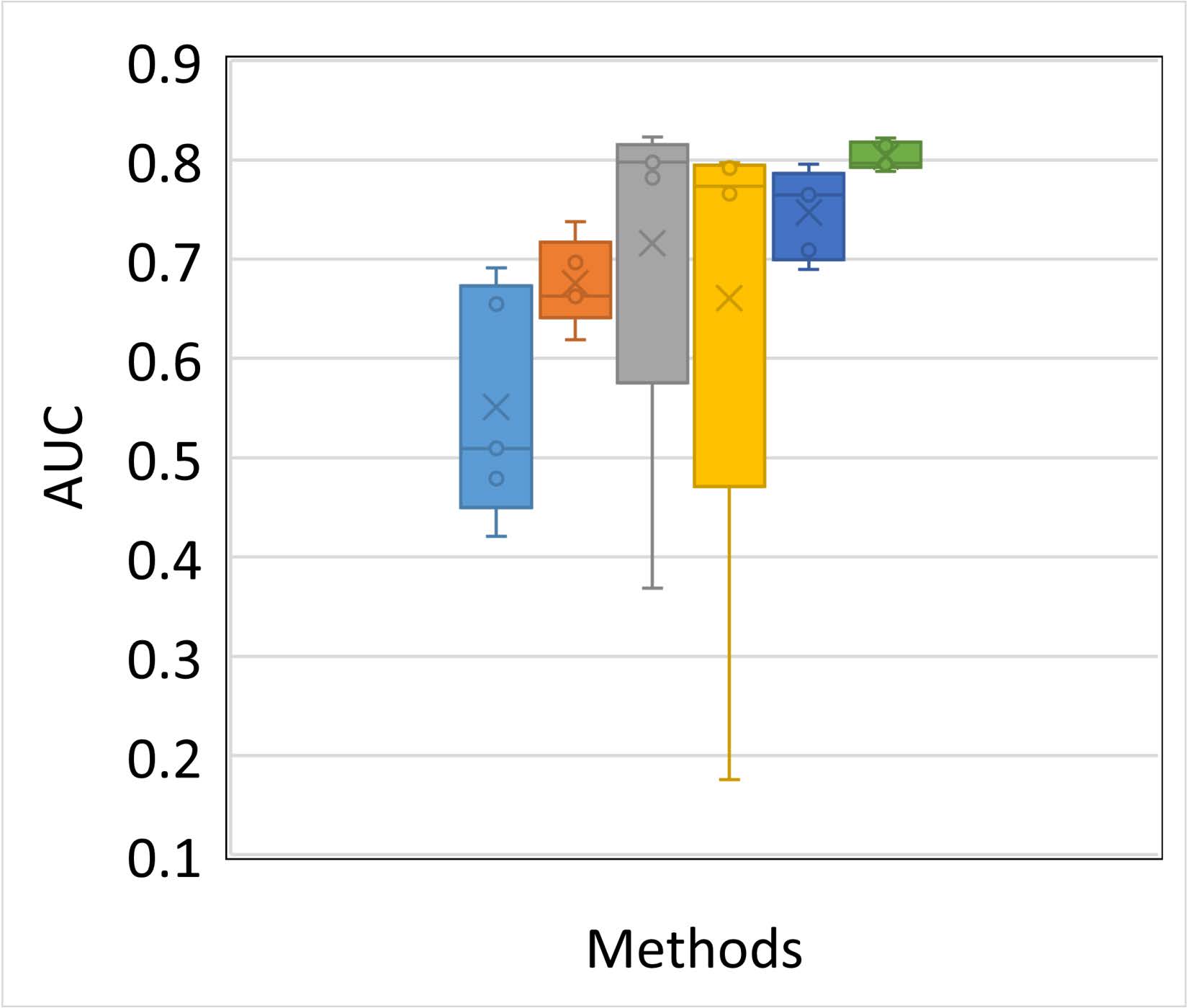}
        \caption{Box plots of Accuracy, Macro F1, and ROC AUC of the compared methods on Pima Indian Diabetes (Diabetes) dataset}
    \end{subfigure}
    \vskip\baselineskip
    \begin{subfigure}[b]{\textwidth}
        \centering
        \includegraphics[scale=\myboxplotscalebig]{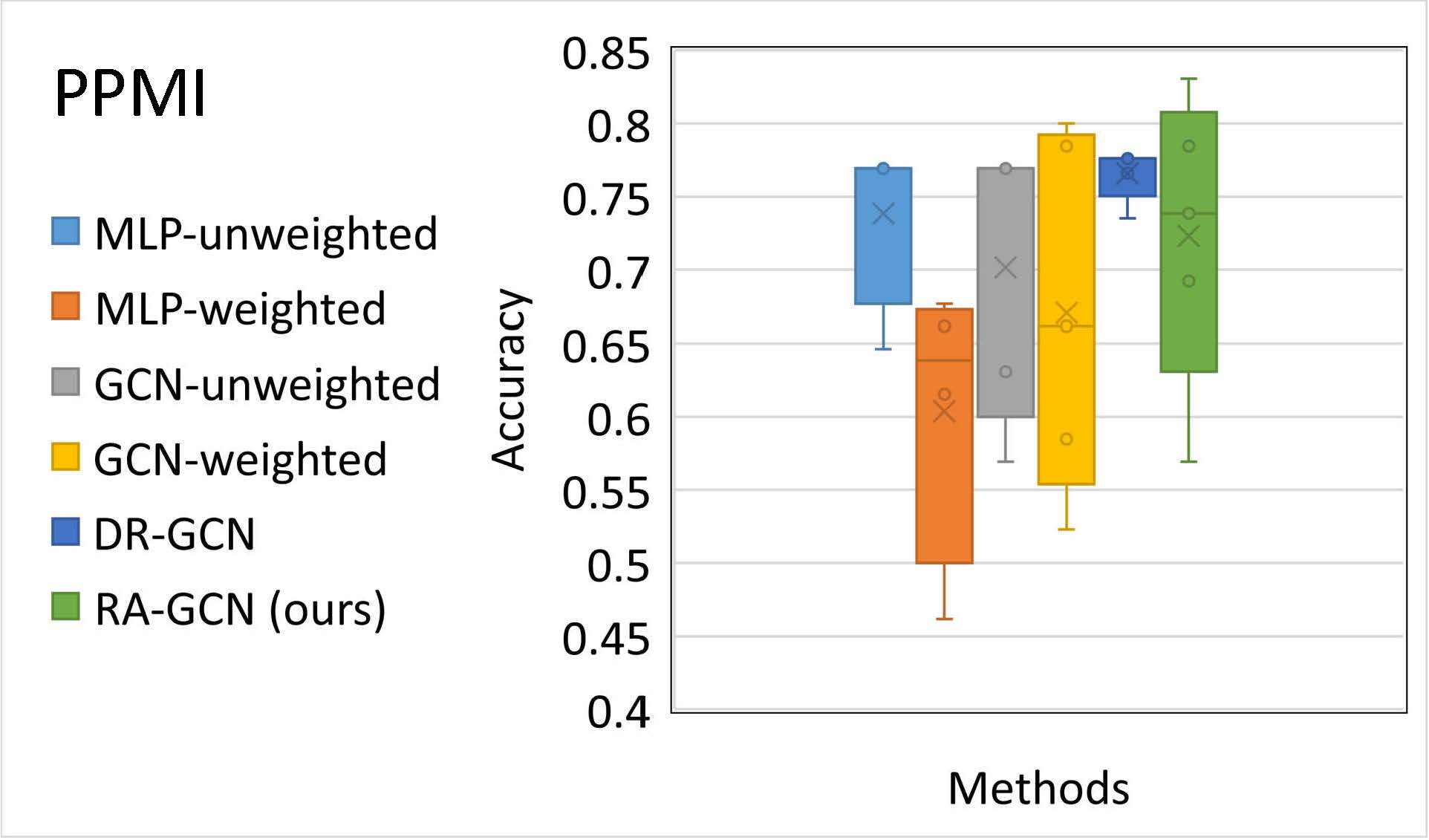}
        \hspace{2em} 
       \includegraphics[scale=\myboxplotscalesmall]{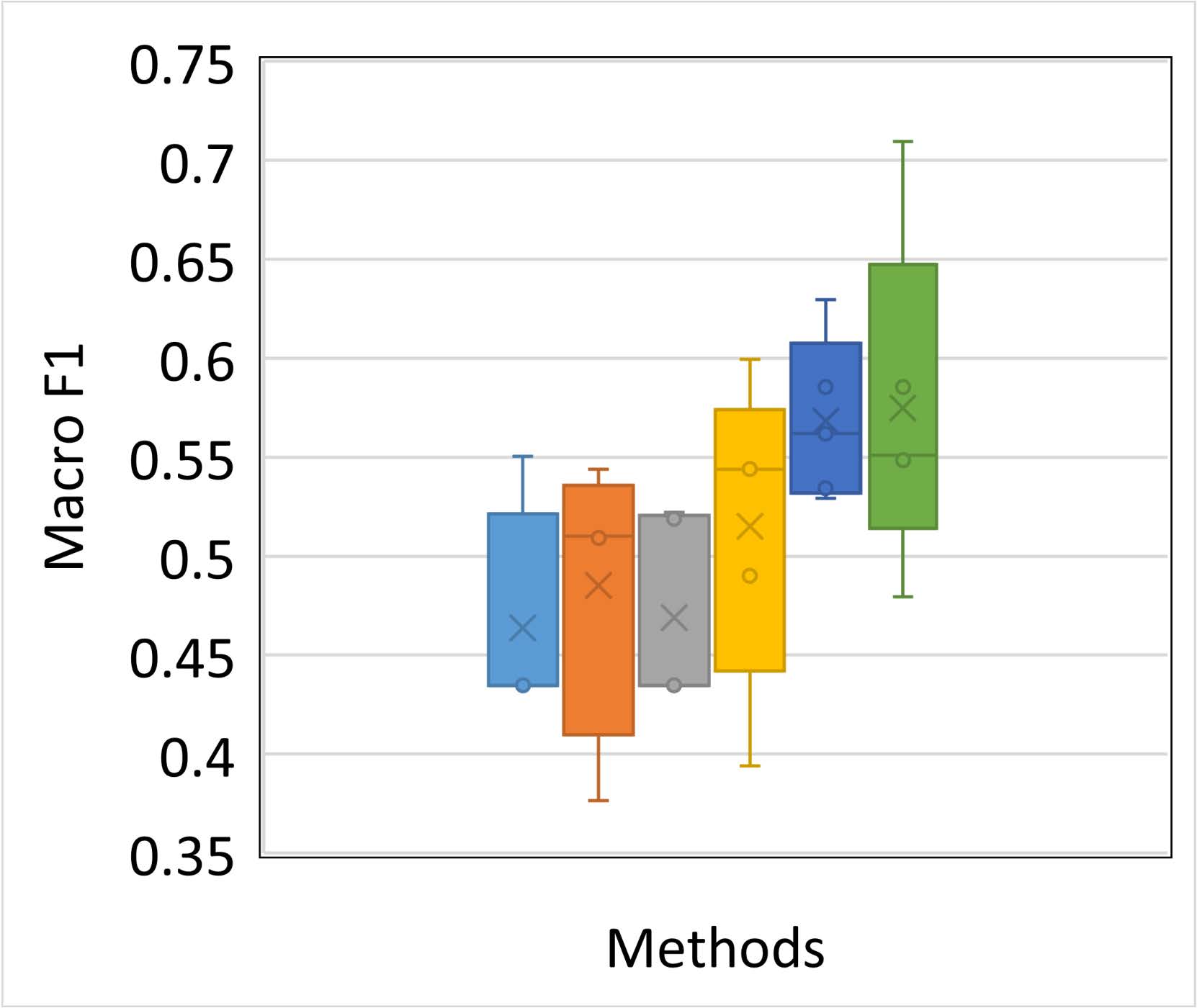}
       \hspace{2em}
       \includegraphics[scale=\myboxplotscalesmall]{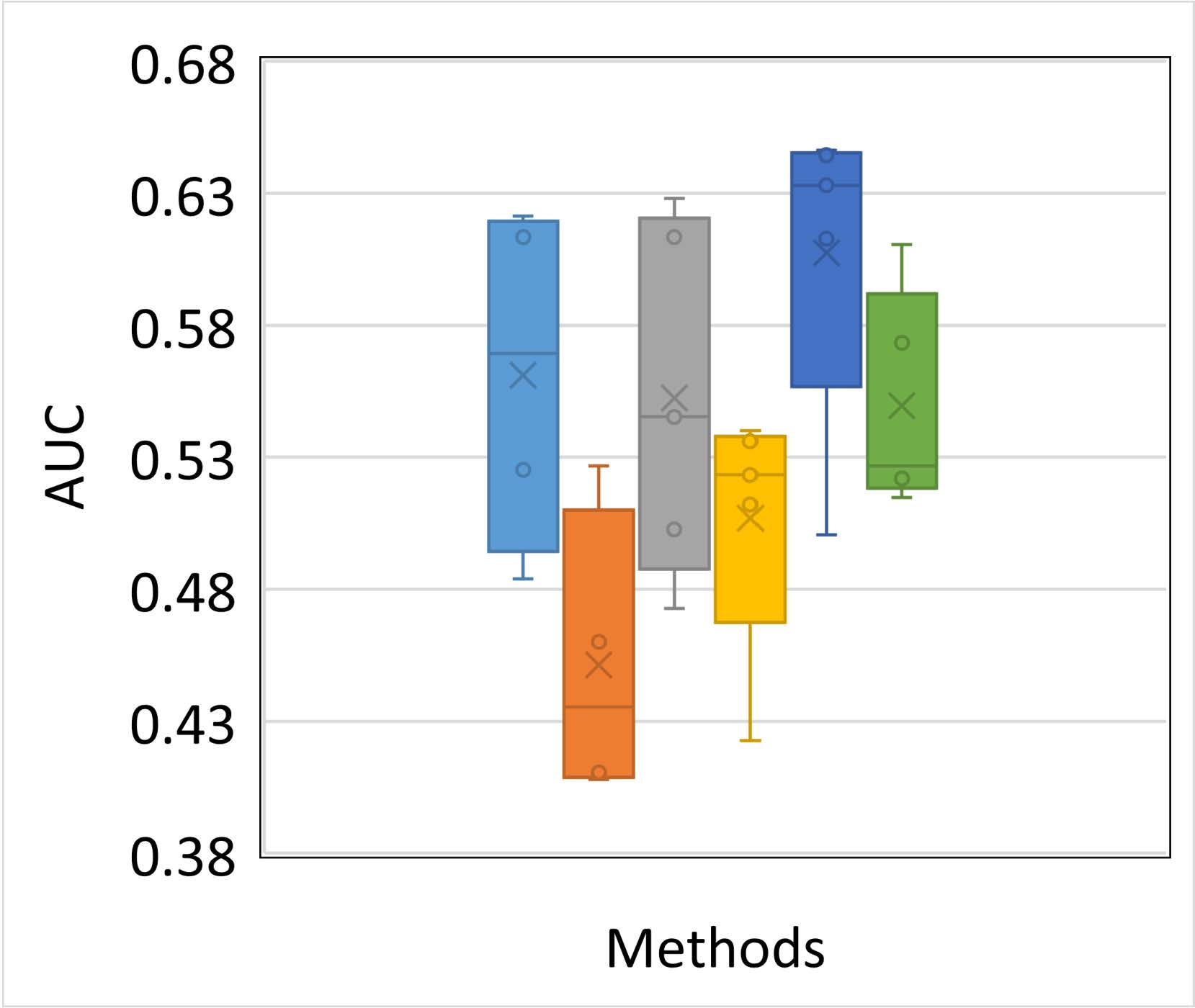}
        \caption{Box plots of Accuracy, Macro F1, and ROC AUC of the compared methods on Parkinson's Progression Markers Initiative (PPMI) dataset}
    \end{subfigure}
    \vskip\baselineskip
    \begin{subfigure}[b]{\textwidth}
        \centering
        \includegraphics[scale=\myboxplotscalebig]{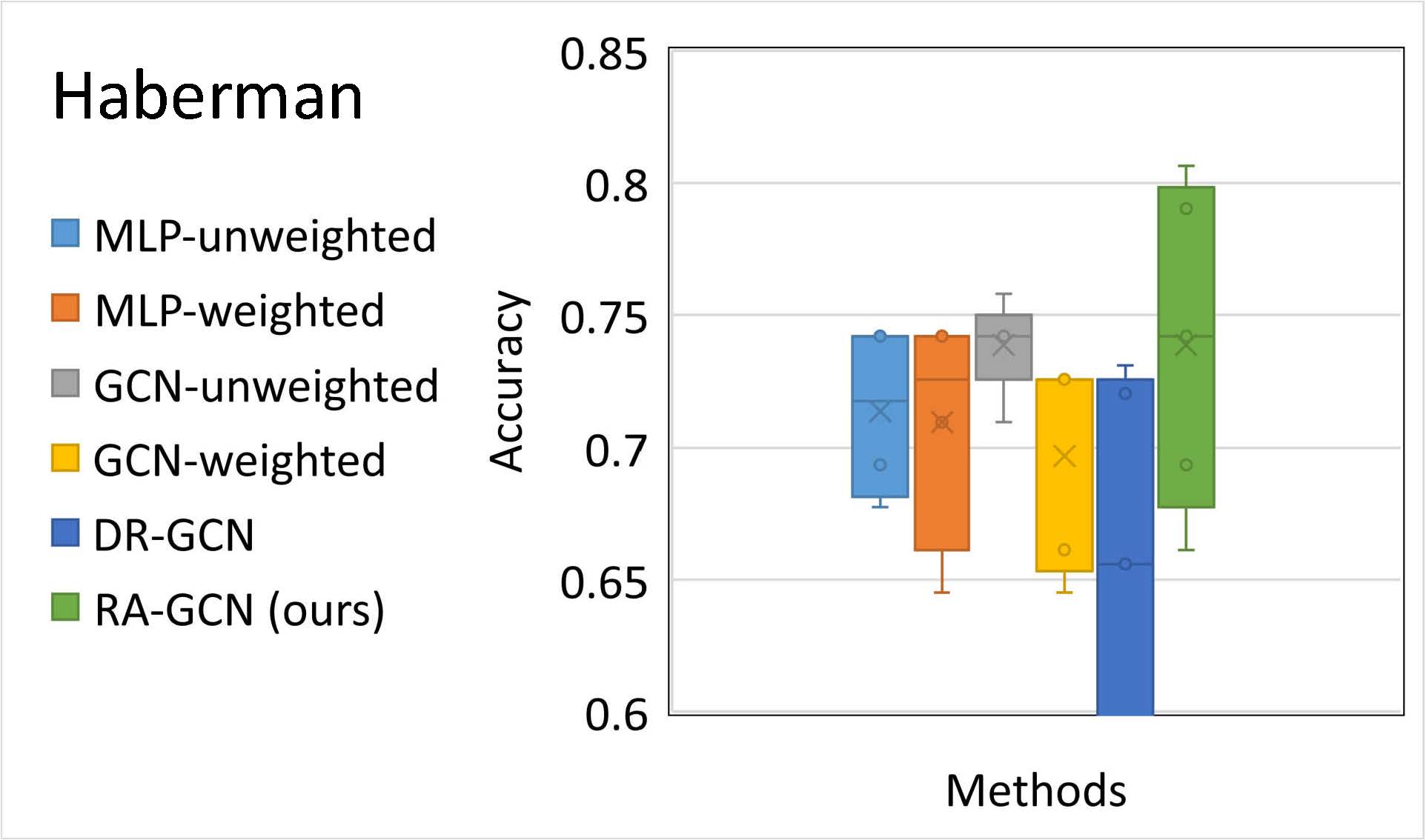}
        \hspace{2em}
       \includegraphics[scale=\myboxplotscalesmall]{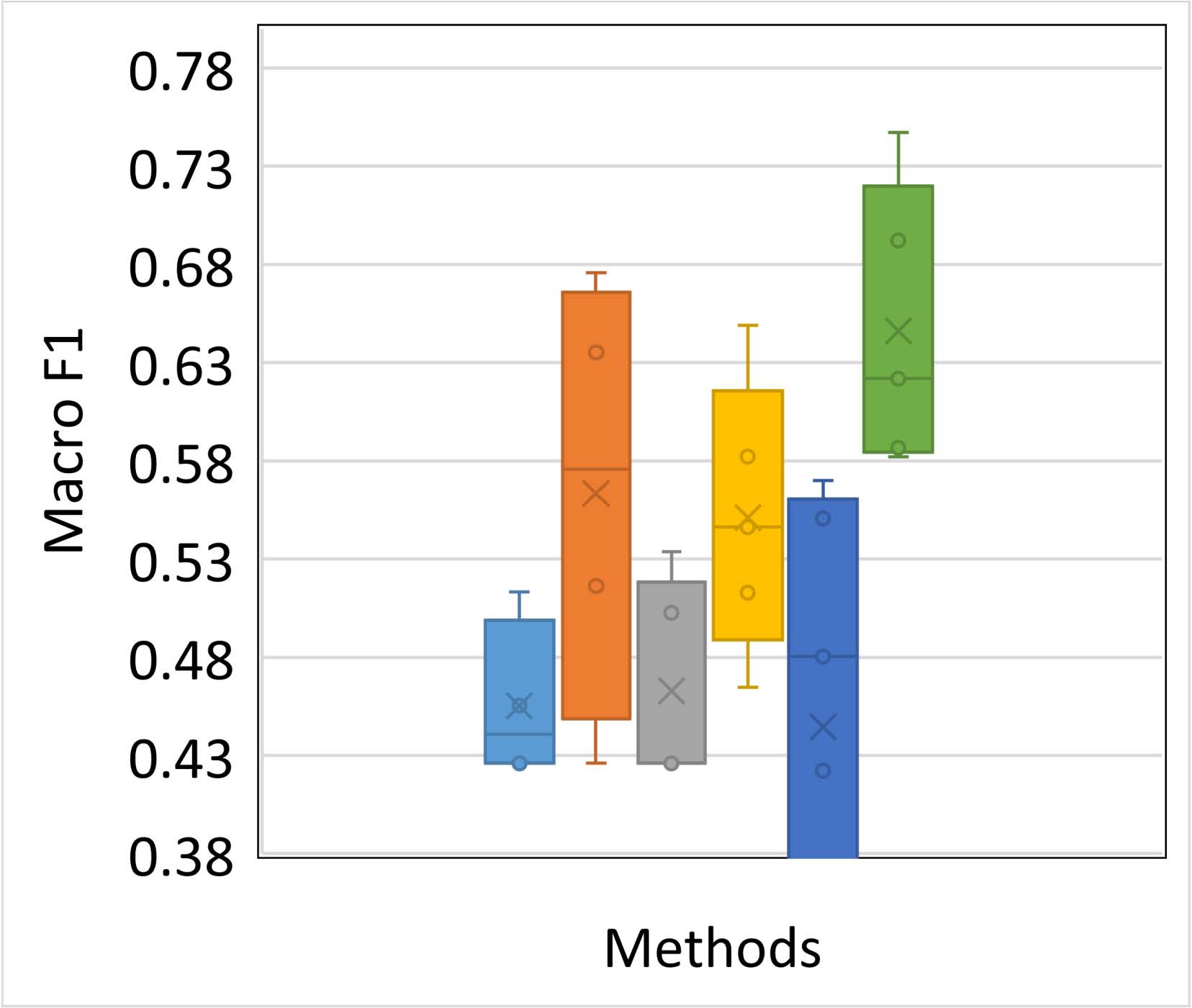}
       \hspace{2em} 
       \includegraphics[scale=\myboxplotscalesmall]{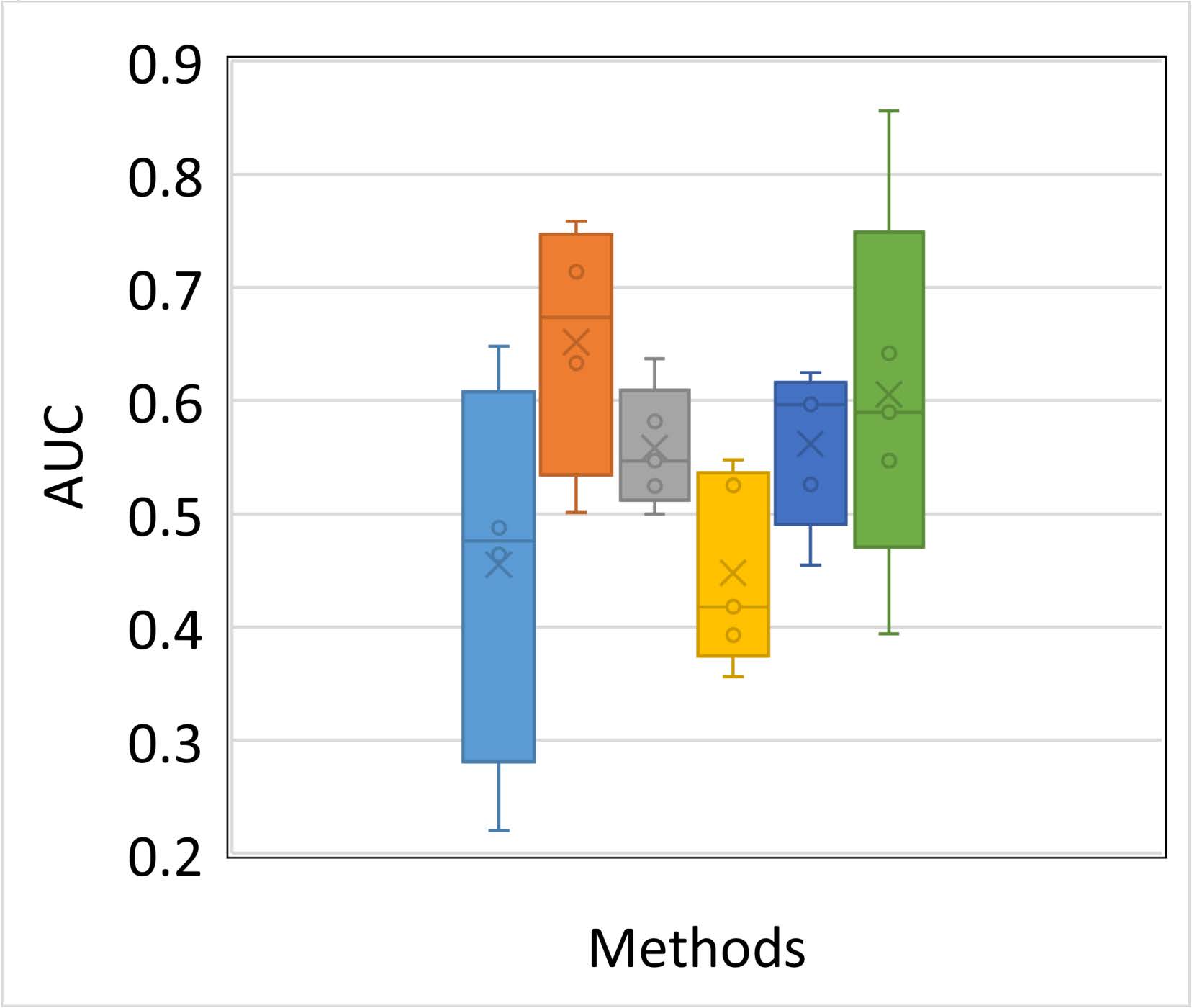}
        \caption{Box plots of Accuracy, Macro F1, and ROC AUC of the compared methods on Haberman's survival (Haberman) dataset}
    \end{subfigure}
    \caption{Box plot of results on real datasets. Each row corresponds to the results of methods on a dataset which from left to right includes accuracy, macro F1, and ROC AUC. From top to bottom, rows are results of methods on Diabetes, PPMI, and Haberman datasets. Each boxplot summarizes the values of a metric on $5$ different random splits of the dataset indicating how the values are spread out.}
    \label{fig:real-res}
\end{figure*}

Accuracy, macro F1, and ROC AUC on the real datasets are provided in Tables ~\ref{tab:real-res-diabetes}, \ref{tab:real-res-ppmi}, and \ref{tab:real-res-haberman}. The boxplots of these results are also provided in Fig.~\ref{fig:real-res} for visual comparison. 
We compare RA-GCN with MLP, GCN, their weighted versions, and one recent method (DR-GCN).

For all datasets, MLP weighted or unweighted perform worse than most of the graph-based methods.
Although GCN performs better than MLP, it has a high variance in all three datasets.
The performance of DR-GCN varies for different datasets. It shows high stability and good performance in the first two datasets, but its efficiency drops on the Haberman. 
As can be seen, our proposed method, RA-GCN, performs better than all the other methods for all datasets, especially in macro F1 that indicates the classifier's performance for both minor and major classes. 
The effect of class imbalance is evident in all the datasets. 

Diabetes dataset is the best-chosen dataset in terms of the imbalance ratio and dataset size. 
It is obvious that utilizing the graph makes a significant improvement in this dataset. 
The effect of weighting in this dataset for MLP is substantial, but its effect for GCN is not considerable; while, the problem of class imbalance still exists for both.
RA-GCN is noticeably the best and most reliable method on this dataset and has an improvement in all metrics. 

PPMI dataset is the most challenging one due to high dimensional input features, a low number of samples, and a high imbalance ratio.
Given such a ratio, all the metrics should be considered for judging a classifier's performance.
Due to the high imbalance ratio, the unweighted methods achieve higher accuracy and ROC AUC than weighted versions. However, the low value of macro F1 is a witness that the trained classifier's bias with unweighted loss function is towards the major class. 
It can be seen from Table~\ref{tab:real-res-ppmi} that the results of the DR-GCN are also interesting. Although it improves the performance of the GCN, it seems that it is still stuck in the trap of class imbalance due to the low value of macro F1. The proposed RA-GCN performs the best in improving the challenge of imbalance.

In terms of the imbalance ratio and the size of the dataset, the Haberman dataset is similar to the PPMI, but its number of input features is much lower.
By comparing the weighted version of MLP and GCN with unweighted ones in this dataset, it can be concluded that the effect of weighting is more substantial in this case.
The graph information is improving the results, but its effect is more limited than other datasets.
The results of the DR-GCN are even worse than weighted GCN. The limited number of samples and features besides the high imbalance ratio might cause poor performance since it is hard to train a generative model with these limitations. 
RA-GCN performs the best in this case, especially in improving the macro F1.

From the experiments, it can be concluded that: 1) None of the metrics is sufficient to judge a model and totally reflect the performance of the minor class against the major one.
2) Using GC-layers instead of FC-layers for feature propagation is helpful.
3) In highly imbalanced datasets, weighting definitely helps the performance, and learning weights enhances the final classifier.  

 \begin{table}[!htb]
\centering
\caption{Results of RA-GCN and compared methods on Diabetes dataset.}
\scriptsize
\begin{tabular}{|c|c|cH|c|}
\hline
Method         & Accuracy  & Macro F1 & Binary F1 & ROC AUC  \\ \hline \hline
MLP-unweighted & $0.58 \pm 0.079$ & $0.43 \pm 0.028$     & $0.15 \pm 0.136$      & $0.55 \pm 0.117$ \\ \hline
MLP-weighted   & $0.62 \pm 0.045$ & $0.60 \pm 0.046$      & $0.53 \pm 0.068$      & $0.68 \pm 0.044$ \\ \hline
GCN-unweighted & $0.70 \pm 0.107$  & $0.65 \pm 0.097$     & $0.51 \pm 0.104$      & $0.72 \pm 0.195$ \\ \hline
GCN-weighted   & $0.71 \pm 0.051$ & $0.64 \pm 0.143$     & $0.51 \pm 0.287$      & $0.66 \pm 0.272$ \\ \hline
DR-GCN   & $0.71 \pm 0.021$ & $0.66 \pm 0.049$     & $0.55 \pm 0.111$      & $0.75 \pm 0.041$ \\ \hline
RA-GCN (ours)        & $\mathbf{0.74 \pm 0.026}$ & $\mathbf{0.73 \pm 0.021}$     & $\mathbf{0.66 \pm 0.016}$      & $\mathbf{0.80 \pm 0.014}$  \\ \hline
\end{tabular}
\label{tab:real-res-diabetes}
\end{table}

\begin{table}[!htb]
\centering
\caption{Results of RA-GCN and compared methods on PPMI dataset.}
\scriptsize
\begin{tabular}{|c|c|cH|c|}
\hline
Method         & Accuracy  & Macro F1 & Binary F1 & ROC AUC  \\ \hline \hline
MLP-unweighted & $0.74 \pm 0.062$ & $0.46 \pm 0.058$     & $0.11 \pm 0.198$      & $0.56 \pm 0.067$ \\ \hline
MLP-weighted   & $0.60 \pm 0.098$ & $0.49 \pm 0.074$      & $0.24 \pm 0.074$      & $0.45 \pm 0.056$ \\ \hline
GCN-unweighted & $0.68 \pm 0.101$  & $0.48 \pm 0.050$     & $0.16 \pm 0.192$      & $0.55 \pm 0.078$ \\ \hline
GCN-weighted   & $0.64 \pm 0.119$ & $0.51 \pm 0.087$     & $0.25 \pm 0.095$      & $0.50 \pm 0.052$ \\ \hline
DR-GCN   & $\mathbf{0.76 \pm 0.016}$ & $0.56 \pm 0.037$     & $0.28 \pm 0.073$      & $\mathbf{0.60 \pm 0.055}$ \\ \hline
RA-GCN (ours)        & $0.71 \pm 0.109$ & $\mathbf{0.58 \pm 0.096}$     & $\mathbf{0.36 \pm 0.123}$      & $0.56 \pm 0.045$  \\ \hline
\end{tabular}
\label{tab:real-res-ppmi}
\end{table}

\begin{table}[!htb]
\centering
\caption{Results of RA-GCN and compared methods on Haberman dataset.}
\scriptsize
\begin{tabular}{|c|c|cH|c|}
\hline
Method         & Accuracy  & Macro F1 & Binary F1 & ROC AUC  \\ \hline \hline
MLP-unweighted & $0.71 \pm 0.033$ & $0.46 \pm 0.041$     & $0.08 \pm 0.109$      & $0.45 \pm 0.177$ \\ \hline
MLP-weighted   & $0.71 \pm 0.046$ & $0.56 \pm 0.114$      & $0.32 \pm 0.239$      & $\mathbf{0.65 \pm 0.113}$ \\ \hline
GCN-unweighted & $0.74 \pm 0.020$  & $0.47 \pm 0.055$     & $0.10 \pm 0.114$      & $0.55 \pm 0.060$ \\ \hline
GCN-weighted   & $0.71 \pm 0.032$ & $0.57 \pm 0.058$     & $0.33 \pm 0.133$      & $0.46 \pm 0.095$ \\ \hline
DR-GCN  & $0.58 \pm 0.178$ & $0.44 \pm 0.134$     & $0.26 \pm 0.137$      & $0.56 \pm 0.063$ \\ \hline
RA-GCN (ours)        & $\mathbf{0.75 \pm 0.065}$ & $\mathbf{0.65 \pm 0.081}$     & $\mathbf{0.47 \pm 0.129}$      & $0.61 \pm 0.193$  \\ \hline
\end{tabular}
\label{tab:real-res-haberman}
\end{table}

\subsection{Experiments on Synthetic Datasets} 
In this section, we investigate the effect of different factors on the proposed model's performance on synthetic datasets.
We generate the synthetic datasets to first examine the effect of imbalance ratio on the performance of the proposed model in Section~\ref{sec:exp-binary-ir}, then, in Section~\ref{sec:exp-ablation}, an ablation study is performed to evaluate how different variants of the weighting networks in the proposed model influence the results, afterward, the sensitivity of the model to its parameter $\alpha$ in the objective function is examined in Section~\ref{sec:exp-ps}. In the final experiment, the effect of graph structure on the results of all compared methods is assessed by changing the sparsity of constructed graphs in Section~\ref{sec:exp-graphsparsity}. In addition to these experiments, we provide a qualitative study to investigate the problem by visualization in Appendix~\ref{sec:exp-qualitative}. We have also explored the class imbalance issue in a more challenging situation for multi-class imbalanced data in Appendix~\ref{sec:exp-multi}. The last experiment in Appendix~\ref{sec:exp-graph-construction} is an study about the compared methods' performance when another component for graph-construction is added to their structure to learn the graph instead of graph construction in pre-processing step.

For generating synthetic samples, we use the scikit-learn library in Python \cite{scikit-learn}. The algorithm of data generation is adopted from \cite{guyon2003design} which is based on creating cluster points normally distributed around vertices on a hypercube. The algorithm introduces interdependence between the features and also is able to add various types of noise to the data.
The library also supports data generation with an imbalanced setting. In order to build a relevant graph with correlations to the output label, we generate a dataset with $2F$ features (where $F$ is a hyper-parameter) and use the first half of the features for the graph construction and the second half as the node features. The graph with node features provides us an appropriate graph-based dataset with the desired imbalance ratio.
\subsubsection{Implementation Details}
We follow the same network architectures for the RA-GCN as the real datasets as well as training settings and parameters.
The architecture of the classifier for all methods is identical. The classifier's dropout and learning rate are common hyper-parameters between all methods and are tuned for each method and dataset separately based on the validation results. 
$60\%$ of the data is used for training, $20\%$ for validation, and $20\%$ for testing. The imbalance ratio is kept the same in the splits. We chose the best classifier of each method based on its macro F1 value on the validation set and reported its results on the test set. Same as Section~\ref{sec:exp-real}, we follow the DR-GCN paper to set up its parameters.
The reported results are the mean and standard deviation of all the metrics for each method on $5$ different random splits of data.
In the following experiments, we generate $1000$ nodes with $10$ features for graph construction and $10$ features as node features. The distance between nodes in Eq.~\ref{eq:graph-dist} is cosine distance.

 \subsubsection{Effect of Imbalance Ratio} 
 \label{sec:exp-binary-ir}
In this experiment, we test the performance of RA-GCN w.r.t imbalance ratio on a set of synthetic datasets consisting of balanced, low imbalanced, and highly imbalanced data. 
In this section, we use the connectivity threshold ($\gamma$) equal to 0.5.
\\
We have two experiments that are both done on binary-class datasets. In the first experiment, the ratio of the data in the major class goes from $50\%$ to $80\%$,
and in the second one, it goes from $85\%$ to $95\%$ at a more granular level. 
The first experiment's goal is to investigate the effect of class imbalance by varying the dataset from a balanced dataset to a highly imbalanced one. In the second one, we want to study the effect of varying imbalance ratios at a higher granular level in the highly imbalanced datasets. The number of samples in all the datasets is $1000$. 
Appendix Table~\ref{tab:syn_data_lh} contains information about the imbalanced ratio and the adjacency matrix in these synthetic datasets. 

\begin{figure*}[!htb]
{
  \begin{center}
    \subfloat[Accuracy]{\label{fig:irl-acc}\includegraphics[scale=\myfigurescale]{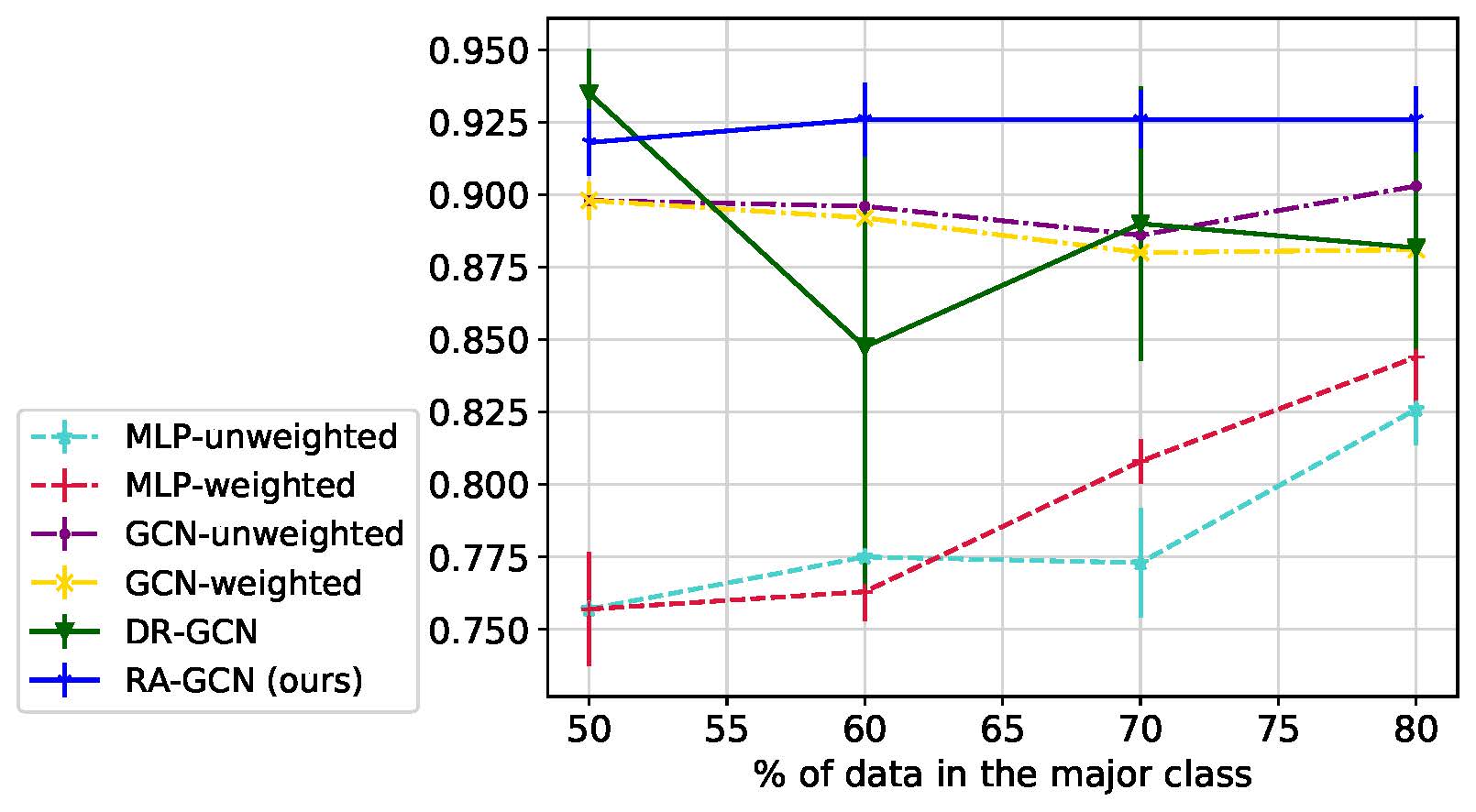}}
    \hfill
    \subfloat[Macro F1]{\label{fig:irl-f1macro}\includegraphics[scale=\myfigurescale]{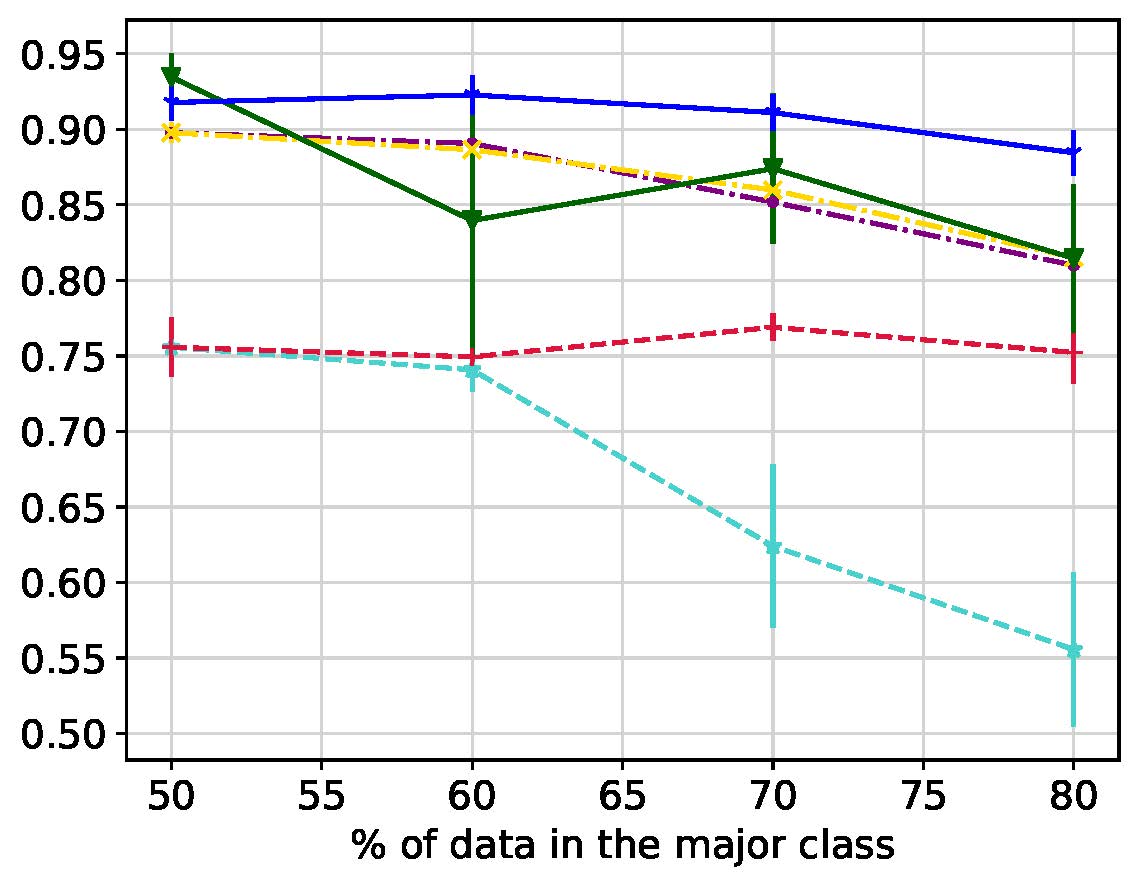}} 
    \hfill 
    \subfloat[ROC AUC]{\label{fig:irl-auc}\includegraphics[scale=\myfigurescale]{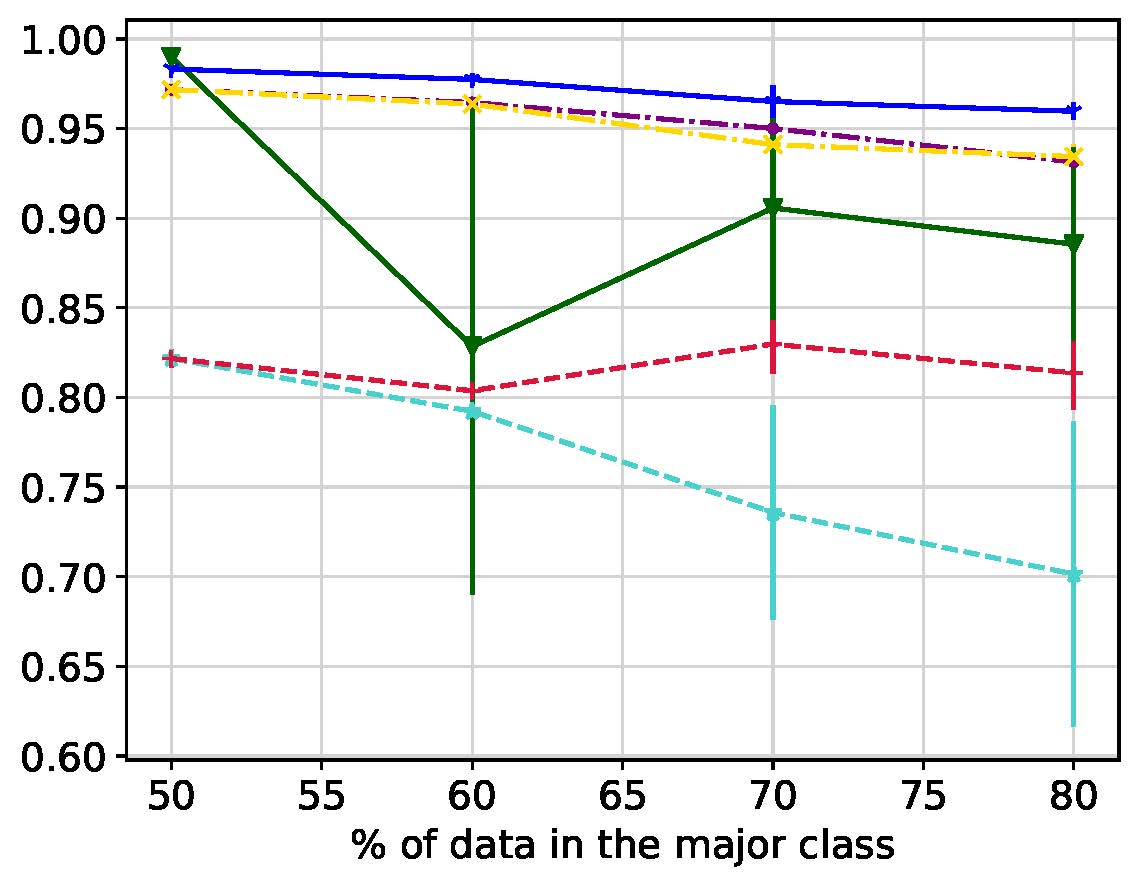}} 
  \end{center}
  \caption{The effect of changing the imbalance ratio on the performance. The figure shows the results of the compared methods. In each figure, by varying the imbalance ratio, the datasets change from balanced dataset to low imbalanced, and highly imbalanced ones.}
  \label{fig:irl}
  \vskip\baselineskip
  \begin{center}
    \subfloat[Accuracy]{\label{fig:irh-acc}\includegraphics[scale=\myfigurescale]{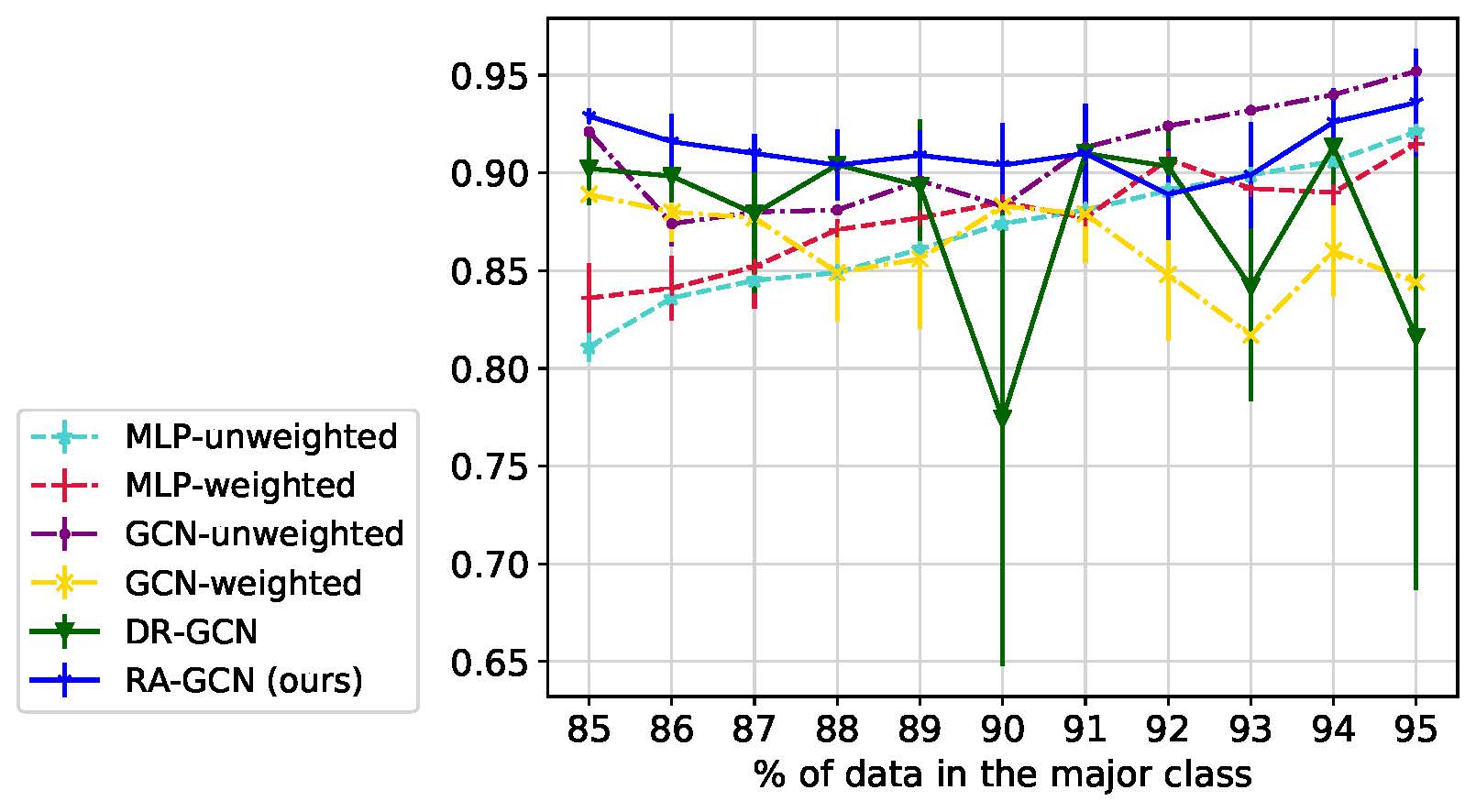}}
    \hfill
    \subfloat[Macro F1]{\label{fig:irh-f1macro}\includegraphics[scale=\myfigurescale]{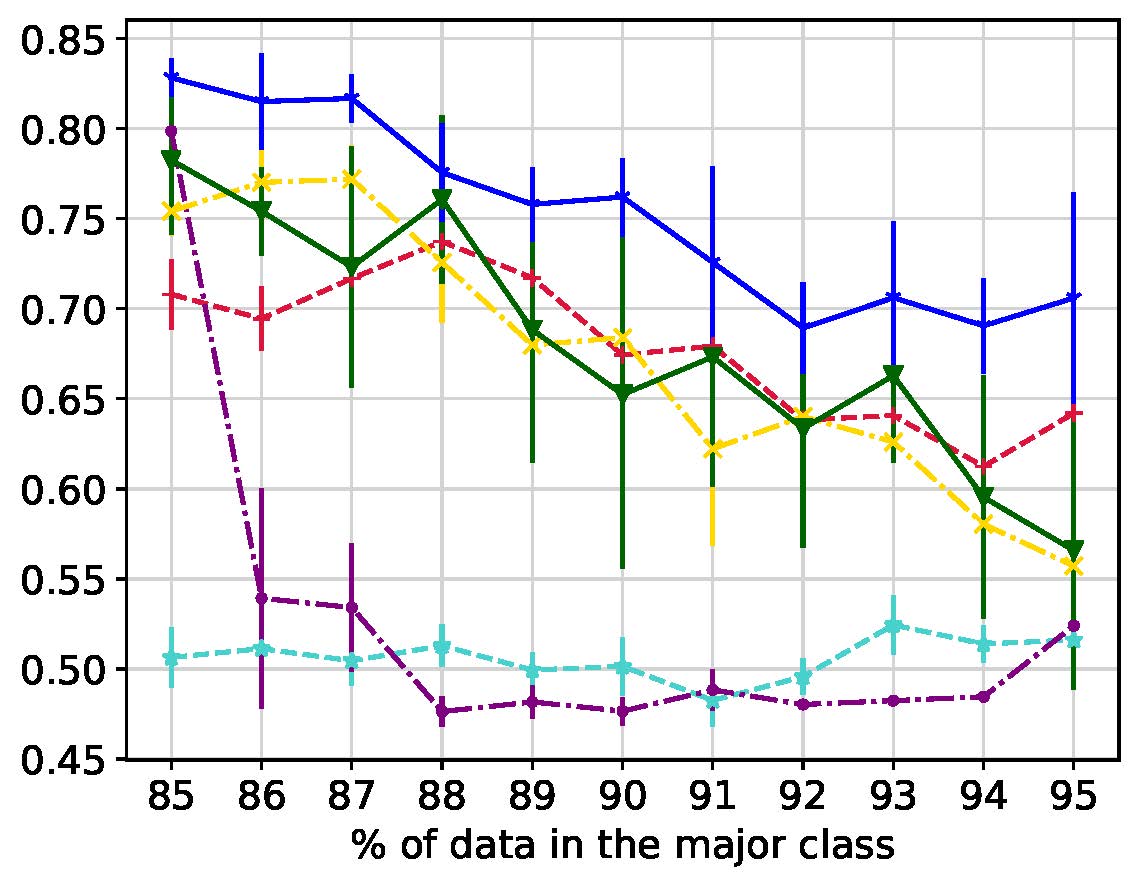}} 
    \hfill
    \subfloat[ROC AUC]{\label{fig:irh-auc}\includegraphics[scale=\myfigurescale]{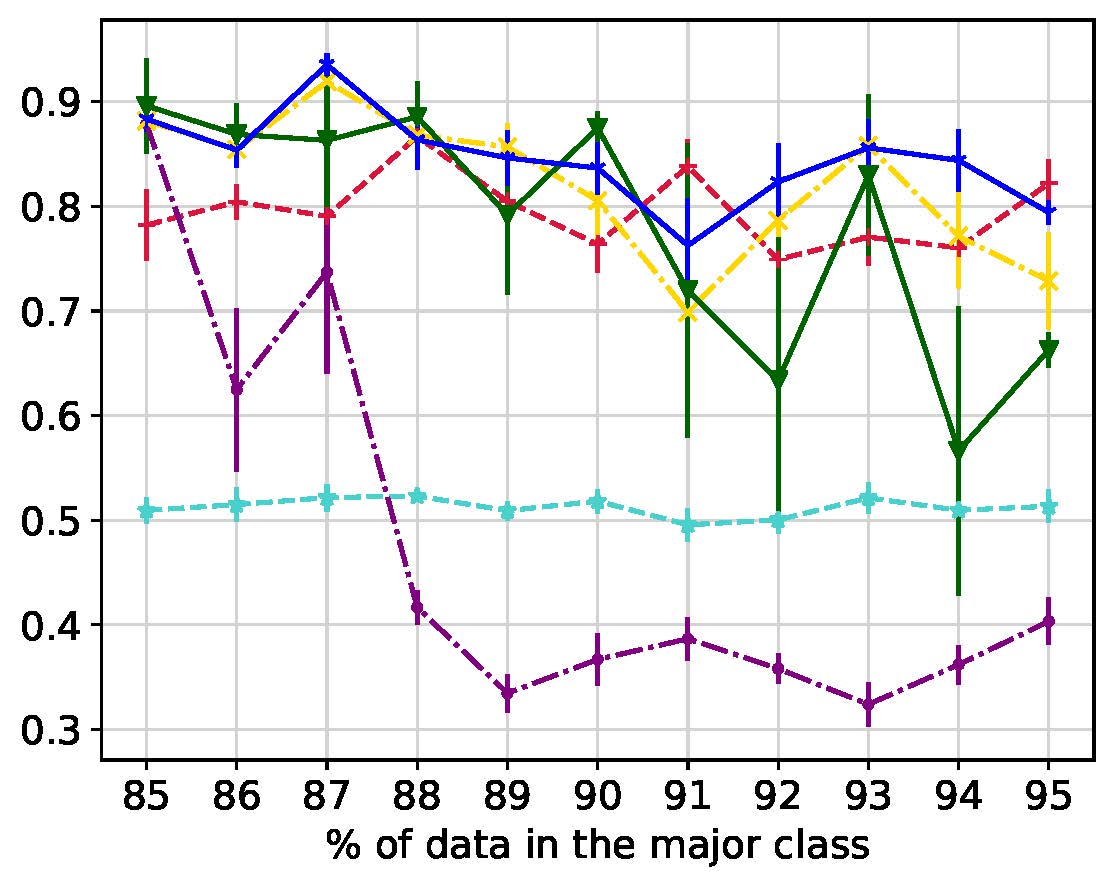}} 
  \end{center}
  \caption{The effect of changing the imbalance ratio on the performance at a granular level. The figure illustrates the results of the compared methods. All the datasets are highly imbalanced.}
  \label{fig:irh}
}
\end{figure*}

\textbf{Results}: 
The first experimental results are provided in Fig.~\ref{fig:irl}. 
When the dataset is completely balanced, there is no difference between weighted and unweighted methods. However, RA-GCN has a minor improvement, which implies that weighting (independent of the class imbalance issue) can help the neural network to find a better low-dimensional space in which classes are more separable than the other methods. 
In severe situations, the performance of all methods, including the proposed method, drops. From the results in Fig.~\ref{fig:irl}, we can conclude that employing the graph results in a better performance. By increasing the imbalance ratio, the accuracy of unweighted MLP increases, but macro F1 drops significantly. 
One interesting point is that unweighted GCN keeps its performance even when the unweighted MLP starts to drop, which shows that utilizing graphs is helpful to deal with the class imbalance in these datasets.
Weighting is an improvement for both GCN and MLP architectures, especially when the imbalance ratio increases; however, it should be noted that the improvement is minor for GCN. DR-GCN instability can also be concluded from Fig.~\ref{fig:irl}. Although this method is more complicated than merely weighting the classes, it ends up with high variance results, which might be due to the convergence problem in the GAN-based methods.
For all imbalanced datasets, the RA-GCN is better in all measures. It has a stable performance by changing the imbalance ratio.

The results of the second experiment are provided in Fig.~\ref{fig:irh}. 
These datasets are more challenging than the previous ones. Although the imbalance ratio is changing slowly, changes in the results are much dramatic. Once again, the difference between the results of weighted methods and unweighted ones proves that class imbalance can have a huge effect on the classification results. For higher imbalance ratios, unweighted methods are biased towards the major class, and although they have high accuracy, they perform poorly based on the other measures.
Interestingly, in the case of GCN with static weighting, the accuracy and ROC AUC are not high. This means that the learned classifier chooses to ignore many samples in the major class against the correct classification of a few samples in the minor one resulting in a high macro F1.
The instability of DR-GCN is more serious in this experiment. Although it improves the results of the unweighted GCN with a clear margin, its results are competitive with weighted GCN. 
Moreover, for highly imbalanced datasets, the results of weighted MLP and weighted GCN are competitive, and for high imbalance ratios, MLP has better performance, so the graph is not helpful in these cases.
We can say that in the case of GCNs, as the size of the major class grows and the minor one shrinks, the minor class samples, due to their small number, do not have many co-labeled neighbors in the graph. Therefore, they are more affected by the major class samples, which causes a drop in weighted GCN performance. This observation is also validated by Shi et al. \cite{ijcai2020-398}. 
RA-GCN deals with the problem by putting more weights on the neglected nodes and highlighting them for the classifier to focus more on the difference between classes instead of just the number of correctly classified samples. 
\textcolor{\mycolor}{
RA-GCN shows a much higher macro F1 and a competitive ROC AUC. In terms of accuracy, it beats the other ones when the percentage of the majority class varies from $85\%$ to $90\%$. The results demonstrate the superiority and robustness of RA-GCN in severely imbalanced datasets.}

\textcolor{\mycolor}{\subsubsection{Effect of Modification in The Weighting Networks}
\label{sec:exp-ablation}
In this section, we investigate the model in detail for a better understanding of weighting networks' role in the proposed model.
For this purpose, two synthetic datasets containing $1000$ nodes with a moderate imbalance ratio ($75\%$ of data in the major class) and a high imbalance ratio ($95\%$ of data in the major class) are generated. The threshold for graph construction is $0.5$ to follow the same steps as the previous experiment. It should be noted the results are provided with a similar structure for the weighting networks in all the compared variants. The compared variations of the RA-GCN are as follows. \\
\textbf{No weighting networks}: In this case, the weighting network is discarded from model and only the classifier is trained. \\
\textbf{Class-weighting networks}: This variant is the class-weighted version of the proposed RA-GCN. It has the same structure as RA-GCN, but instead of the softmax function, we first calculate the output of weighting networks for all samples, then apply an average function on the samples of each class to get class weights. In the last step, we normalize the class weights and use them in the WCE loss function. \\
\textbf{One sample-weighting network for all samples}: In this variant of RA-GCN, instead of having a separate weighting network for each class, one sample-weighting network with the same structure is supposed to learn the weights for all training samples. \\
\textbf{ Separate sample-weighting network per class with shared parameters}: In this form of model, like the original RA-GCN, one weighting network is responsible for the weights of samples in each class, but all the network parameters are shared between W-GCNs except the parameters of the last layer. 
}

\textbf{Results}: Table~\ref{tab:ablation} shows the results of different models on both moderate ($75\%$ of data in the major class) and highly imbalanced ($95\%$ of data in the major class) datasets. 
In both datasets, the imbalance issue is apparent in the results of the classifier without weighting networks. 
The results are competitive on the moderate imbalanced dataset (right-side of Table~\ref{tab:ablation}) and the main difference between the models shows up in the highly imbalanced setting.
By comparing the method for class-weighting (the second row) against sample-weighting methods, it can be concluded that the sample-weighting is superior approach over the class-weighting in terms of accuracy, macro F1 and ROC AUC, especially for highly imbalanced datasets.  \\
From the results we can infer that finding appropriate weights for all samples from different classes with one weighting network is a difficult task and needs a network with more parameters that might overfit due to the few number of samples.
On the other hand, although sharing the parameters between the networks in the moderate imbalanced dataset has better results than one weighting network for all samples, it is more unstable with lower performance in a high imbalance setting. When the parameters are shared between the weighting networks, they use the same low-dimensional representation (hidden space of the network) to fit a weight distribution over classes. This can be a limitation when the best weight distributions over classes need more than one layer (the last one layer which is not shared between classes) to be learned. 
Therefore, dedicating a separate weighting network to each class makes the problem simpler for each network and increases the ability of the model to learn complicated weight distributions.

\begin{table*}[!htb]
\centering
\caption{Ablation study for RA-GCN. The table shows the results of different variants of RA-GCN with respect to the design of weighting networks and the effect of each variation on the performance.}
\notsotiny
\color{\mycolor}
\begin{tabular}{|c|c|cH|c||c|cH|c|}
\hline 
\begin{tabular}[c]{@{}c@{}}Class distribution \end{tabular} 
& \multicolumn{4}{c||}{[0.95-0.05]} & \multicolumn{4}{c|}{[0.75-0.25]}
\\ \hhline{|-|-|-|-|-||-|-|-|-|}
Method/Metric & Accuracy & Macro F1 & Binary F1  & ROC AUC & Accuracy & Macro F1 & Binary F1  & ROC AUC 
\\ \hline 
\begin{tabular}[c]{@{}c@{}}No weighting networks \\ (GCN-unweighted)\end{tabular}  
& $\mathbf{0.952 \pm 0.0012}$ & $0.524 \pm 0.0226$ & $0.073 \pm 0.0445$ & $0.403 \pm 0.0225$ 
& $0.914 \pm 0.0037$ & $0.872 \pm 0.0055$ & $0.799 \pm 0.0087$ & $0.964 \pm 0.0025$
\\ \hline 
\begin{tabular}[c]{@{}c@{}}Class-weighting networks\end{tabular}  
 & $0.874 \pm 0.0954$ & $0.616 \pm 0.0903$ & $0.304 \pm 0.1283$ & $0.74 \pm 0.0713$ 
 & $0.913 \pm 0.0227$ & $0.884 \pm 0.0325$ & $0.825 \pm 0.0507$ & $0.943 \pm 0.0326$ 
 \\ \hline 
\begin{tabular}[c]{@{}c@{}}One sample-weighting network \\ for all samples\end{tabular}
& $0.918 \pm 0.0273$  & $0.698 \pm  0.0731$  &  $ 0.440 \pm 0.1352$ & $0.864 \pm 0.0665$
& $0.913 \pm 0.0284$  & $0.887 \pm 0.0375 $  &  $0.834 \pm 0.0564 $ & $0.956 \pm 0.0266$ 
\\ \hline 
\begin{tabular}[c]{@{}c@{}} Separate sample-weighting network \\ per class with shared parameters\end{tabular}
&  $ 0.933 \pm 0.0301 $  & $0.695 \pm 0.1380$ & $0.426 \pm 0.2605$  & $0.783 \pm 0.2169 $
&  $0.921 \pm 0.0201 $  & $0.895 \pm 0.0299$ & $0.844 \pm 0.0473$  & $0.956 \pm 0.0298$ 
\\ \hline 
\begin{tabular}[c]{@{}c@{}}Separate sample-weighting network \\ per class (proposed model)\end{tabular}   
& $0.934 \pm 0.0102$  & $\mathbf{0.710 \pm 0.0366} $ &  $\mathbf{0.456 \pm 0.0698}$ & $\mathbf{0.871 \pm 0.0451}$ 
& $\mathbf{0.927 \pm 0.0163}$  & $\mathbf{0.905 \pm 0.0224}$ &  $\mathbf{0.859 \pm 0.0342}$ & $\mathbf{0.960 \pm 0.0188}$
\\ \hline
\end{tabular}
\label{tab:ablation}
\end{table*}

\textcolor{\mycolor}{
\subsubsection{Parameter Sensitivity}
\label{sec:exp-ps}
RA-GCN algorithm involves the parameter $\alpha$ as a coefficient of a regularization term added to the objective function. 
In order to evaluate how changes to the parameter $\alpha$ affect the performance on node classification, we conduct the experiments on two synthetic datasets. Likewise the previous experiment, we use two datasets of $1000$ nodes with $95\%$ and $75\%$ of nodes in the major class. For this experiment, we fixed the structure of the classifier and weighting networks and changed the coefficient of entropy term in Eq.~\ref{eq:ra-gcn}.}

\begin{figure*}[!htb]
{
  \begin{center}
    \subfloat[Accuracy]{\label{fig:ps-acc}\includegraphics[scale=\myfigurescale]{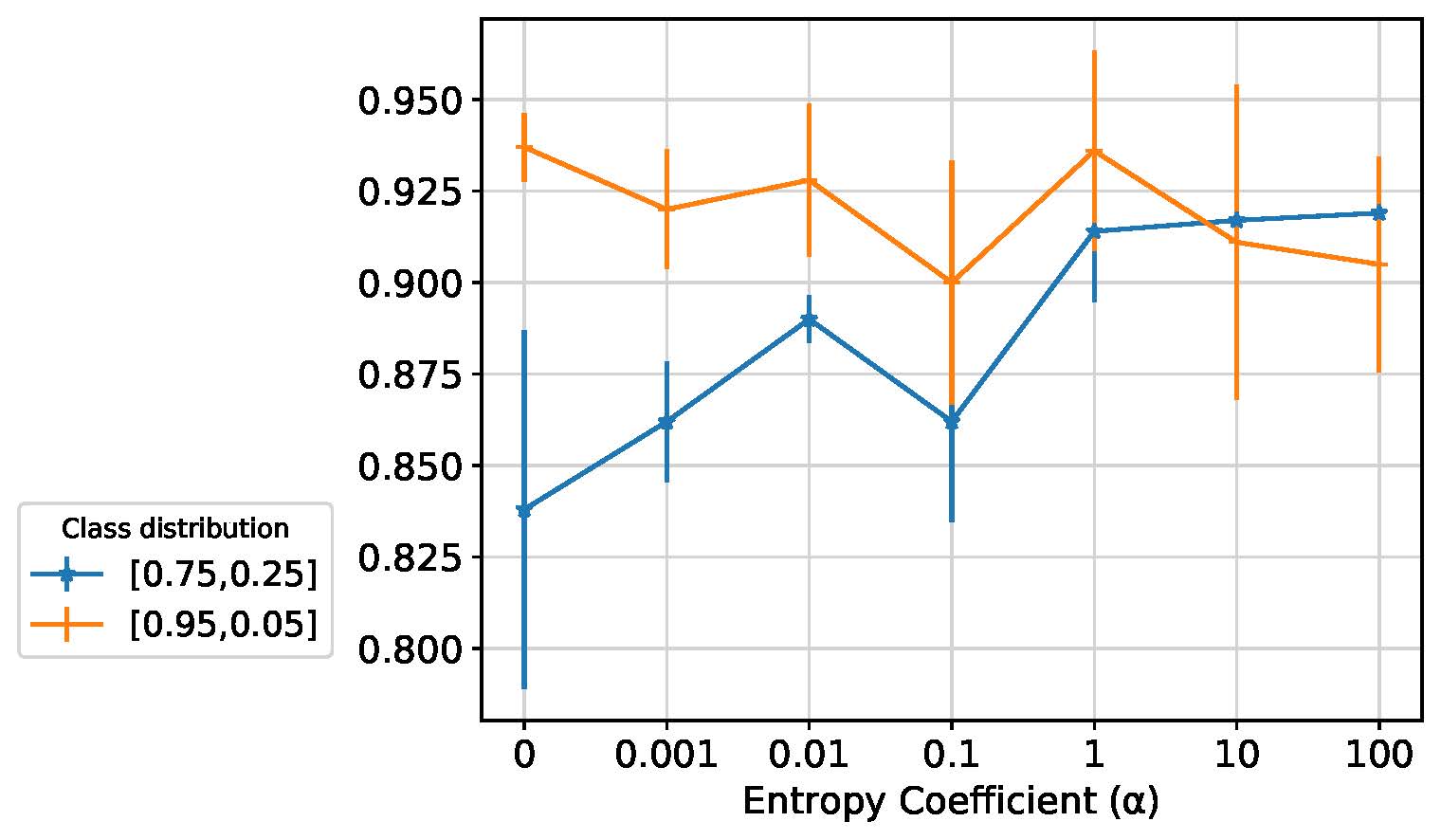}}
    \subfloat[Macro F1]{\label{fig:ps-f1macro}\includegraphics[scale=\myfigurescale]{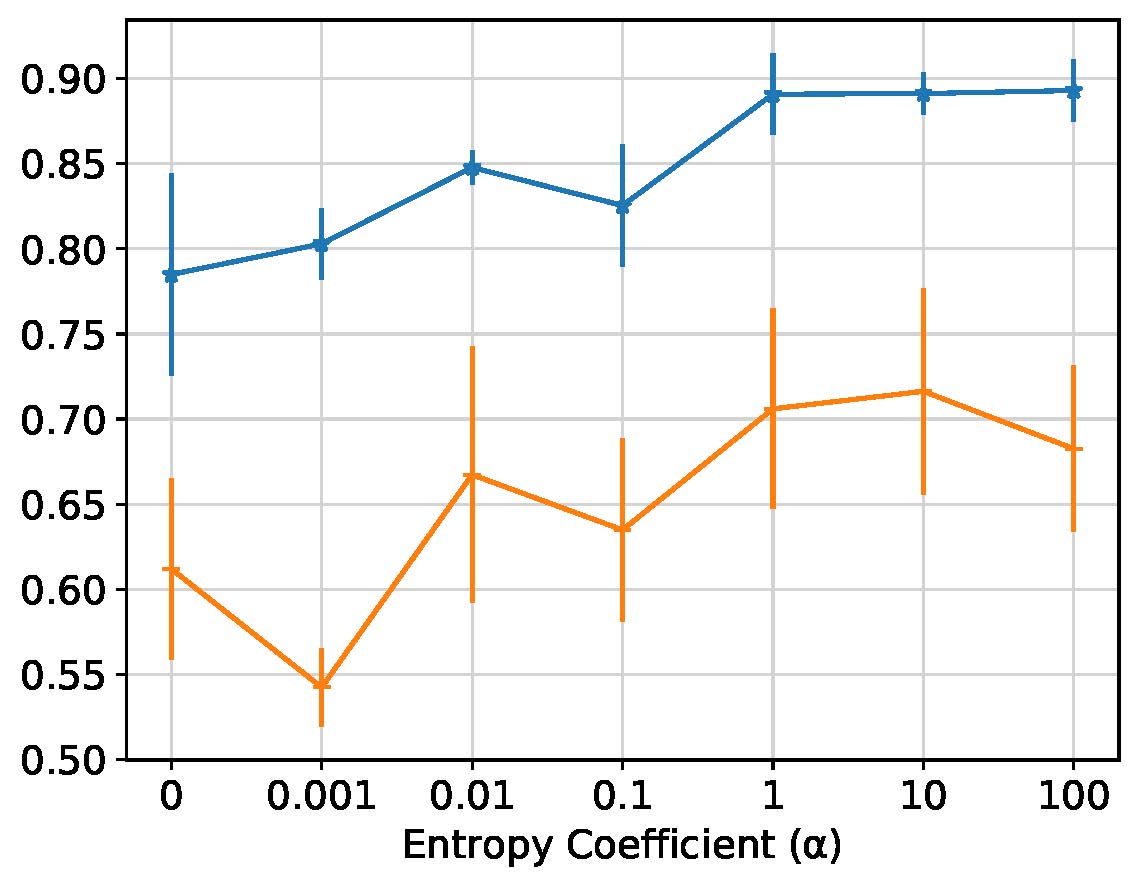}} 
    \subfloat[ROC AUC]{\label{fig:ps-auc}\includegraphics[scale=\myfigurescale]{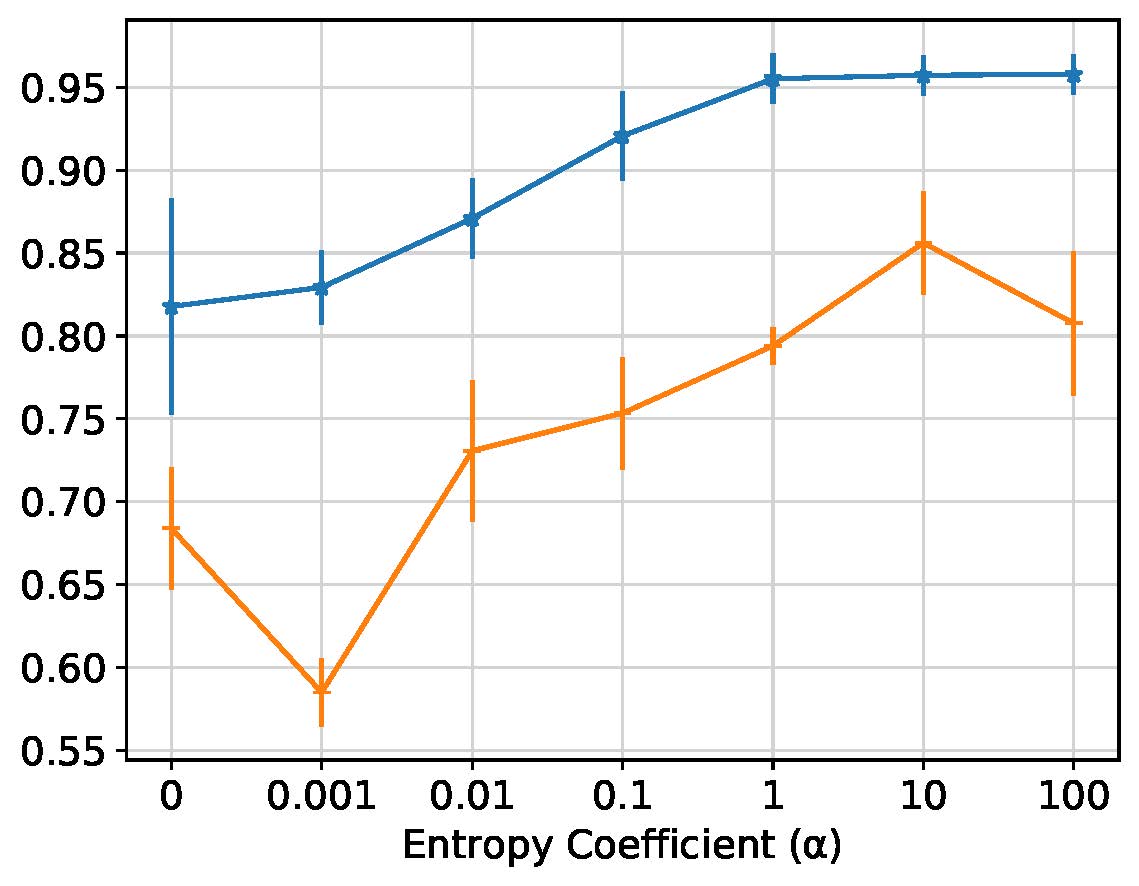}} 
  \end{center}
  \caption{\textcolor{\mycolor}{Performance evaluation of RA-GCN on varying amount of $\alpha$. The influence of the entropy term in the objective function is controlled by $\alpha$. Adjusting $\alpha$ is more effective when the imbalance ratio increases.}}
  \label{fig:ps}
}
\end{figure*}

\textbf{Results}: Fig.~\ref{fig:ps} shows the effect of increasing the coefficient of entropy term on the results.
In the moderate imbalanced dataset (the blue line), the performance of RA-GCN across different values of $\alpha$ is more stable than the other dataset.
From this observation, we can conclude that a high entropy weight distribution can also work out for this type of dataset.
This confirms the results from the previous experiment saying that learning the class-weights instead of sample-weights has more acceptable results in the moderate imbalanced dataset than the highly imbalanced one.
Moreover, when we discard the entropy term from the objective function ($\alpha=0$), the results on the moderate imbalanced data get unstable with higher variance than for the other values of $\alpha$. 
On the other hand, in the highly imbalanced dataset (orange line), the coefficient of entropy is more effective. In this dataset, when the $\alpha$ is zero or close to zero ($\alpha=0.001$), the performance drops, because, with a small number of samples in the minor class, the weighting network is more likely to assign high weights to a few number of samples.  
The experimental results demonstrate that the entropy term is more effective for the learning process of sample-weighting when the class imbalance issue gets more severe.

 \subsubsection{Effect of Graph Sparsity}
 \label{sec:exp-graphsparsity}
 Although this paper's primary focus is to deal with the challenge of class imbalance by learning the weights automatically, we cannot avoid the fact that GCNs are sensitive to the graph structure \cite{kazi2019inceptiongcn, rossi2020sign}.
 In this section, we study the effect of the graph construction on the imbalanced classification problem. For this purpose, we generate two sets of datasets. In each set, the imbalance ratio is fixed. Following the graph construction from Section~\ref{sec:graph_cons}, we vary the threshold $\gamma$ for graph construction from $0.1$ to $0.9$. When the threshold is $0.1$, the graph is sparse (refer to Eq.\ref{eq:graph-dist}), and when the threshold is $0.9$, the graph is becoming denser, and even samples from different classes are connected. All the datasets contain $1000$ samples. In the first and second sets of the generated datasets, the percentage of data in the major class is $95$ and $75$, respectively. 
 The information about the sparsity of graphs is provided in Table~\ref{tab:graph-thr-info}.

\begin{figure*}[!htb]
{
  \begin{center}
    \subfloat[Accuracy]{\label{fig:ir-acc}\includegraphics[scale=\myfigurescale]{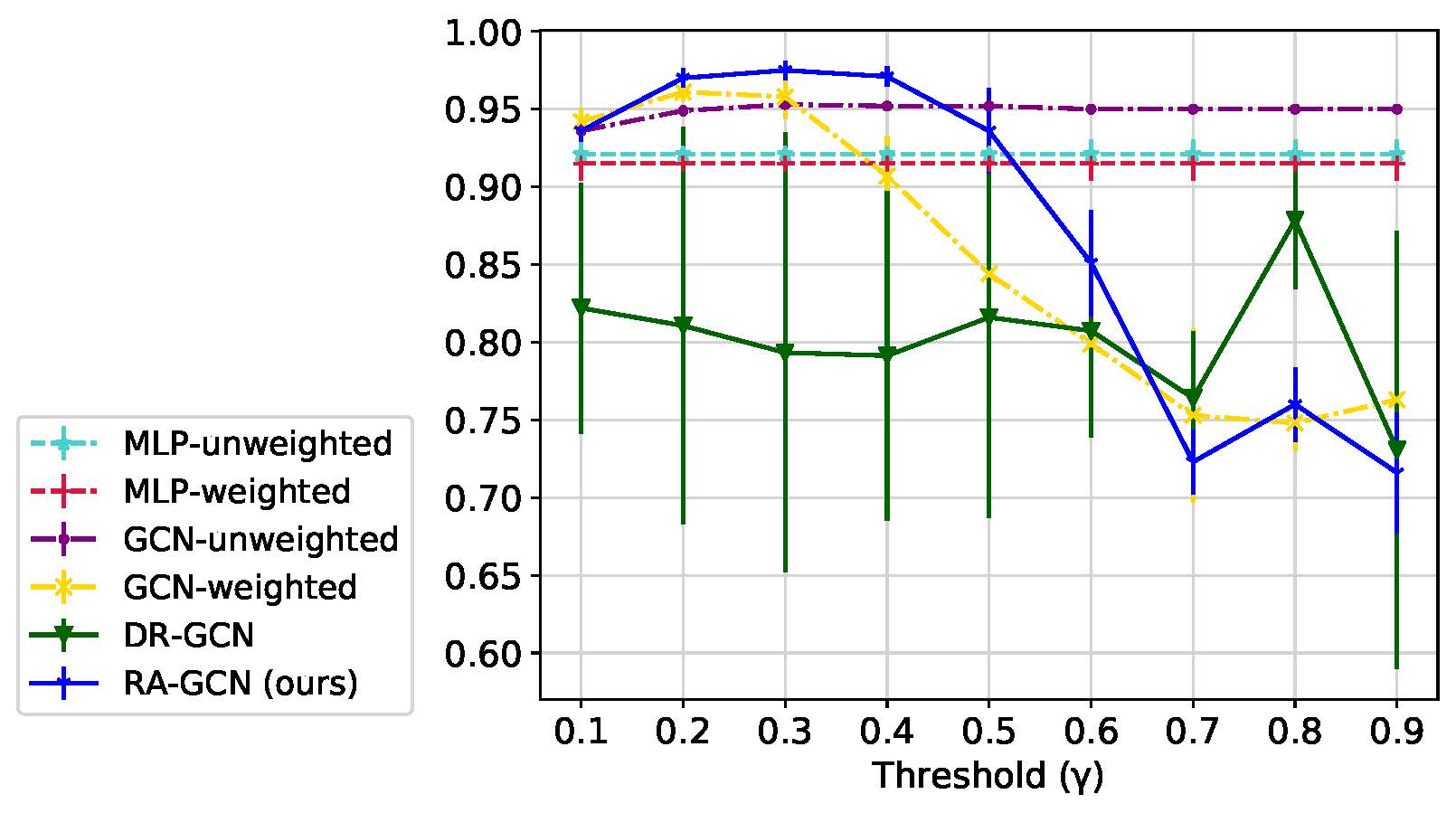}}
    \hfill
    \subfloat[Macro F1]{\label{fig:ir-f1macro}\includegraphics[scale=\myfigurescale]{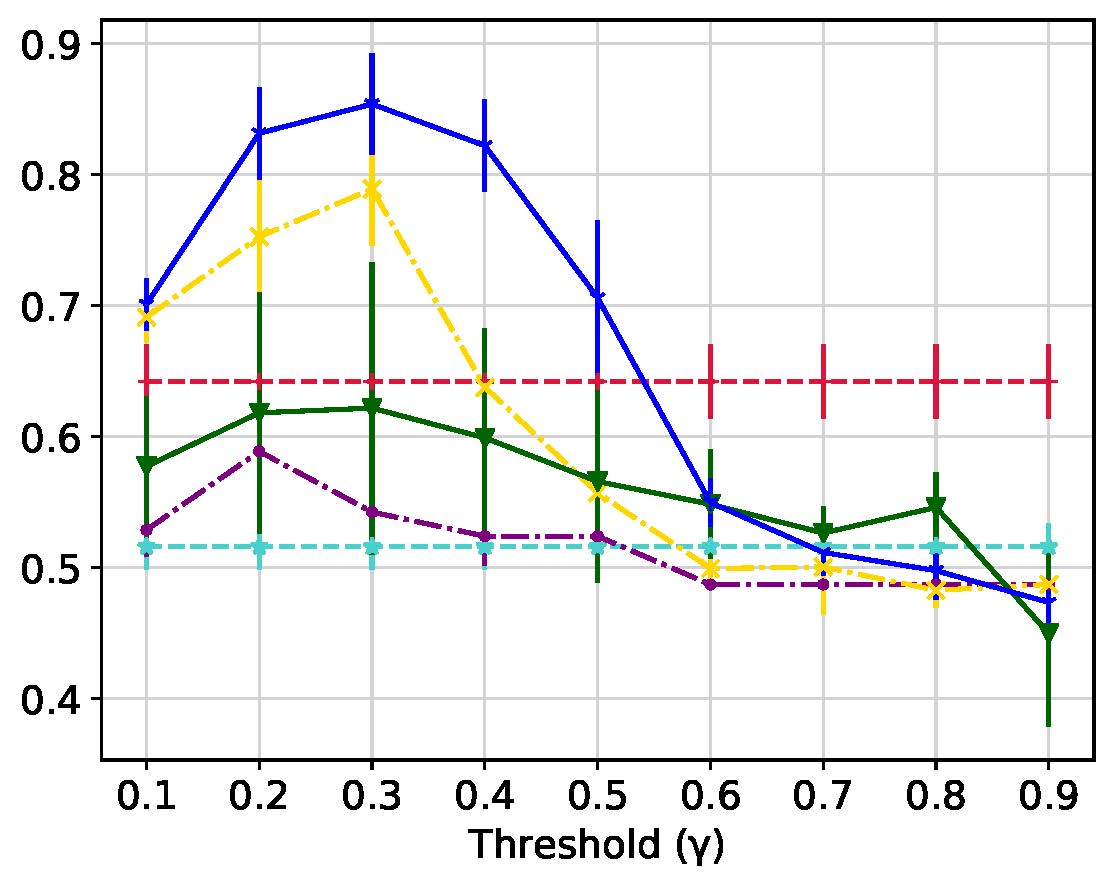}} 
    \hfill
    \subfloat[ROC AUC]{\label{fig:ir-auc}\includegraphics[scale=\myfigurescale]{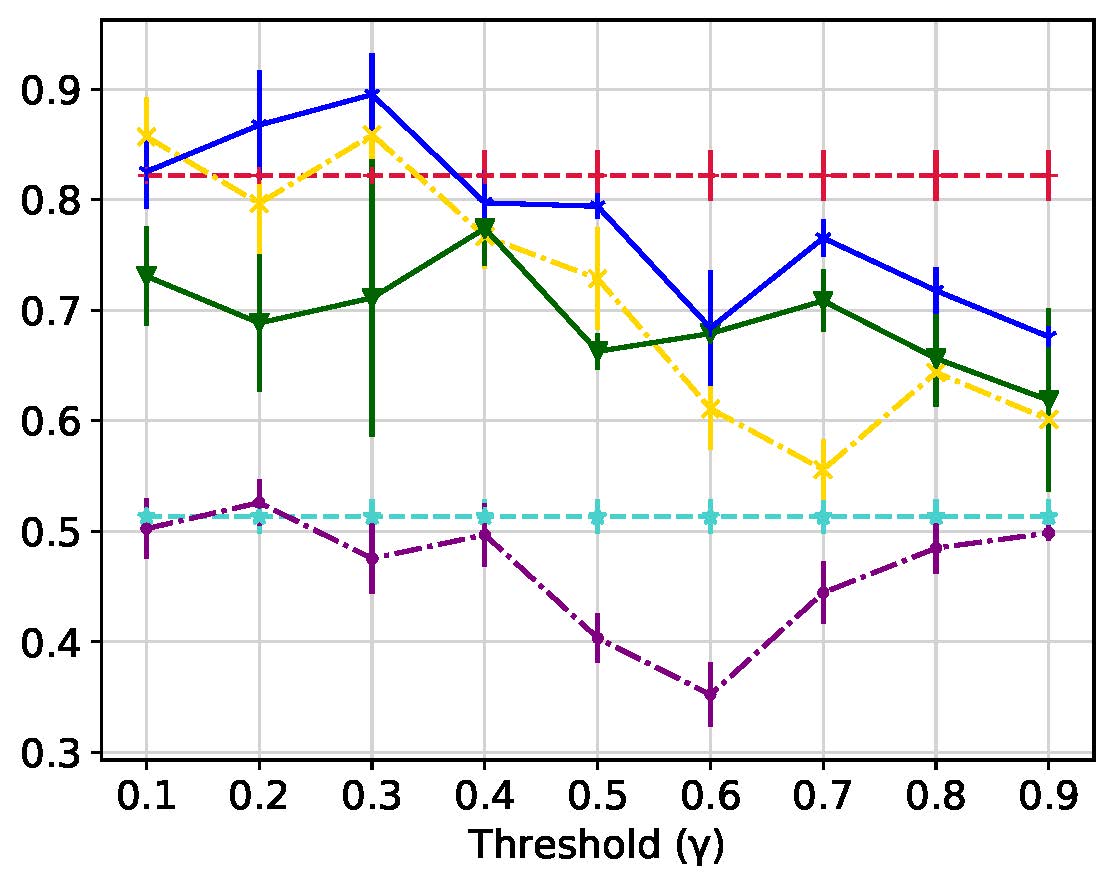}} 
  \end{center}
   \caption{The effect of changing the threshold for graph construction on the performance when $95\%$ of data is in the major class. The figure depicts the performance of the compared methods. In each figure, the threshold for connecting nodes changes from $0.1$ to $0.9$. Increasing the threshold makes the distant nodes to be connected in the graph.}
   \label{fig:thr-95}
   \bigskip
     \begin{center}
    \subfloat[Accuracy]{\label{fig:thr-75-acc}\includegraphics[scale=\myfigurescale]{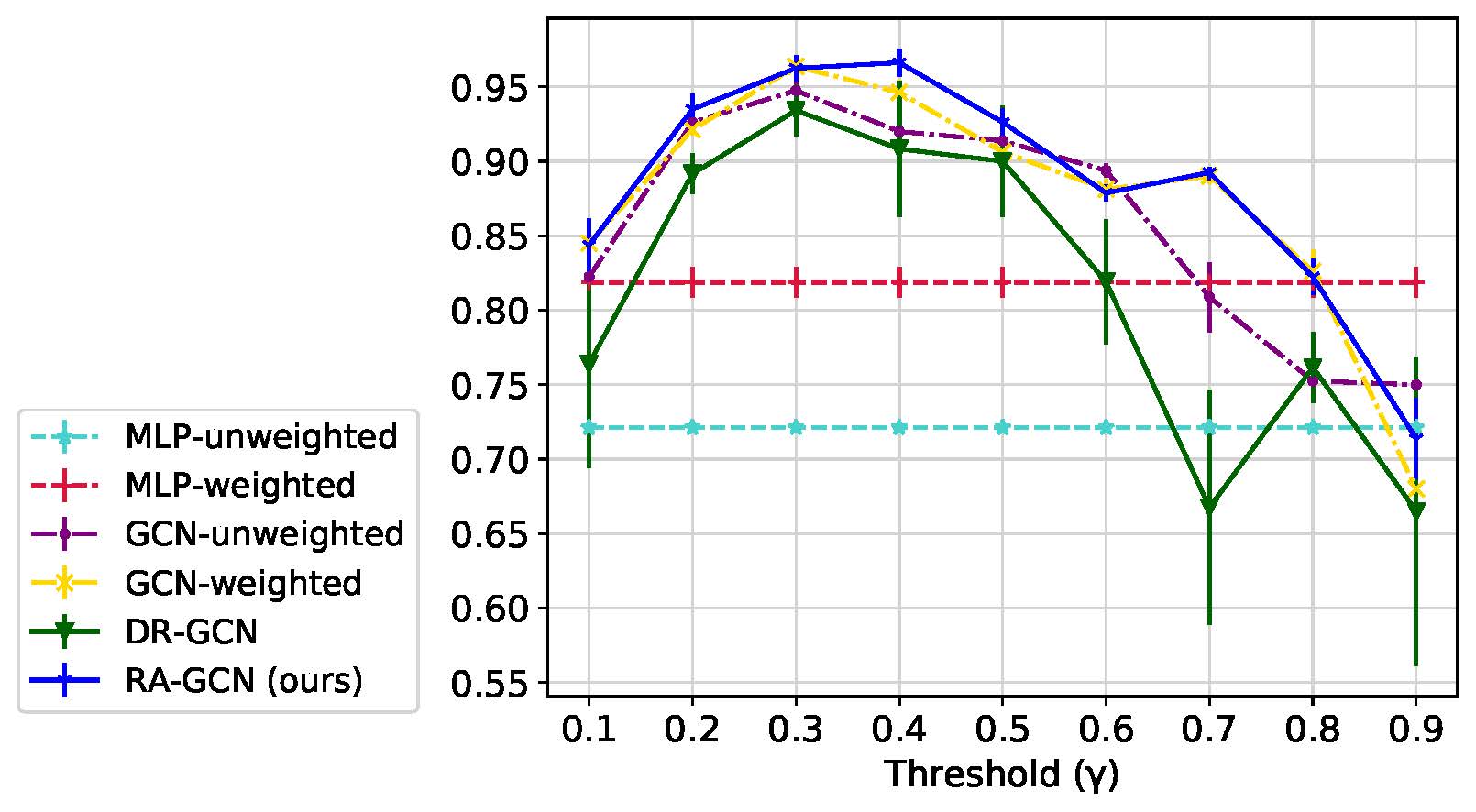}}
    \hfill
    \subfloat[Macro F1]{\label{fig:thr-75-macrof1}\includegraphics[scale=\myfigurescale]{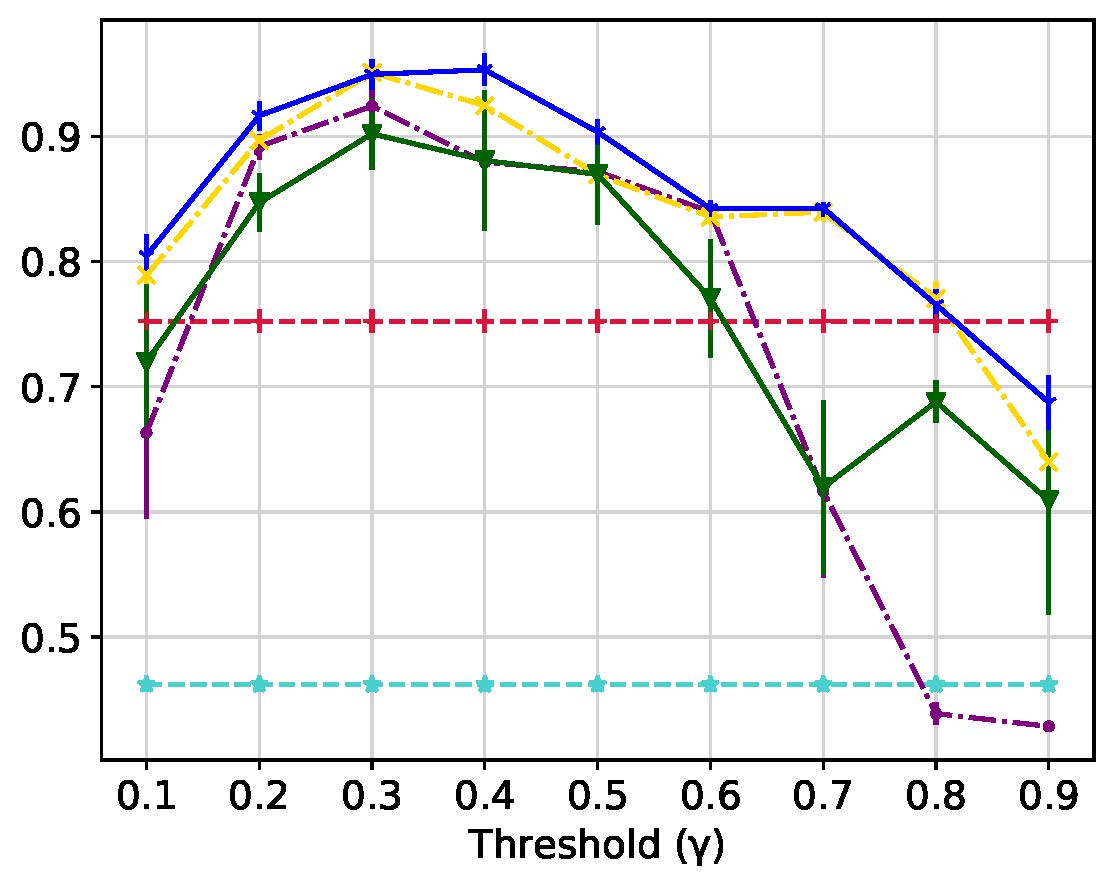}} 
    \hfill
    \subfloat[ROC AUC]{\label{fig:thr-75-auc}\includegraphics[scale=\myfigurescale]{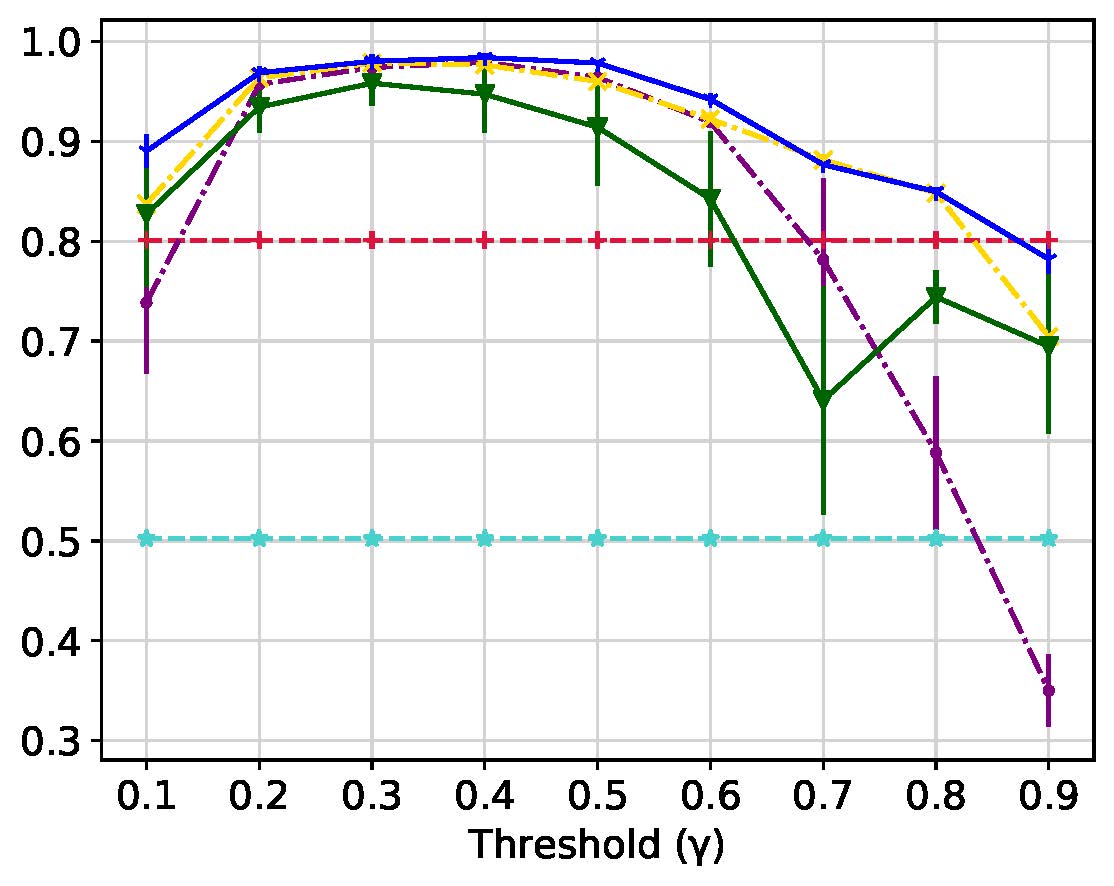}} 
  \end{center}
   \caption{The effect of changing the threshold for graph construction on the performance when $75\%$ of data is in the major class. The figure depicts the performance of the compared methods. In each figure, the threshold for connecting nodes changes from $0.1$ to $0.9$. Increasing the threshold makes the distant nodes to be connected in the graphs.}
   \label{fig:thr-75}
}
\end{figure*}

\textbf{Results}: The results of the highly imbalanced setting are depicted in Fig.~\ref{fig:thr-95}. The results of weighted and unweighted MLP are constant as they are not dependent on the graph. The results frequently illustrate that as the threshold increases and the graph becomes dense, the performance of the graph-based classifiers drops. This reveals that tuning the threshold for graph construction is essential. This experiment once again confirms that in the case of an imbalanced dataset, a dense graph might corrupt the performance by feature propagation through the connections between samples from different classes. Especially when the number of samples in the minor one is low, the inter-class connectivities become more dominant. 
Further, as the graph convolution takes an average between the node features and its neighbors, even two GC layered networks may lead to the smoothed features.

The second experiment's results, with the moderately imbalanced dataset, are depicted in Fig.~\ref{fig:thr-75}. This study also implies that tuning the graph threshold can have a high impact even when the dataset is not highly imbalanced. The best results of the classifier belong to the RA-GCN when the graph threshold is set to $0.4$. 
The weighted and unweighted GCN performances start to drop dramatically for $\gamma > 0.3$ when encountering a highly imbalanced dataset (previous experiment); however, they are more stable on the moderate imbalanced datasets (current experiment). 
This confirms the importance of graph construction alongside the class imbalance issue for the node classification task.

\section{Conclusion}
In this paper, we proposed RA-GCN, a new model to tackle the class imbalance problem by dynamically learning the cross-entropy sample weights besides the GNN-based classifier using adversarial training. We studied the behavior of the proposed model on various input data compared to baselines and the one recent method \cite{ijcai2020-398} in this field.
We tested the method on three real and a set of synthetic datasets. Our results indicate that the proposed model enhances the classifier based on the adopted metrics and outperforms the static weighting of samples with margin. To investigate the impact of different factors on the class imbalance problem, we generated synthetic datasets with different imbalance ratios, different sparsity for the population graph, and a different number of classes. Synthetic experiments show that although the competitive methods have acceptable performance on moderate imbalanced datasets, their efficiency drops in highly imbalanced ones. It is also evident that in GCN-based methods, the graph structure is crucial due to feature propagation throughout the networks.
The detailed benefits of the proposed method in dealing with graph-based class imbalanced datasets are demonstrated in the quantitative and qualitative experimental results.  

Although in this paper, we aimed at handling the class imbalance issue for the graph-based data, a similar approach can be adopted to tackle the imbalance problem in the non-graph datasets. For this purpose, an appropriate neural network architecture has to be used as the classifier and the weighting networks (e.g., convolutional neural network for imaging data).
Moreover, we should note that in the case of a large number of classes in which training the weighting networks is computationally inefficient, a general weighting network instead of class-specific ones could present a scalable solution.
For future research studies, we point out two possible directions. The proposed model is constructed based on the transductive setting. It would be interesting to investigate the models' performance in inductive settings where one has no access to testing node features during the training.
Additionally, the way one can generalize the proposed RA-GCN to a multi-label classification problem where each data can have more than one label in the dataset is also challenging.

\appendix 
\section{Detailed Definition of Metrics}
\label{supsec:metric}
The utilized metrics in the experiments are defined based on the following four measurements: 

\textbf{True-positive ($tp^c$)}: The number of nodes whose true label is $c$ and the predicted class by the classifier is also $c$.

\textbf{True-negative ($tn^c$)}: The number of nodes whose true label is not $c$ and the predicted class by the classifier is also not $c$.

\textbf{False-positive ($fp^c$)}: The number of nodes whose true label is not $c$ but the predicted class by the classifier is $c$.

\textbf{False-negative ($fn^c$)}: The number of nodes whose true label is $c$ but the predicted class by the classifier is not $c$.
 
The following metrics are reported in this section.  

\textbf{Accuracy}: Accuracy is the ratio of nodes that are classified correctly, without considering whether that node is a member of major classes or minor classes. 
 \begin{equation}
     Acc = \frac{\sum_{c=1}^{|C|}tp^c}{N}
 \end{equation}
 
\textbf{Macro F1}: In the classification problem, the F1-score can be defined for every class. F1-score is the harmonic mean of the precision and recall of the classifier performance for that class. The precision of class $c$ is the number of correctly identified samples of class $c$ divided by the number of samples predicted as class $c$. The recall of class $c$ is the number of correctly identified samples of class $c$ divided by the number of samples whose true label is $c$. In binary or multi-class classification, macro averaging means assigning equal weight to all classes and compute an average over their scores. Thus, macro F1 is a useful metric when there is a class imbalance issue in the dataset.
Macro F1 can be calculated as follow:
\begin{equation}
    Macro F1 = \frac{1}{|C|}\sum_{c=1}^{|C|}\frac{2tp^c}{2tp^c+fp^c+fn^c}
\end{equation}

\textbf{Area Under Receiver Operating Characteristic Curve (ROC AUC)}:  \textcolor{\mycolor}{An ROC curve plots true-positive rate vs. false-positive rate at different classification threshold.} ROC AUC means the area under the ROC curve.
The ROC AUC of a classifier is the probability that a randomly selected positive sample will be ranked higher than a randomly selected negative sample.

\section{Details of Synthetic Datasets}
The details about the synthetic datasets of experiments in Section~\ref{sec:exp-binary-ir} (the effect of imbalance ratio) and experiments in Section~\ref{sec:exp-graphsparsity} (the effect of graph sparsity) are provided in Tables.~\ref{tab:syn_data_lh} and \ref{tab:graph-thr-info}, respectively.

\begin{table}[!htb]
\centering
\caption{\textcolor{\mycolor}{Setting of the synthetic datasets with moderate imbalance ratio. The table contains the imbalance ratio and the density of the adjacency matrix generated for evaluating the performance of RA-GCN with changing the imbalance ratios (data of experiments in Section~\ref{sec:exp-binary-ir}).}}
\scriptsize
\begin{tabular}{|c|c|c|c|c|c|}
\hline
\begin{tabular}[c]{@{}c@{}}Dataset \end{tabular}  
& Adj Density & Imbalance Ratio & 
 \begin{tabular}[c]{@{}c@{}}$\%$ of data in  \\ the major class \end{tabular}  
 \\ \hline \hline
\multirow{4}{*}{
 \begin{tabular}[c]{@{}c@{}}Moderate imbalanced  \\ data of Fig.~\ref{fig:irl} \end{tabular} 
}
& $0.109$               & $500/500   = 1$ & $50$   \\ \cline{2-4}
& $0.115$               & $600/400   = 1.5$ & $60$ \\ \cline{2-4}
& $0.126$               & $700/300   = 2.333$ & $70$ \\ \cline{2-4}
& $0.144$               & $800/200   = 4$ & $80$ \\ \hline \hline
\multirow{11}{*}{
\begin{tabular}[c]{@{}c@{}}Highly imbalanced  \\ data of Fig.~\ref{fig:irh} \end{tabular} 
} 
& $0.156$               & $850/150   = 5.667$ & $85$ \\ \cline{2-4}
& $0.159$               & $860/140   = 6.143$  & $86$   \\ \cline{2-4}
& $0.161$               & $870/130   = 6.692$  & $87$  \\ \cline{2-4}
& $0.164$               & $880/120   = 7.334$  & $88$ \\ \cline{2-4}
& $0.166$               & $890/110   = 8.091$  & $89$ \\ \cline{2-4}
& $0.169$               & $900/100   = 9.0$  & $90$  \\ \cline{2-4}
& $0.171$               & $910/90   = 10.112$  & $91$ \\ \cline{2-4}
& $0.173$               & $920/80   = 11.500$  & $92$ \\ \cline{2-4}
& $0.176$               & $930/70   = 13.286$  & $93$  \\ \cline{2-4}
& $0.178$               & $940/60   = 15.667$  & $94$ \\ \cline{2-4}
& $0.181$               & $950/50   = 19$   & $95$ \\ \hline
\end{tabular}
\label{tab:syn_data_lh}
\end{table}

 \begin{table}[!htb]
\centering
\caption{Setting of the synthetic datasets. The table contains the imbalance ratio and the density of the generated adjacency matrix for evaluating the performance of RA-GCN with changing the threshold for connecting the nodes in the graph construction step (data of experiments in Section~\ref{sec:exp-graphsparsity}).}
\scriptsize
\begin{tabular}{|c|c|c|}
\hline
\multirow{3}{*}{
\begin{tabular}[c]{@{}c@{}}Threshold ($\gamma$) \end{tabular} 
} & \multicolumn{2}{c|}{Adj Density}   \\ \cline{2-3} 
                           & IR = 750/250 = 3 & IR = 950/50 = 19  \\
                           & (Datasets of Fig.~\ref{fig:thr-75}) & (Datasets of Fig.~\ref{fig:thr-95}) 
                           \\ \hline \hline
0.1                        & 0.0037                                 & 0.0049                                 \\ \hline
0.2                        & 0.0215                                 & 0.0301                                \\ \hline
0.3                        & 0.0518                                 & 0.0727                                \\ \hline
0.4                        & 0.0894                                 & 0.1242                                 \\ \hline
0.5                        & 0.1347                                 & 0.1813                                \\ \hline
0.6                        & 0.1899                                 & 0.2433                                \\ \hline
0.7                        & 0.2580                                 & 0.3117                                 \\ \hline
0.8                        & 0.3406                                 & 0.3886                                \\ \hline
0.9                        & 0.4355                                 & 0.4735                                \\ \hline
\end{tabular}
\label{tab:graph-thr-info}
\end{table}

\section{Additional Experiments}
\textcolor{\mycolor}{
In this section, we provide three additional experiments on synthetic datasets. In the first experiment, it has been shown that how the class imbalance changes the behavior of a linear classifier trained by RA-GCN and other competitors. In the second experiment, the imbalance issue is studied in datasets with  $3$, $4$, and $5$ classes with various class distributions, and in the last experiment, one of the latest methods for graph learning is employed to study how adding graph learning to the classification task effects the performance in the presence of class imbalance.}
\subsection{Qualitative Study}
\label{sec:exp-qualitative}
\begin{figure*}[!htb]
{
\captionsetup[subfloat]{font=scriptsize,labelfont=scriptsize}
  \begin{center}
  \subfloat[test]
  [ \hspace*{3mm}
  \begin{tabular}{|c|c|c|c|}
    \cline{1-1}
    \multicolumn{1}{|c}{MLP-unweighted} & \multicolumn{1}{|c}{}  \\ \hline 
      Accuracy & $0.884$ \\
      Macro F1 & $0.494$ \\
      ROC AUC & $0.504$ \\ \hline
  \end{tabular}
  ]
  {\label{fig:vis-mlp-unweighted}\includegraphics[width=0.20\textwidth,height=0.20\textheight,keepaspectratio]{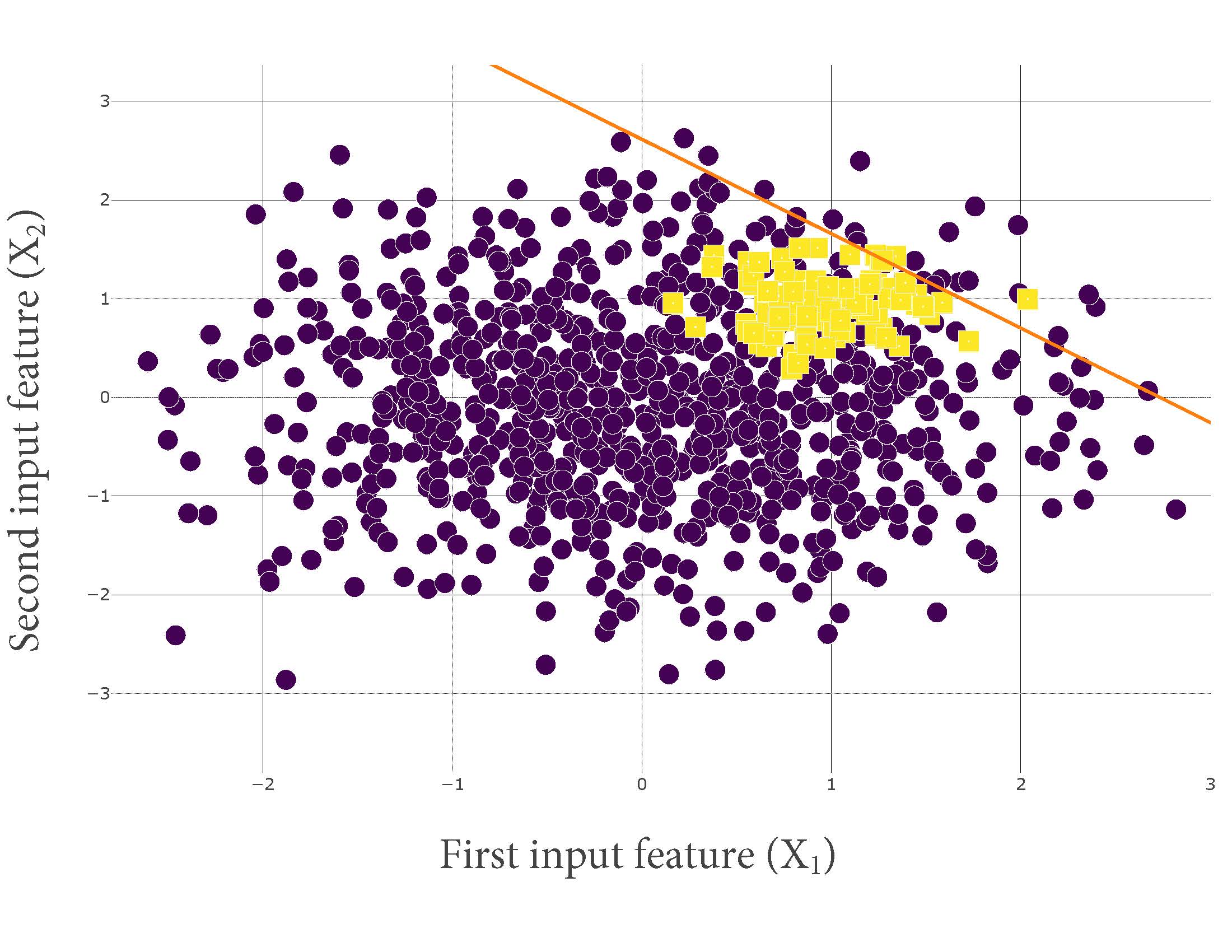}}
 \hfill
  \subfloat
  [ \hspace*{3mm}
  \begin{tabular}{|c|c|c|c|}
    \cline{1-1}
    \multicolumn{1}{|c}{MLP-weighted} & \multicolumn{1}{|c}{}  \\ \hline 
      Accuracy & $0.817$ \\
      Macro F1 & $0.696$ \\
      ROC AUC & $0.867$ \\ \hline
  \end{tabular}
  ]
  {\label{fig:vis-mlp-weighted}\includegraphics[width=0.20\textwidth,height=0.20\textheight,keepaspectratio]{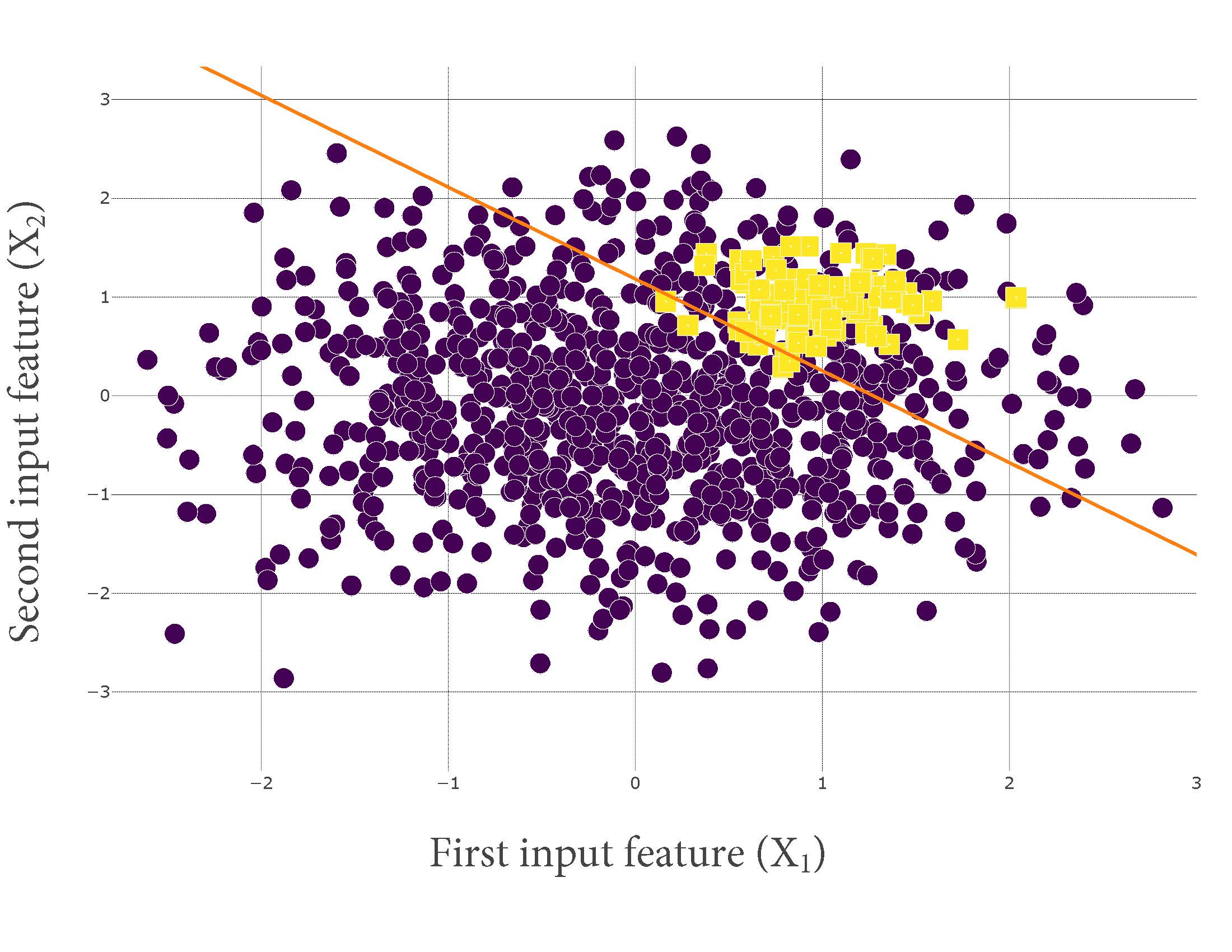}} 
  \hfill
  \subfloat[test]
  [ \hspace*{3mm}
  \begin{tabular}{|c|c|c|c|}
    \cline{1-1}
    \multicolumn{1}{|c}{GCN-unweighted} & \multicolumn{1}{|c}{}  \\ \hline 
      Accuracy & $0.9$ \\
      Macro F1 & $0.474$ \\
      ROC AUC & $0.5$ \\ \hline
  \end{tabular}
  ]
  {\label{fig:vis-gcn-unweighted}\includegraphics[width=0.20\textwidth,height=0.20\textheight,keepaspectratio]{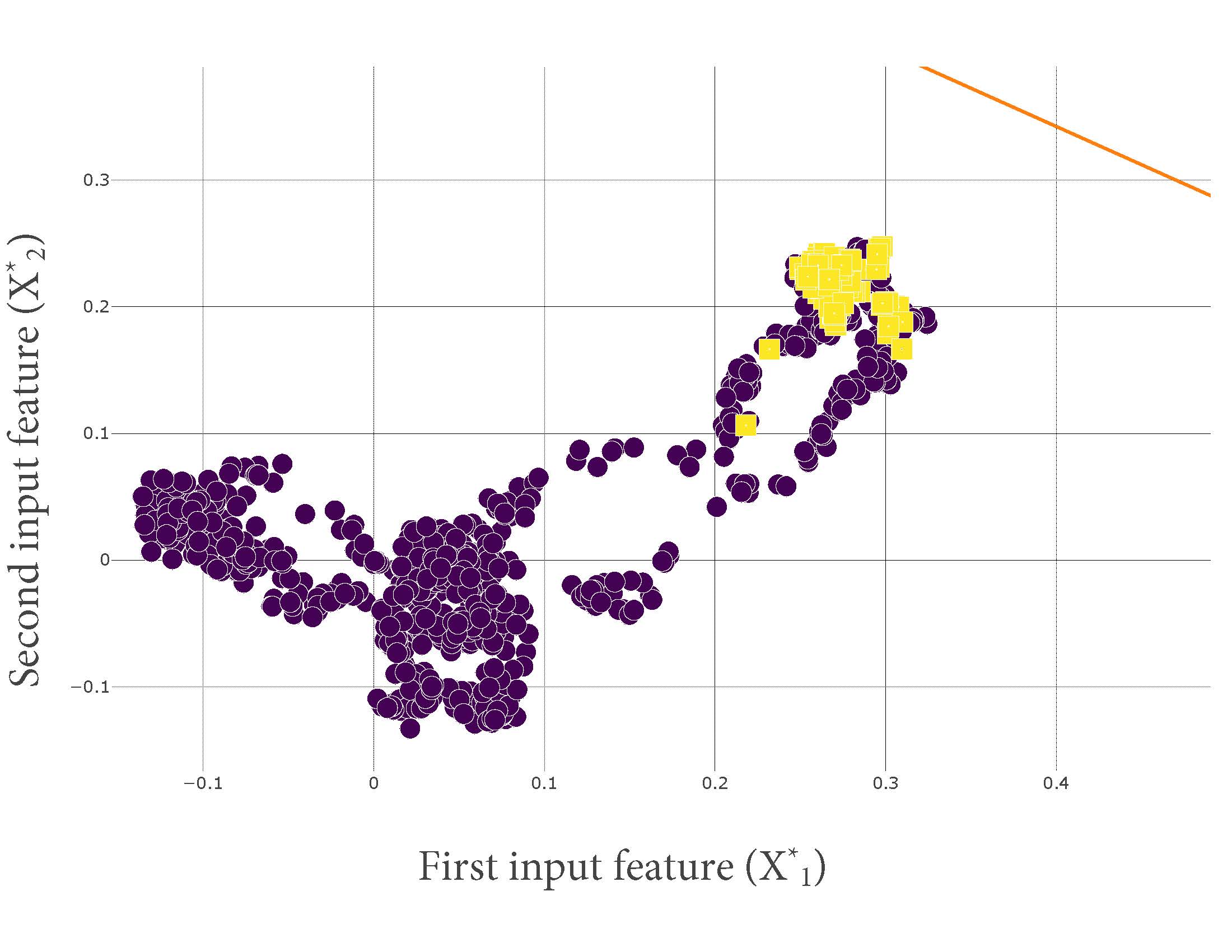}}  
    \hfill
  \subfloat[test]
  [ \hspace*{3mm}
  \begin{tabular}{|c|c|c|c|}
    \cline{1-1}
    \multicolumn{1}{|c}{GCN-weighted} & \multicolumn{1}{|c}{}  \\ \hline
      Accuracy & $0.769$ \\
      Macro F1 & $0.658$ \\
      ROC AUC & $0.872$ \\ \hline
  \end{tabular}
  ]
  {\label{fig:vis-gcn-weighted}\includegraphics[width=0.20\textwidth,height=0.20\textheight,keepaspectratio]{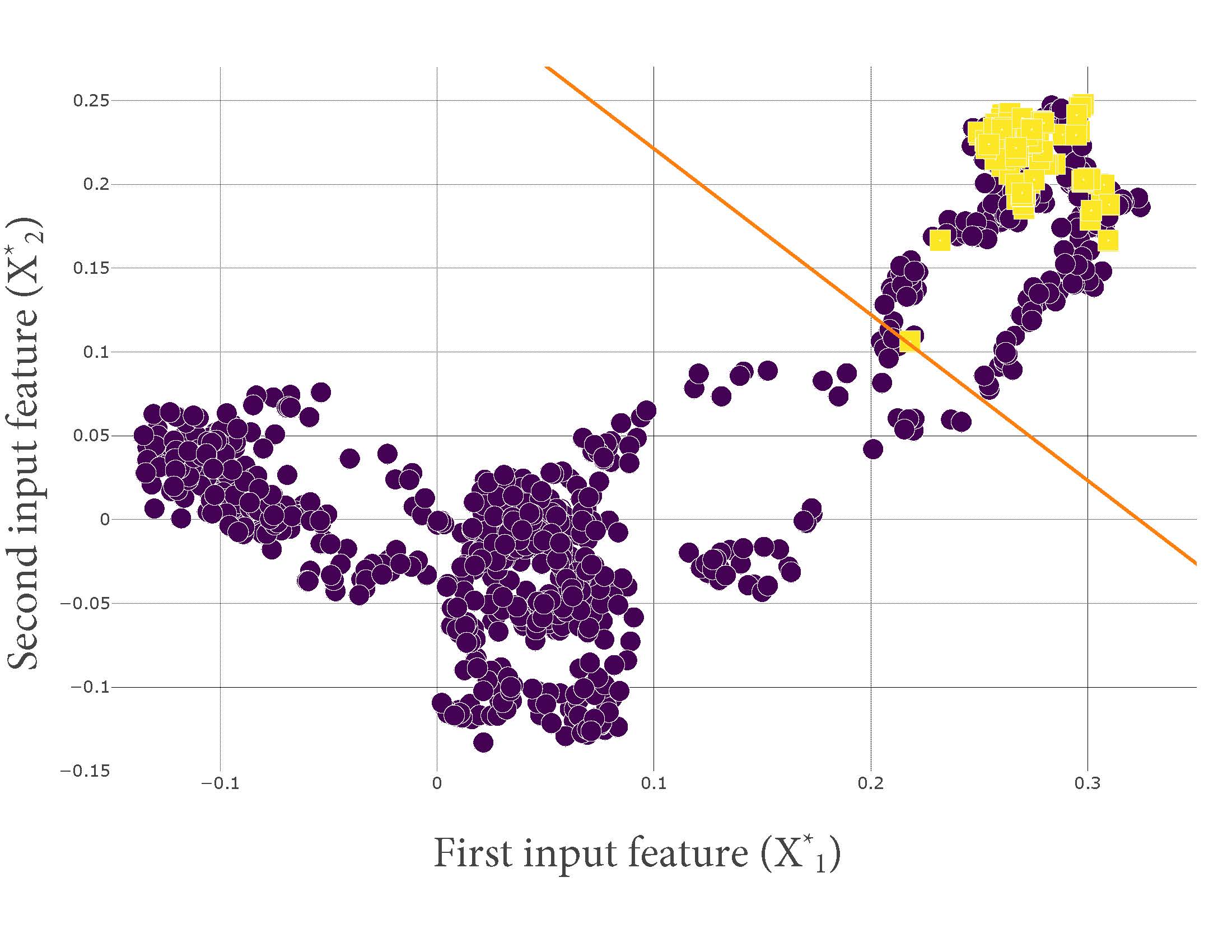}} 
  \hfill
  \subfloat[test]
  [ \hspace*{3mm}
  \begin{tabular}{|c|c|c|c|}
    \cline{1-1}
    \multicolumn{1}{|c}{RA-GCN (ours)} & \multicolumn{1}{|c}{}   \\ \hline 
      Accuracy & $0.873$ \\
      Macro F1 & $0.762$ \\
      ROC AUC & $0.907$ \\ \hline
  \end{tabular}
  ]
  {\label{fig:vis-ragcn}\includegraphics[width=0.20\textwidth,height=0.20\textheight,keepaspectratio]{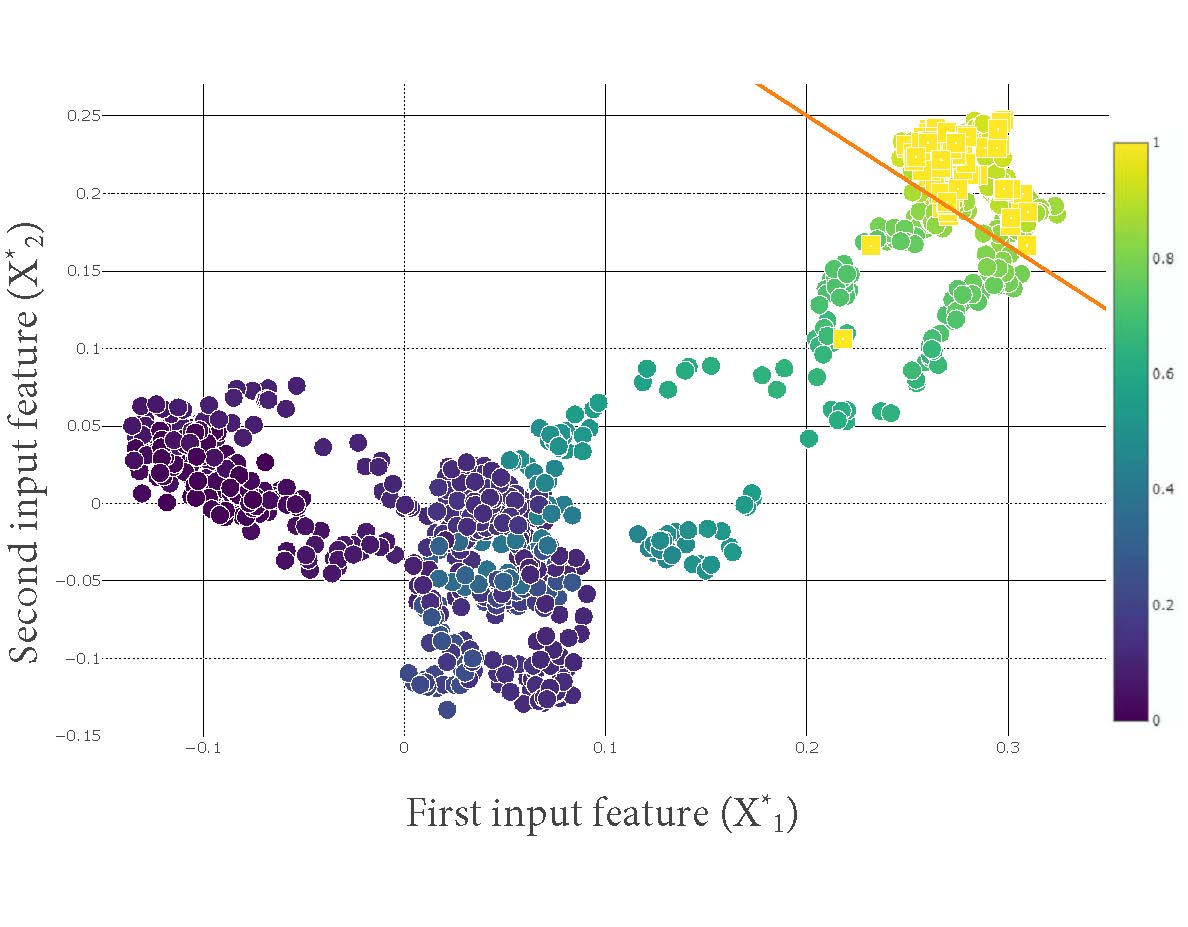}} 
  \end{center}
   \caption{Visualization of the trained classifier on a dataset with two input features. Figure represents the analysis of linear classifiers trained by evaluated methods. MLP, GCN (in weighted and unweighted form), and the proposed method (RA-GCN) are compared. Due to the linearity of the classifiers (without hidden space), the output of DR-GCN is the same as GCN-unweighted. In all of the sub-figures, the orange line represents the learned classifier. In figure (a),(b),(c) and (d) two colors represent the classes. In sub-figure (e) color represents the weight learned by the proposed method. Horizontal and vertical axes show the first and second input features respectively. Since GCN-based methods (c, d, and e) multiply the input feature by adjacency matrix and use a new feature space, their classifier is drawn in their own feature space.}
   \label{fig:vis}
}
\end{figure*}

In this section, we investigate the problem by visualization. For this purpose, we generate $1000$ samples distributed in two classes with a class imbalance ratio of 90:10. Samples of the first class are drawn from a $4$-dimensional Gaussian distribution with mean $0$ and identity covariance matrix ($\CMcal{N}(\bm{0}_{4\times 1},I_{4\times 4})$). On the other hand, the samples from the second class are drawn from another $4$-dimensional Gaussian distribution with mean $1$ and covariance matrix $0.3I_{4\times 4}$ ($\CMcal{N}(\bm{1}_{4\times 1},I_{4\times 4})$).
Out of the four features, two are used for graph construction, and the other two are input features. The 2-dimensional visualization of the input features is illustrated in Fig.~\ref{fig:vis}. In all of the subfigures of Fig.~\ref{fig:vis}, each color (purple, yellow) or shape (circle, square) represents the class. The orange line shows the respective classifier after training. The table below each subfigure reports the performance metrics of the classifier. 
To avoid a confusing diagram, we have abstained from drawing the graph between samples. 

For simplicity, we train each classifier in a linear manner. 
The learned classifier of DR-GCN is not provided here, because the idea of DR-GCN is to apply regularization to the hidden space. Since the linear classifier does not have any hidden layer, the output of DR-GCN is the same as GCN-unweighted. 
Fig.~\ref{fig:vis} depicts the visualization of the trained classifiers with different methods. 

Fig.~\ref{fig:vis-mlp-unweighted} indicates that a simple unweighted classifier ignores most of the minor class samples (Yellow). It acquires high accuracy and low macro F1, as expected.  On the other hand, MLP-weighted (Fig.~\ref{fig:vis-mlp-weighted}) ignores lots of samples in the major class because one sample in the minor class is weighted $9$ times more. On the other hand, it can be seen from Fig.~\ref{fig:vis-gcn-unweighted} and \ref{fig:vis-gcn-weighted} that multiplying the features by the modified adjacency matrix reduces the variance of features and results in the denser clusters with potential class overlap. 
Although the samples' clustering can make them more class separable, the inter-class overlap might make the problem of class imbalance more serious for the graph-based methods. 
In Fig.~\ref{fig:vis-gcn-unweighted}, the high density of samples in one class changes the behavior of the unweighted GCN in comparison to the unweighted MLP. 
Due to the high imbalance, the model becomes highly biased towards the major class. This bias encourages the classifier to misclassify all the minor class samples, represented by very similar features due to low variance in modified features space.
On the other hand, in Fig.~\ref{fig:vis-gcn-weighted}, two outliers in the minor class that are away from other samples can deteriorate the classifier learned with conventional weighted cross-entropy. 
This is because the classifier prefers to correctly classify the outliers (due to the high weight) instead of many samples in the other class. Hence, outliers (noise or mislabeled data) have a significant effect.
Fig.~\ref{fig:vis-ragcn} depicted the classifier learned by RA-GCN. The weights trained by RA-GCN are also illustrated in Fig.~\ref{fig:vis-ragcn} in the heatmap. To better visualize weights, the logarithm of weights is scaled to $[0,1]$ in this figure. As can be seen, each sample is differently weighted, unlike the weighted cross-entropy. RA-GCN resolves the problem by automatically weighing the samples of each class. The entropy term in the objective function from Eq.\ref{eq:ra-gcn} prevents the classifier from emphasizing outliers. Although the minor class's samples have high weights, RA-GCN assigns more weights to the misclassified and boundary points of the major class which keeps the classifier balanced and hinders it from sacrificing the class samples against each other. Metrics also confirm the superiority of the RA-GCN in all measurements. 

\subsection{Experiments on Synthetic Datasets for Multi-Class Classification}
\label{sec:exp-multi}
So far, we have been through analysis, results, and validations on binary class datasets covering the different levels of difficulty.  
In this section, we target more challenging conditions. We examine the proposed method when the class imbalance issue happens in the multi-class classification problem. In this case, there are more than two classes. It should be noted that the datasets are not multi-label, which means that each sample belongs to only one class.

\textbf{Dataset}: We generate $6$ different datasets with $3$, $4$, and $5$ classes. All the datasets contain $2000$ samples. Like the previous experiment, $10$ features are the node features, and the other $10$ are utilized for graph construction. The generated datasets cover the class imbalance problem in different ways. For example, in three classes scenario, we test two cases: 1) Two classes being minor classes ($0.7, 0.2, 0.1$ for data distribution), 2) One class being the minor class ($0.5, 0.4, 0.1$ for data distribution). More details about the six datasets are provided in Table~\ref{tab:multi-class}. The adjacency density results from thresholding the graphs with $0.5$.

\begin{table}[!htb]
\caption{Details about the generated synthetic datasets for multi-class classification problem. The table contains the number of classes, class distributions and the density of the constructed adjacency matrix.}
\scriptsize
\centering
\begin{tabular}{|c|c|c|c|}
\hline
\begin{tabular}[c]{@{}c@{}}Dataset\end{tabular} & \begin{tabular}[c]{@{}c@{}}Number of \\ Classes\end{tabular} & \begin{tabular}[c]{@{}c@{}}Percentage of Data \\ Spread in Classes\end{tabular} & \multicolumn{1}{l|}{Adj Density} \\ \hline \hline
DS1 & 3   & 70, 20, 10            & 0.0025                           \\ \hline
DS2 & 3   & 50,40,10              & 0.0020                           \\ \hline
DS3 & 4   & 60, 25, 10, 5         & 0.0023                           \\ \hline
DS4 & 4   & 45, 35, 12, 8         & 0.0018                           \\ \hline
DS5 & 4   & 35, 30, 25, 10        & 0.0016                           \\ \hline
DS6 & 5   & 50, 20, 15, 10, 5     & 0.0015                           \\ \hline
\end{tabular}
\label{tab:multi-class}
\end{table}

\textbf{Results}: In this experiment, there are more classes, so the problem gets more challenging. The macro F1 is the average of F1-score over all classes. Hence, improvement in the macro F1 means an improvement in all classes, on average, despite their size, and a small increase is meaningful. From the results in Table~\ref{tab:multi-res}, the overall drop in performance across the methods proves that an increase in the number of classes makes the classification task more difficult. When there are three classes, the results on DS1 and DS2 are consistent, although the distribution of data is different. When there are four classes, the problem of class imbalance is more obvious, as can be seen in DS3, DS4, and DS5 compared to DS1 and DS2. 
With the four class scenarios (DS3, DS4, and DS5), the impact of class imbalance is higher in DS3, where there are three minor classes against one major class. 
By comparing the weighted and unweighted methods, it can be concluded that weighting is beneficial for all datasets and methods. With four classes, for the DS5 in which the class imbalance is less severe, weighting is less effective than for DS3 and DS4.
DR-GCN improves the results of GCN-unweighted in DS1, DS3, and DS6. In all these sets, there is one major class. Needless to mention that the information provided by the graph is helpful in all datasets. All the results demonstrate that the proposed method, RA-GCN, outperforms all the competitors. This implies the power of automatically and dynamically weighting for multi-class datasets.

\begin{table*}[!htb]
\caption{Results of the methods on the multi-class datasets. Table includes the performance of the compared methods for different levels of difficulty for the node classification task. The first row represents the class distributions of the corresponding dataset in graphical form.}
\centering
\scalebox{0.65}{
\begin{tabular}{|c|c|c|c|c|c|c|c|c|c|c|c|c|}
\hhline{~|-|-|-|-|-|-|-|-|-|-|-|-|}
\multicolumn{1}{c}{} & 
 \multicolumn{2}{|c}{\includegraphics[width=0.05\textwidth]{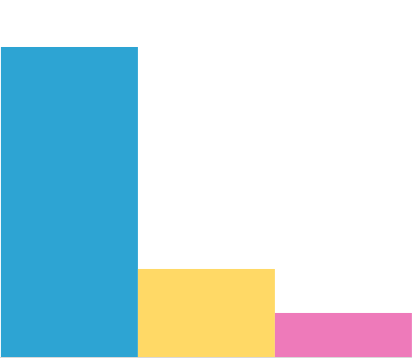}}& 
 \multicolumn{2}{|c}{\includegraphics[width=0.049\textwidth]{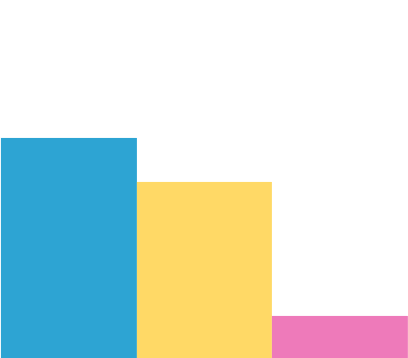}} &  
\multicolumn{2}{|c}{\includegraphics[width=0.05\textwidth]{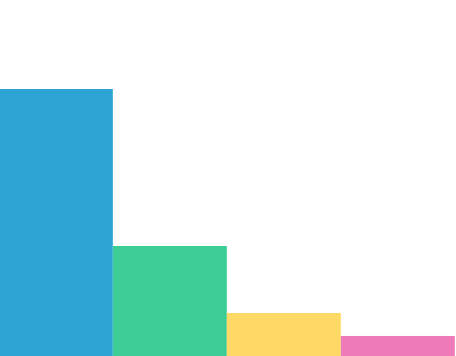}} & 
\multicolumn{2}{|c}{\includegraphics[width=0.05\textwidth]{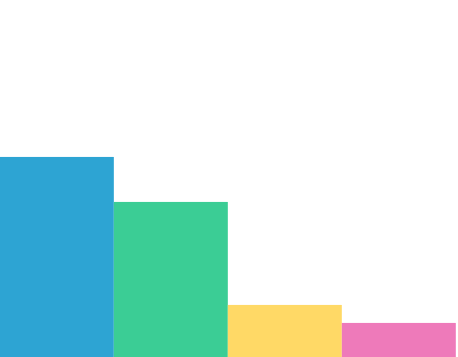}}  & 
\multicolumn{2}{|c}{\includegraphics[width=0.05\textwidth]{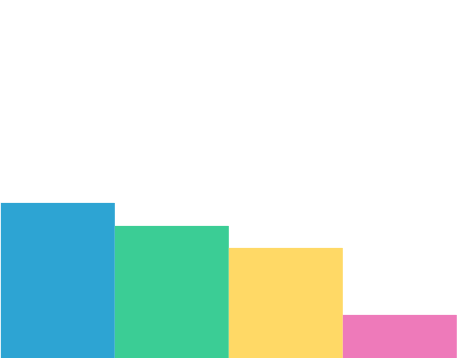}}  & 
\multicolumn{2}{|c|}{\includegraphics[width=0.05\textwidth]{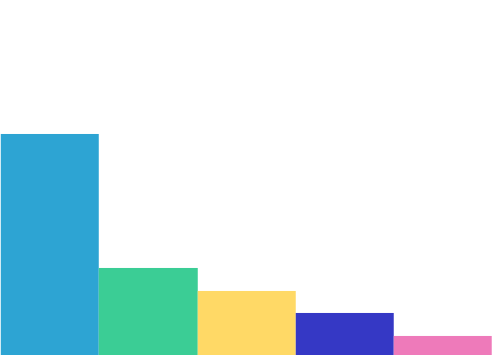}} 
 \\ \hline 
Distribution of Data        & \multicolumn{2}{c|}{DS1: 70, 20, 10} & \multicolumn{2}{c|}{DS2: 50, 40, 10} & \multicolumn{2}{c|}{DS3: 60, 25, 10, 5} & \multicolumn{2}{c|}{DS4: 45, 35, 12, 8} & \multicolumn{2}{c|}{DS5: 35, 30, 25, 10} & \multicolumn{2}{c|}{DS6: 50, 20, 15, 10, 5} \\ \hline \hline
Method/Metric  & Acc          & Macro F1         & Acc          & Macro F1         & Acc           & Macro F1           & Acc           & Macro F1           & Acc            & Macro F1           & Acc             & Macro F1             \\ \hline
MLP-unweighted & $0.81 \mp 0.015$ & $0.54 \mp 0.018$	 & $0.8 \mp 0.014$ & $0.56 \mp 0.01$	 & $0.72 \mp 0.008$ & $0.38 \mp 0.012$	 & $0.69 \mp 0.008$ & $0.4 \mp 0.008$	 & $0.64 \mp 0.009$ & $0.51 \mp 0.018$	 & $0.65 \mp 0.011$ & $0.31 \mp 0.026$\\ \hline
MLP-weighted & $0.77 \mp 0.018$ & $0.67 \mp 0.02$	 & $0.78 \mp 0.018$ & $0.7 \mp 0.019$	 & $0.66 \mp 0.042$ & $0.54 \mp 0.036$	 & $0.64 \mp 0.023$ & $0.56 \mp 0.031$	 & $0.62 \mp 0.022$ & $0.59 \mp 0.018$	 & $0.61 \mp 0.032$ & $0.47 \mp 0.012$\\ \hline
GCN-unweighted & $0.86 \mp 0.004$ & $0.59 \mp 0.011$	 & $0.86 \mp 0.008$ & $0.64 \mp 0.041$	 & $\mathbf{0.79 \mp 0.01}$ & $0.44 \mp 0.035$	 & $\mathbf{0.76 \mp 0.018}$ & $0.49 \mp 0.046$	 & $\mathbf{0.72 \mp 0.021}$ & $0.59 \mp 0.041$	 & $\mathbf{0.7 \mp 0.023}$ & $0.4 \mp 0.053$\\ \hline
GCN-weighted & $0.84 \mp 0.018$ & $0.75 \mp 0.021$	 & $0.84 \mp 0.017$ & $0.77 \mp 0.02$	 & $0.74 \mp 0.062$ & $0.61 \mp 0.047$	 & $0.74 \mp 0.036$ & $0.65 \mp 0.041$	 & $0.7 \mp 0.03$ & $0.65 \mp 0.026$	 & $0.68 \mp 0.023$ & $0.55 \mp 0.028$\\ \hline
DR-GCN & $0.85 \mp 0.012$ & $0.66 \mp 0.031$	 & $0.83 \mp 0.044$ & $0.62 \mp 0.020$	 & $0.75 \mp 0.022$ & $0.45 \mp 0.037$	 & $0.67 \mp 0.038$ & $0.44 \mp 0.026$	 & $0.71 \mp 0.041$ & $0.57 \mp 0.026$	 & $0.69 \mp 0.037$ & $0.42 \mp 0.062$\\ \hline
RA-GCN (ours) & $\mathbf{0.88 \mp 0.023}$ & $\mathbf{0.79 \mp 0.031}$	 & $\mathbf{0.88 \mp 0.013}$ & $\mathbf{0.8 \mp 0.023}$	 & $0.76 \mp 0.037$ & $\mathbf{0.63 \mp 0.027}$	 & $\mathbf{0.76 \mp 0.012}$ & $\mathbf{0.67 \mp 0.016}$	 & $\mathbf{0.72 \mp 0.013}$ & $\mathbf{0.68 \mp 0.017}$	 & $0.68 \mp 0.028$ & $\mathbf{0.56 \mp 0.038}$\\ \hline
\end{tabular}}
\label{tab:multi-res}
\end{table*}

\textcolor{\mycolor}{
\subsection{End-to-end Graph Construction}
\label{sec:exp-graph-construction}
In the previous experiments, an unsupervised approach for graph construction is used. In this experiment, we intend to study the performance of the proposed method besides learning the graph in an end-to-end manner. For this purpose, we adopt the method proposed by Cosmo et al. \cite{cosmo2020latent} for graph learning. The method proposed by Cosmo et al. \cite{cosmo2020latent} is one of the latest graph-learning methods which aims to learn the best graph in an end-to-end manner to optimally support the node-classification task. In the method, first, the node features are embedded into a lower-dimensional space by an MLP network, then a sigmoid function is applied on the linear transformation of the Euclidean distances between every pair of nodes in the low-dimensional space. The parameters of the linear transformation are hyper-parameters and need to be set by the results on validation set. The outcome provides a weighted graph to the rest of the model to complete the node classification task. We added graph-learning method to RA-GCN and GCN as a component called graph-constructor. For DR-GCN, since the implementation is done by authors, to make sure that we are fair to their method, we used the learned graph by GCN-unweighted and input that graph to DR-GCN. Therefore, the construction of the input graph for DR-GCN is not end-to-end. The results are reported on two synthetic datasets with class distributions $[0.95,0.05]$ (highly imbalanced dataset) and $[0.75,0.25]$ (moderate imbalanced dataset).}

\textbf{Results}: The results of the experiment are provided in Table~\ref{tab:graph-learning}.
When the graph-constructor component is added to each method, the number of the parameters grows, and the need for more data increases.
In addition to that, one of the most important factors for constructing the best graph is to have enough samples that enables the model to detect similar samples. 
In the imbalanced datasets, usually, the number of samples in the minor class is limited. 
This issue can be a reason that the results of classification with end-to-end graph construction for each method cannot reach the best results of their counterpart with manual graph construction. 
A comparison between the results of the two datasets also confirms that the methods with graph-constructor shows more deficiency on the higher imbalanced dataset.
However, in the graph construction based on the distance between nodes and without parameters \cite{parisot2017spectral}, the number of samples in each class is not involved. 
It should be also noted that as the imbalance issue biases the classifier, it can also bias the structure of the final graph. Hence, better handling of the issue leads to a better graph and results. 
This can be seen in the results where RA-GCN has achieved superiority over other methods when the graph is constructed during the training.
Moreover, the poor performance of DR-GCN can come from the fact that the learned graph by GCN-unweighted has deteriorated its outcomes.

\begin{table*}[!htb]
\centering
\caption{Results of the compared methods with adding the graph-constructor component to the structure.}
\notsotiny
\color{\mycolor}
\begin{tabular}{|c|c|cH|c||c|cH|c|}
\hline 
\begin{tabular}[c]{@{}c@{}}Class distribution \end{tabular} 
& \multicolumn{4}{c||}{[0.95-0.05]} & \multicolumn{4}{c|}{[0.75-0.25]}
\\ \hhline{|-|-|-|-|-||-|-|-|-|}
Method/Metric & Accuracy & Macro F1 & Binary F1  & ROC AUC & Accuracy & Macro F1 & Binary F1  & ROC AUC 
\\ \hline \hline
\begin{tabular}[c]{@{}c@{}}GCN - unweighted \\ (Best threshold=0.2)\end{tabular}                                    
 & $0.949 \pm 0.0037$ & $0.589 \pm 0.036$ & $0.204 \pm 0.0708$ & $0.526 \pm 0.0207$
 & $0.948 \pm 0.0091$ & $0.924 \pm 0.014$ & $0.883 \pm 0.0224$ & $0.974 \pm 0.0032$
 \\ \hline
\begin{tabular}[c]{@{}c@{}}GCN - weighted\\ (Best threshold=0.3) \end{tabular}                                          
 & $0.958 \pm 0.0142$ & $0.789 \pm 0.0429$ & $0.601 \pm 0.0806$ & $0.859 \pm 0.0443$
 & $0.964 \pm 0.0037$ & $0.951 \pm 0.0053$ & $0.926 \pm 0.0083$ & $0.978 \pm 0.0037$
 \\ \hline
\begin{tabular}[c]{@{}c@{}}RA - GCN\\ (Best threshold=0.3) \end{tabular}                                                        
 & $0.975 \pm 0.0061$ & $0.854 \pm 0.0385$ & $0.721 \pm 0.0737$ & $0.895 \pm 0.037$ 
 & $0.934 \pm 0.0171$ & $0.902 \pm 0.0281$ & $0.846 \pm 0.0459$ & $0.958 \pm 0.0228$
 \\ \hline
  \begin{tabular}[c]{@{}c@{}}DR - GCN\\(Best threshold=0.3)  \end{tabular}
  & $0.793 \pm 0.1414$ & $0.622 \pm 0.1114$ & $0.406 \pm 0.1135$ & $0.711 \pm 0.1256$ 
  & $0.966 \pm 0.0091$ & $0.953 \pm 0.0129$ & $0.929 \pm 0.0199$ & $0.984 \pm 0.0039$ 
  \\ \specialrule{.13em}{.05em}{.05em} 
\begin{tabular}[c]{@{}c@{}}GCN - unweighted \\ + Graph Constructor\end{tabular} 
 & $0.919 \pm 0.0233$ & $0.504 \pm 0.0103$ & $0.051 \pm 0.0321$ & $0.617 \pm 0.0565$ 
 & $0.877 \pm 0.0164$ & $0.809 \pm 0.0286$ & $0.695 \pm 0.0476$ & $0.931 \pm 0.0185$ 
 \\ \hline
\begin{tabular}[c]{@{}c@{}}GCN - weighted\\ + Graph Constructor\end{tabular}
 & $0.804 \pm 0.0827$ & $0.559 \pm 0.0551$ & $0.242 \pm 0.0687$ & $0.717 \pm 0.0239$
 & $0.886 \pm 0.0209$ & $0.857 \pm 0.0233$ & $0.793 \pm 0.0312$ & $0.933 \pm 0.0196$ 
 \\ \hline
 \begin{tabular}[c]{@{}c@{}}DR - GCN\\ + Graph Constructor\end{tabular}
   & $0.895 \pm 0.0321$ & $0.532 \pm 0.0292$ & $0.122 \pm 0.0579$ & $0.663 \pm 0.0334$
   & $0.767 \pm 0.0224$ & $0.687 \pm 0.0216$ & $0.533 \pm 0.0401$ & $0.788 \pm 0.0383$
   \\ \hline
\begin{tabular}[c]{@{}c@{}}RA - GCN\\ + Graph Constructor\end{tabular}          
 & $0.892 \pm 0.0347$ & $0.643 \pm 0.0396$ & $0.347 \pm 0.0657$ & $0.775 \pm 0.0488$ 
 & $0.906 \pm 0.0121$ & $0.876 \pm 0.0157$ & $0.816 \pm 0.0234$ & $0.953 \pm 0.0121$
 \\ \hline
\end{tabular}
\label{tab:graph-learning}
\end{table*}
 
\section*{Acknowledgements}
The authors would like to thank Mojtaba Bahrami for all his help in editing and proof reading the article.

\bibliographystyle{IEEEtran}
\bibliography{IEEEabrv,main}

\end{document}